%% file: main.tex
\documentclass[runningheads]{llncs}

\usepackage{eccv}

\usepackage{eccvabbrv}

\usepackage{graphicx}
\usepackage{booktabs}

\usepackage[accsupp]{axessibility}  

\usepackage[pagebackref,breaklinks,colorlinks,citecolor=eccvblue,hypertexnames=false]{hyperref}

\usepackage{orcidlink}

\usepackage{tabularx}
\usepackage{listings}
\usepackage{algorithm}
\usepackage{multirow}
\usepackage[table]{xcolor}

\begin{document}

\title{Continuous Adversarial Flow Models} 

\titlerunning{Continuous Adversarial Flow Models}

\author{Shanchuan Lin \and
Ceyuan Yang \and
Zhijie Lin \and
Hao Chen \and
Haoqi Fan}

\authorrunning{S.~Lin et al.}

\institute{
ByteDance Seed
}

\maketitle

\footnotetext[1]{Correspondence to: Shanchuan Lin <\email{peterlin@bytedance.com}>}
\footnotetext[2]{Code: \href{https://github.com/ByteDance-Seed/Adversarial-Flow-Models}{https://github.com/ByteDance-Seed/Adversarial-Flow-Models}}

\input{sec/0.abstract}
\input{fig/hero}
\input{sec/1.introduction}
\input{sec/2.background}
\input{sec/3.method}
\input{sec/4.experiment}
\input{sec/5.related}
\input{sec/6.conclusion}
\input{sec/7.acknowledgment}


%
%
\bibliographystyle{splncs04}
\bibliography{main}

\newpage
\appendix
\input{sup/sup}

\end{document}

%% file: sec/0.abstract.tex
\begin{abstract}
    We propose continuous adversarial flow models, a type of continuous-time flow model trained with an adversarial objective. Unlike flow matching, which uses a fixed mean-squared-error criterion, our approach introduces a learned discriminator to guide training. This change in objective induces a different generalized distribution, which empirically produces samples that are better aligned with the target data distribution. Our method is primarily proposed for post-training existing flow-matching models, although it can also train models from scratch. On the ImageNet 256px generation task, our post-training substantially improves the guidance-free FID of latent-space SiT from 8.26 to 3.63 and of pixel-space JiT from 7.17 to 3.57. It also improves guided generation, reducing FID from 2.06 to 1.53 for SiT and from 1.86 to 1.80 for JiT. We further evaluate our approach on text-to-image generation, where it achieves improved results on both the GenEval and DPG benchmarks.

    \keywords{Generative models \and Adversarial training \and Flow models}
\end{abstract}

%% file: fig/hero.tex
\vspace{-20pt}
\begin{figure}
    \setlength{\tabcolsep}{0pt}
    \centering
    \captionsetup{justification=raggedright,singlelinecheck=false}
    \begin{subfigure}[t]{\linewidth}
        \begin{tabularx}{\linewidth}{XXXXX}
            \includegraphics[width=\linewidth]{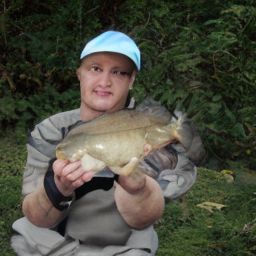} &
            \includegraphics[width=\linewidth]{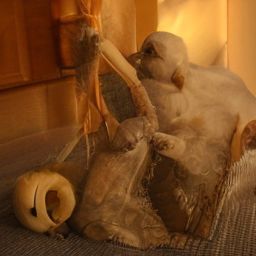} &
            \includegraphics[width=\linewidth]{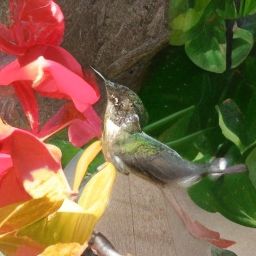} &
            \includegraphics[width=\linewidth]{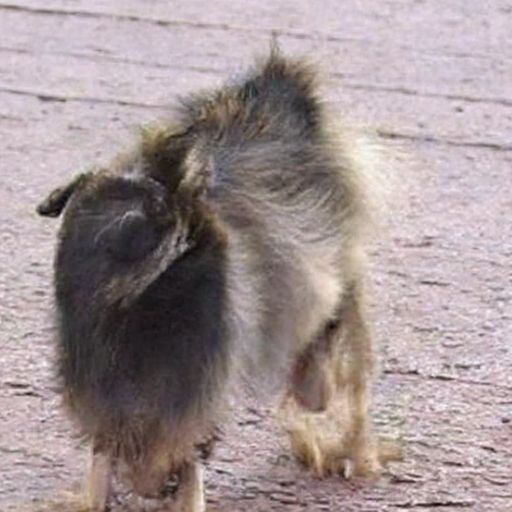} &
            \includegraphics[width=\linewidth]{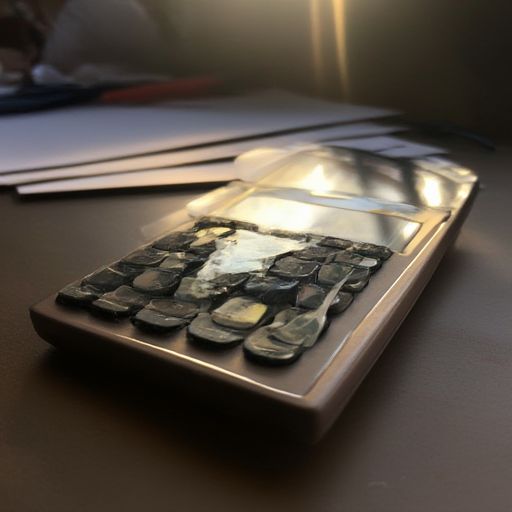}\\
        \end{tabularx}
        \vspace{-5pt}
        \caption*{\textbf{Top:} Flow Matching}
        \vspace{2pt}
    \end{subfigure}
    \begin{subfigure}[t]{\linewidth}
        \begin{tabularx}{\linewidth}{XXXXX}
            \includegraphics[width=\linewidth]{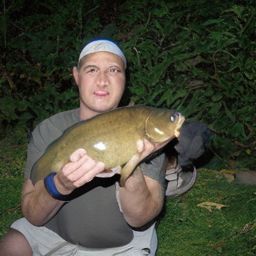} &
            \includegraphics[width=\linewidth]{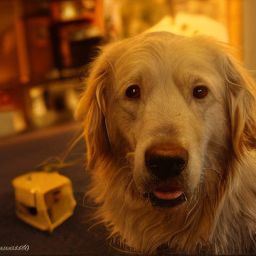} &
            \includegraphics[width=\linewidth]{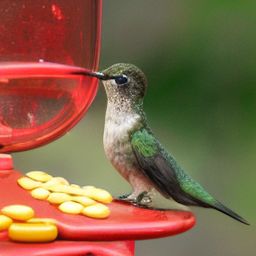} &
            \includegraphics[width=\linewidth]{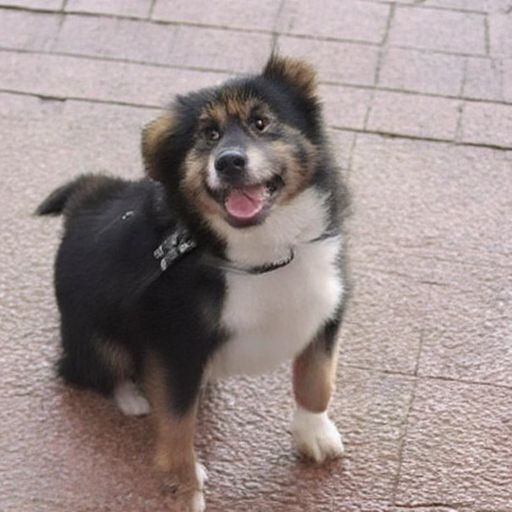} &
            \includegraphics[width=\linewidth]{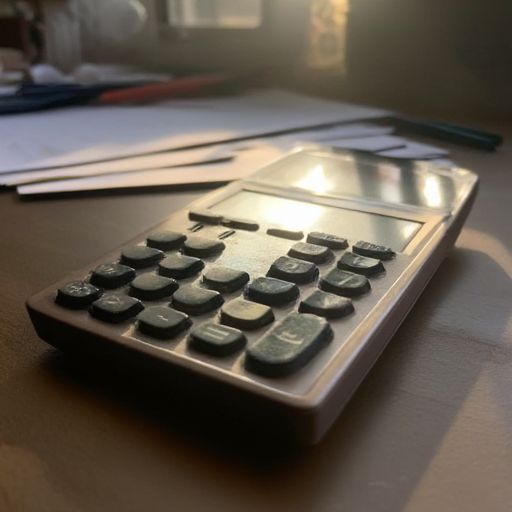}\\
        \end{tabularx}
        \vspace{-5pt}
        \caption*{\textbf{Bottom:} Continuous Adversarial Flow Models (Ours)}
    \end{subfigure}
    \vspace{-5pt}
    \captionsetup{justification=centering}
    \caption{Generation without guidance. Our method yields better generalization.}
    \label{fig:placeholder}
\end{figure}
\vspace{-25pt}

%% file: sec/1.introduction.tex
\section{Introduction}

Flow matching~\cite{lipman2023flow} has achieved significant success in recent years, yet a critical problem remains. The issue is particularly evident in the generation of visual modalities, such as image~\cite{seedream2025seedream,cai2025z} and video~\cite{seawead2025seaweed,gao2025seedance,seedance2025seedance} synthesis, where models often produce out-of-distribution samples unless guidance is applied~\cite{dhariwal2021diffusion,ho2021classifierfree,karras2024guiding}. While guidance improves sample quality, it alters the sampling distribution. How to more faithfully model the underlying distribution of the original data remains an open problem.

One reason flow matching generates out-of-distribution samples is that it uses a Euclidean distance criterion rather than a manifold-aware one. Concretely, flow matching (FM) learns the velocity field of a probability flow~\cite{song2021scorebased} between noise and data distributions. Training minimizes the squared $L_2$ loss between predicted and target velocities. In theory, this objective converges to the ground-truth flow under an infinite-capacity model, which would overfit and reproduce the training samples exactly. In practice, models with finite capacity must generalize, and therefore results in the generation of new data samples. However, the squared $L_2$ objective measures Euclidean distance rather than the manifold-aware distance, inducing incorrect generalization relative to the underlying data distribution.

Recent work has attempted to tackle the issue from different angles. Representational autoencoders~\cite{zheng2025diffusion} convert the data space on which flow matching operates and have empirically reported improvements in generation quality, but this requires operating in a latent space instead of the original data space. Riemannian flow matching~\cite{chen2023flow} extends flow matching to non-Euclidean geometries, but this requires manual definition of the data manifold, which is often unknown for general datasets. Other work~\cite{lin2023diffusion} replaces Euclidean loss with perceptual distances derived from frozen feature networks, motivated by the empirical finding that deep networks can serve as better perceptual metrics~\cite{zhang2018unreasonable}. However, a fixed criterion network can be exploited by the generator~\cite{goodfellow2014explaining}, leading to artifacts in the generated samples. A way to mitigate generator hacking is to jointly train the criterion network with the generator, which yields a dynamic reminiscent of generative adversarial networks.

Generative adversarial networks (GANs)~\cite{goodfellow2014generative} are a standalone class of generative methods. They achieve strong performance on ImageNet benchmarks~\cite{sauer2022stylegan,huang2024gan,hyun2025scalable,lin2025adversarial} and are widely used in flow-model distillation for sharp image synthesis~\cite{lin2025diffusion,lin2024sdxl,lin2024animatediff,ren2024hyper,xu2024ufogen,yin2024improved,sauer2024adversarial,sauer2024fast,lin2025autoregressive}. We hypothesize that this advantage arises because the discriminator networks are more sensitive to perceptual details, \eg texture, sharpness, contour, \etc, than pointwise Euclidean losses, because they may have learned to better capture the manifold structure. Recent work, adversarial flow models (AFMs)~\cite{lin2025adversarial}, combines adversarial and flow modeling, improving training stability and extending adversarial objectives to multi-step flow training. However, AFMs are formulated in discrete time, leaving open the question of how to incorporate adversarial training into continuous-time flow modeling.

In this paper, we introduce continuous adversarial flow models (CAFMs), which extend AFMs to continuous time. CAFMs are a type of continuous normalizing flow (CNF)~\cite{chen2018neural} that generates samples by integrating an ordinary differential equation (ODE) from noise to data. Like flow-matching models (FMs), CAFMs also learn the velocity field of a predefined probability flow with a simulation-free objective. Although FMs and CAFMs target the same ground-truth flow, they differ in finite-capacity generalization because CAFMs use a learned discriminator rather than a fixed Euclidean criterion. Empirically, our experiments find that CAFMs produce more in-distribution samples, both perceptually and by various metrics. To the best of our knowledge, our work is the first to apply adversarial training in continuous-time flow modeling.

Since FMs and CAFMs learn the same ground-truth flow and differ mainly in model generalization, our method is primarily designed to post-train existing FMs for efficiency and practicality, although the objective can also be used for training from scratch. In class-conditional ImageNet~\cite{russakovsky2015imagenet} 256px generation, CAFM post-training improves the guidance-free FID for latent-space SiT~\cite{ma2024sit} from 8.26 to 3.63, and for pixel-space JiT~\cite{li2025back} from 7.17 to 3.57, using only 10 epochs of finetuning. CAFMs also achieve better guided generation, improving the FID from 2.06 to 1.53 for SiT and from 1.86 to 1.80 for JiT. In text-to-image experiments, CAFMs increase the GenEval~\cite{ghosh2023geneval} score from 0.81 to 0.85 and the DPG~\cite{hu2024ella} score from 83.7 to 85.2. These results suggest promising prospects for integrating adversarial training into continuous-time flow modeling---not for few-step generation, but for improving sample fidelity and distribution matching.

%% file: sec/2.background.tex
\section{Background}

\subsection{Flow Matching}

Flow matching (FM)~\cite{lipman2023flow} formulates the generation problem as transporting samples from a prior distribution $z\sim \mathcal{Z} \in \mathbb{R}^n$, often Gaussian $\mathcal{N}(0,\mathbf{I})$, to the data distribution $x\sim\mathcal{X} \in \mathbb{R}^n$ over a probability flow, defined by an interpolation function:
\begin{equation}
    x_t = A(t) x + B(t)z,
\end{equation}
where $t\in[0,1]$. Linear interpolation~\cite{liu2023flow,lipman2023flow} is commonly used, where $A(t)=1-t$, $B(t)=t$, and $x_t=(1-t)x+tz$.

The time derivative at position $x_t$ conditioned on $x$ and $z$, called the conditional velocity $\bar{v}_t$, can be derived as:
\begin{equation}
    \bar{v}_t = \frac{dA(t)}{dt}x+\frac{dB(t)}{dt}z,
\end{equation}
which for linear interpolation, $\frac{dA(t)}{dt}=-1$, $\frac{dB(t)}{dt}=1$, and $\bar{v}_t=-x+z$.

Flow matching trains a generator $G(x_t, t):\mathbb{R}^n\times[0,1]\rightarrow\mathbb{R}^n$ to match the conditional velocity $\bar{v}_t$:
\begin{equation}
    \mathcal{L}_{\mathrm{FM}}= \mathbb{E}_{x,z,t} \left[d(G(x_t, t), \bar{v}_t)\right],
\end{equation}
and finds that this conditional flow matching objective in expectation over independent coupling of $x,z$ learns the marginal velocity $v_t=\mathbb{E}[\bar{v}_t\mid x_t]$ of the probability flow when the criterion $d(a,b)$ satisfies:
\begin{equation}
    \arg\min_a \mathbb{E}_b[d(a, b)] = \mathbb{E}[b].
    \label{eq:d_requirement}
\end{equation}
The squared $L_2$ criterion is adopted. The mean squared error (MSE) variant, with an additional factor of $\frac{1}{n}$, is most commonly used:
\begin{equation}
    \mathcal{L}_{\mathrm{FM}}= \mathbb{E}_{x,z,t} \left[\frac{1}{n}\|G(x_t, t)-\bar{v}_t\|^2_2\right].
\end{equation}

The resulting generator $G(x_t,t)$ defines a continuous-time flow model that predicts the marginal velocity field $v_t$ at each state $x_t$ along the probability flow. Samples are transported from the noise distribution to the data distribution by integrating the ODE:
\begin{equation}
    x_0 = x_1 + \int_0^1 G(x_t, t)\ dt, \quad x_1\sim\mathcal{Z},
\end{equation}
where the integration runs backward from $t=1$ to $t=0$.

\paragraph{The limitation of flow matching.} Using any criterion $d(a,b)$ satisfying \cref{eq:d_requirement} in theory ensures the model converges to $v_t =\mathbb{E}[\bar{v}_t\mid x_t]$, but this overfits to generating only the training samples. In practice, models parameterized by neural networks have finite capacity and learn a generalized distribution. In this case, the loss objective affects the way of generalization. Consider:
\begin{equation}
    d(a,b) = (a-b)^\top M (a-b),
\end{equation}
where the squared $L_2$ criterion corresponds to the special case $M=I$. In general, $M$ can be any strictly positive definite matrices (\cref{app:flow-matching-objective}). These objectives converge to the same ground-truth flow, but can induce different generalizations. Flow matching minimizes the isotropic Euclidean distance without awareness of the data manifold, leading to incorrect generalization and out-of-distribution generation.

\paragraph{Using a manifold criterion.} A natural idea is to replace the squared $L_2$ criterion with one that measures distance on the data manifold. However, the underlying data manifold is not known in advance and must itself be inferred and generalized from the limited training data. Our work explores adversarial training, where a criterion network is learned simultaneously along with the generator. This is encouraged by previous empirical findings that deep networks can better capture the data manifold, as evidenced by their ability to serve as a better perceptual distance than the Euclidean metric~\cite{lin2023diffusion,zhang2018unreasonable}. 

\subsection{Adversarial Flow Models}

Adversarial flow models (AFMs)~\cite{lin2025adversarial} are a type of discrete-time flow model trained with an adversarial objective. The training involves a generator $G(x_s, s, t) : \mathbb{R}^n\times[0,1]\times[0,1] \rightarrow \mathbb{R}^n$ that transports samples from source $x_s$ to target $x_t$ on the probability flow, and a discriminator $D(x_t, t):\mathbb{R}^n\times[0,1]\rightarrow\mathbb{R}$ that differentiates the real and generated $x_t$ samples.

Adversarial training involves a minimax optimization game where $D$ aims to maximize discrimination while $G$ aims to minimize discrimination by $D$. The adversarial objective is defined as:
\begin{align}
    \mathcal{L}^D_\textrm{adv} = \mathbb{E}_{x,z,s,t} \left[ f\big(D(x_t, t), D(G(x_s, s, t), t)\big) \right],\\
    \mathcal{L}^G_\textrm{adv} = \mathbb{E}_{x,z,s,t} \left[ f\big(D(G(x_s, s, t), t), D(x_t, t)\big) \right],
\end{align}
where $f(a,b)=-\log(\mathrm{sigmoid}(a - b))$ is one of many viable contrastive functions used by recent work~\cite{lin2025adversarial,huang2024gan,jolicoeur2018relativistic,hudson2021generative}. Training updates $G$ and $D$ in alternation and reaches equilibrium when $G(x_s, s, t)$ produces the same distribution of $x_t$.

AFMs additionally introduce an optimal transport objective on $G$:
\begin{equation}
    \mathcal{L}^G_\textrm{ot} = \mathbb{E}_{x,z,s,t} \left[ \frac{1}{n} \cdot \frac{1}{|t-s|}\cdot \| G(x_s, s, t) - x_s \|^2_2 \right].
    \label{eq:ot_discrete}
\end{equation}
It minimizes the distance to $x_s$, not $\bar{v}_s$, unlike flow matching. Over the expectation, this encourages $G$ to predict targets $x_t$ that are closest to the sources $x_s$, allowing $G$ to learn a unique optimal transport for stable training.

Additionally, $D$ is regulated by gradient penalties $R_1$ and $R_2$~\cite{roth2017stabilizing} to mitigate the problem of vanishing gradient~\cite{arjovsky2017towards} and a centering penalty~\cite{karras2018progressive} to prevent logit drifting:
\begin{align}
    \mathcal{L}_\mathrm{r1}^D &= \mathbb{E}_{x,z,s,t} \left[\| \nabla_{x_t} D(x_t,t) \|^2_2\right],\\
    \mathcal{L}_\mathrm{r2}^D &= \mathbb{E}_{x,z,s,t}\left[ \| \nabla_{G(x_s,s,t)}D(G(x_s,s,t), t)\|^2_2 \right], \\
    \mathcal{L}_\mathrm{cp}^D &= \mathbb{E}_{x,z,s,t} \left[ ( D(x_t,t) + D(G(x_s, s, t), t) )^2 \right].
\end{align}

The final training objectives of AFMs are:
\begin{align}
    \mathcal{L}^D_\textrm{AFM} &= \mathcal{L}^D_\textrm{adv} + \lambda_\textrm{gp}\mathcal{L}^D_\textrm{r1} + \lambda_\textrm{gp}\mathcal{L}^D_\textrm{r2} + \lambda_\textrm{cp}\mathcal{L}_\mathrm{cp}^D, \\
    \mathcal{L}^G_\textrm{AFM} &= \mathcal{L}^G_\textrm{adv} + \lambda_\textrm{ot}\mathcal{L}^G_\textrm{ot}.
\end{align}

To generate, AFMs transport samples from the noise distribution to the data distribution by solving the difference equation:
\begin{equation}
    x_0 = x_1 + \sum_{i=1}^S \big(G(x_{\tau_i}, \tau_i, \tau_{i-1})-x_{\tau_i}\big), \quad x_1 \sim \mathcal{Z},
\end{equation}
where the summation runs backward from $i=S$ to $i=1$ with a total of $S$ sampling steps, and $\tau$ is a list of discrete timesteps satisfying $\tau_0=0,\ \tau_S=1$.

\paragraph{The limitation of adversarial flow models.} AFMs are a form of discrete-time flow models. Although the timestep interval $|t-s|$ can be made arbitrarily small, the training becomes increasingly unstable, and the objective breaks down when $|t-s| = 0$. It is not clear how to extend adversarial training to continuous-time flow modeling. Furthermore, AFMs still have the gradient-vanishing problem~\cite{arjovsky2017towards}. They rely on gradient penalties~\cite{roth2017stabilizing}, discriminator augmentation~\cite{karras2020training}, and discriminator reset~\cite{lin2025adversarial} to mitigate the issue.

%% file: sec/3.method.tex
\section{Method}

\subsection{Continuous Adversarial Flow Models}

We propose continuous adversarial flow models (CAFMs) to extend adversarial training to continuous-time flow modeling. Our method involves a generator $G(x_t, t):\mathbb{R}^n\times[0,1]\rightarrow\mathbb{R}^n$ of the same form as in flow matching, which predicts the velocity field $v_t$ at $x_t$, and a discriminator $D(x_t, t) : \mathbb{R}^n\times[0,1]\rightarrow\mathbb{R}$ of the same form as in AFMs. Unlike discrete-time adversarial training, we discriminate $v_t$ in the derivative space of $D$, explicitly reflecting the physical property of velocity $v_t$ as a derivative of position $x_t$.

Specifically, we denote the Jacobian-Vector Product (JVP) of $D$ with primal $(x_t, t)$ and tangent $(\dot{x}_t,\dot{t})$ as:
\begin{equation}
    D_\mathrm{jvp}(x_t, t, \dot{x}_t, \dot{t}) = \frac{\partial D(x_t,t)}{\partial x_t}\dot{x}_t + \frac{\partial D(x_t,t)}{\partial t}\dot{t},
    \label{eq:d_jvp}
\end{equation}
where:
\begin{equation}
    \frac{\partial D(x_t,t)}{\partial x_t}\in\mathbb{R}^{1\times n} \quad \text{and} \quad \frac{\partial D(x_t,t)}{\partial t}\in\mathbb{R}^{1\times 1}
\end{equation}
are the Jacobian matrices of the actual network $D(x_t, t)$ with respect to the primal variables $x_t$ and $t$. The entire JVP function also outputs a scalar, which we use as the discrimination logit:
\begin{equation}
    D_\mathrm{jvp}(x_t, t, \dot{x}_t, \dot{t}) : (\mathbb{R}^n\times[0,1]\times\mathbb{R}^n\times[0,1])\rightarrow\mathbb{R}.
\end{equation}

During training, $D_\mathrm{jvp}$ is evaluated using $(x_t, t)$ as primal and $(\bar{v}_t, T)$ as tangent, where $T=1$ for networks trained with $t\in[0,1]$. The continuous-time adversarial objectives are defined as:
\begin{align}
    \mathcal{L}_\mathrm{adv'}^D = \mathbb{E}_{x,z,t} [ f(&D_\mathrm{jvp}(x_t, t, \bar{v}_t, T),
    D_\mathrm{jvp}(x_t, t, G(x_t,t), T)) ],\label{eq:cadv_D} \\
    \mathcal{L}_\mathrm{adv'}^G = \mathbb{E}_{x,z,t} [ f(&D_\mathrm{jvp}(x_t, t, G(x_t,t), T),
    D_\mathrm{jvp}(x_t, t, \bar{v}_t, T)) ],\label{eq:cadv_G}
\end{align}
where we adopt a bounded contrastive function, similar to prior work~\cite{mao2017least}:
\begin{equation}
    f(a, b) = (a - 1)^2 + (b + 1)^2.
    \label{eq:bounded_contrastive_function}
\end{equation}
The gradients with respect to the model parameters are backpropagated through the JVP.

\input{fig/dis}

\cref{fig:dis} visualizes the intuition and training dynamics of our method. Intuitively, our discriminator $D$ learns a scalar potential whose directional derivative distinguishes real and fake flows. $D$ learns to assign higher potential to more realistic directions, and $G$ is optimized toward the direction that maximizes $D$'s potential. Training reaches equilibrium when $G$ learns the ground-truth flow and $D$ outputs flat potentials everywhere.

Since the objectives in \cref{eq:cadv_D,eq:cadv_G} only penalize the derivative while the absolute value of $D$ is free to drift, we also include a centering penalty to keep the absolute value of $D$ centered around zero:
\begin{equation}
    \mathcal{L}_\mathrm{cp'}^D = \mathbb{E}_{x,z,t} \left[ D(x_t, t)^2 \right].
\end{equation}

When training on high-dimensional flows where $n>1$, the discriminator, which projects an $n$-dimensional input to a scalar value, creates ambiguity because multiple $v_t\in\mathbb{R}^n$ can yield the same value. $G$ may learn to exploit the null space, and $D$ is then updated to counter this behavior. However, this causes slow convergence. A regularizer can be added to encourage $G$ to pick the minimum-norm solution, which is related to optimal transport regularization. The continuous-time optimal transport regularization on $G$ is equivalent to its discrete counterpart in \cref{eq:ot_discrete} in the limit of $|t-s|\rightarrow 0$:
\begin{equation}
    \mathcal{L}_\mathrm{ot'}^G = \mathbb{E}_{x,z,t} \left[ \frac{1}{n} \|G(x_t, t)\|^2_2 \right].
\end{equation}

Extending adversarial training to continuous time also mitigates the gradient vanishing problem (\cref{app:vanishing-gradient}). Empirically, we find that CAFMs can be trained without gradient penalties in our experiments. We also find it beneficial to train $D$ toward optimality by updating it for $N$ steps per update of $G$.

The final objectives for CAFMs resemble the discrete-time counterparts, except for the removal of the gradient penalties:
\begin{align}
    \mathcal{L}_\mathrm{CAFM}^D &= \mathcal{L}_\mathrm{adv'}^D + \lambda_\mathrm{cp}\mathcal{L}_\mathrm{cp'}^D, \\
    \mathcal{L}_\mathrm{CAFM}^G &= \mathcal{L}_\mathrm{adv'}^G + \lambda_\mathrm{ot}\mathcal{L}_\mathrm{ot'}^G.
\end{align}
We follow AFMs to gradually reduce $\lambda_\mathrm{ot}$ when training from scratch. For post-training existing flow-matching models, we set $\lambda_\mathrm{ot}=0$ to completely eliminate the bias of the Euclidean norm. The centering penalty is set to $\lambda_\mathrm{cp}=0.001$. We provide additional proofs and discussions in \cref{app:jvp-design} of the appendix.

\newpage
\subsection{Practical and Efficient Implementation}

JVP can be efficiently computed with forward-mode automatic differentiation. It computes both $D(x_t, t)$ and $D_\mathrm{jvp}(x_t, t, \dot{x}_t, \dot{t})$ in a single forward pass, allowing us to derive the adversarial loss $\mathcal{L}_\mathrm{adv'}^{D}$ and the centering penalty $\mathcal{L}_\mathrm{cp'}^D$ together efficiently. Additionally, we use vectorizing map (vmap) to efficiently compute multiple tangents at the same primal when updating $D$. A concise PyTorch implementation is provided in \cref{algo:cafm}. For larger-scale training, JVP and vmap are compatible with PyTorch's DDP~\cite{li2020pytorch}, FSDP~\cite{zhao2023pytorch}, and gradient checkpointing~\cite{chen2016training}. Implementation details are in \cref{app:implementation}.

\input{algo/cafm}

In terms of the network architecture, there are no restrictions on $G$ because it does not involve JVP computation and can use any architectures, same as in flow matching. For $D$, we find that switching LayerNorm~\cite{ba2016layer} to RMSNorm~\cite{zhang2019root} significantly improves training stability, consistent with the findings from previous research involving JVP computation~\cite{zhou2025terminal}. Unlike prior work, we do not find additional normalization on modulation necessary~\cite{zhou2025terminal,lu2024simplifying}. Our experiments show that CAFMs work well with standard transformers~\cite{vaswani2017attention} as both $G$ and $D$.

\subsection{Pre-training \vs Post-training}
\label{sec:pre-vs-post}

Although CAFMs can be trained from scratch, it is inherently less efficient than FMs due to the involvement of an extra discriminator network, the forward and backward computation of JVP, and the multiple steps of discriminator learning per generator update. Since both FMs and CAFMs learn the same probability flow and differ only in model generalization, it is much more efficient to pre-train models with FM objective and post-train with CAFM objective. Therefore, we primarily propose CAFMs for post-training, but still show that the objective can be used to train models from scratch, albeit less efficiently.

%% file: fig/dis.tex
\begin{figure}[h]
    \centering
    \small
    \begin{minipage}{0.8\columnwidth}
    \setlength{\tabcolsep}{2pt}
    \renewcommand{\arraystretch}{0.5}
    \newcolumntype{Y}{>{\centering\arraybackslash}X}
    \begin{tabularx}{\linewidth}{@{}cYYYY@{}}
        \raisebox{-0.5\height}{$G$} &
        \raisebox{-0.5\height}{\includegraphics[width=\linewidth]{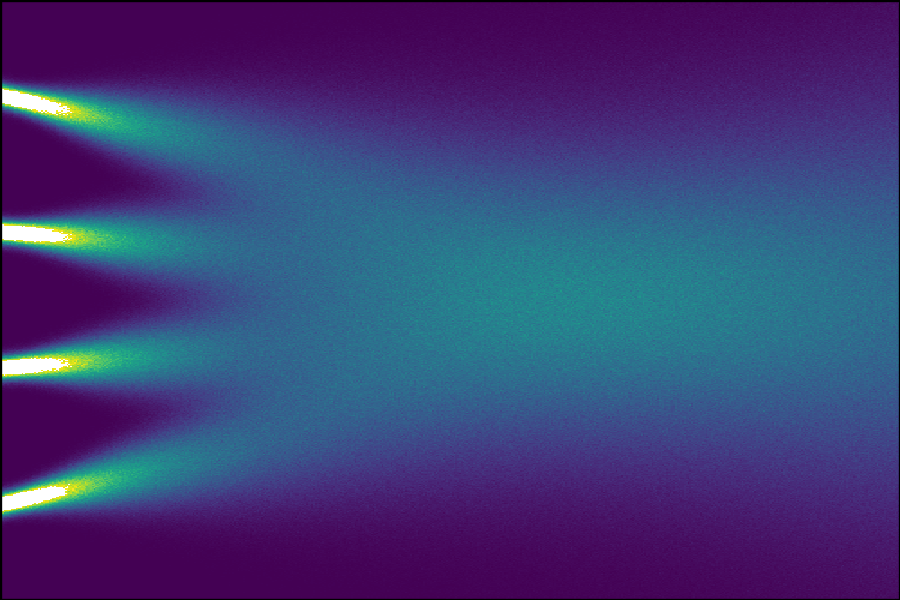}} &
        \raisebox{-0.5\height}{\includegraphics[width=\linewidth]{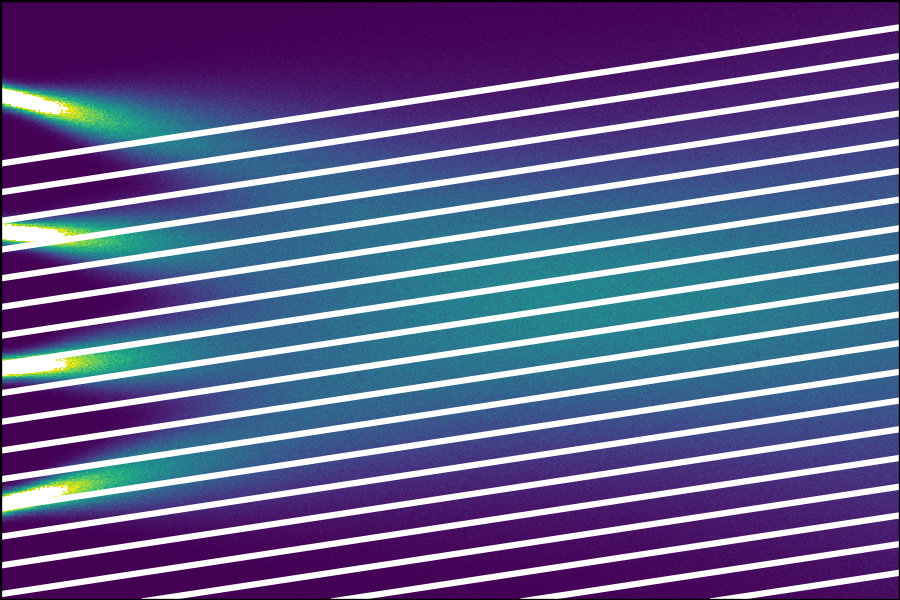}} &
        \raisebox{-0.5\height}{\includegraphics[width=\linewidth]{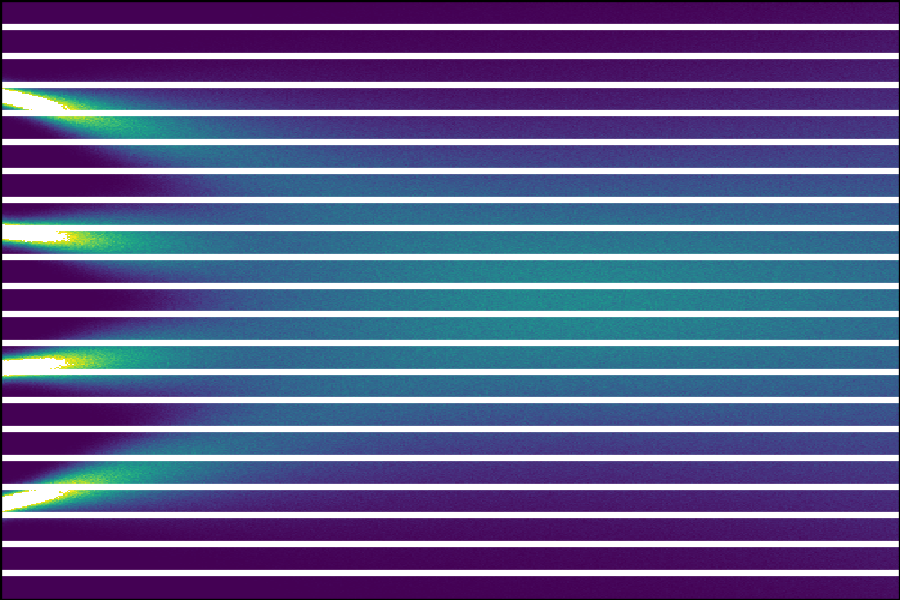}} &
        \raisebox{-0.5\height}{\includegraphics[width=\linewidth]{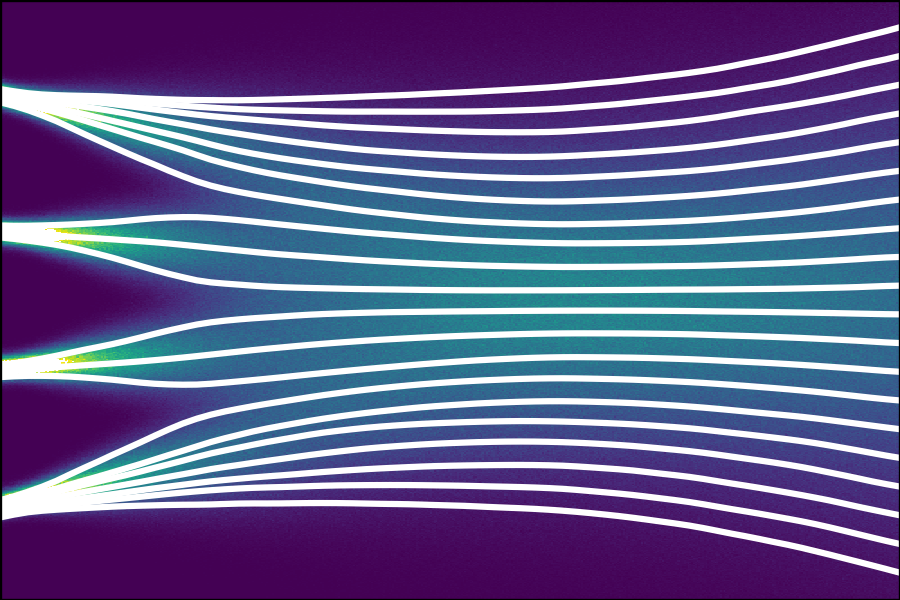}} \\
        \raisebox{-0.5\height}{$D$} &
        \raisebox{-0.5\height}{\includegraphics[width=\linewidth]{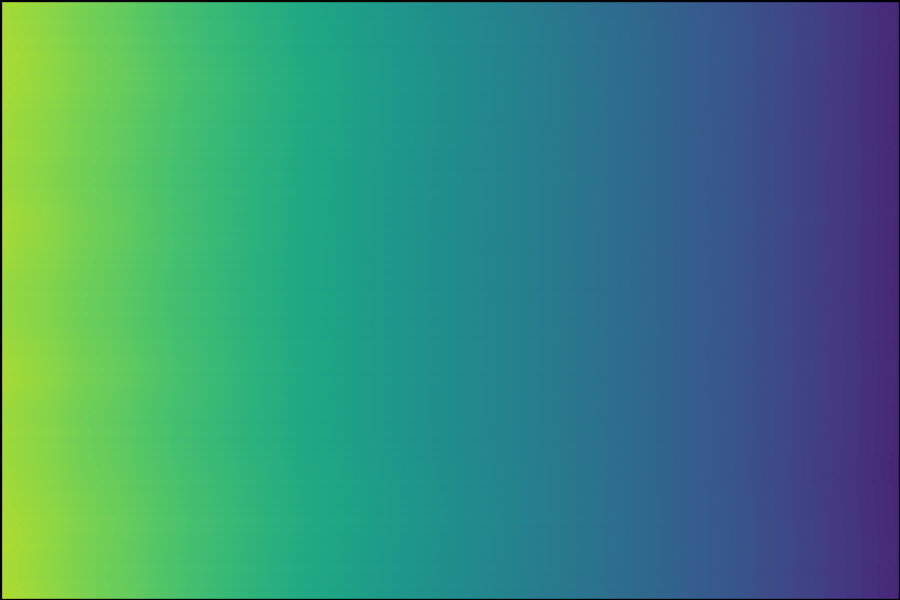}} & 
        \raisebox{-0.5\height}{\includegraphics[width=\linewidth]{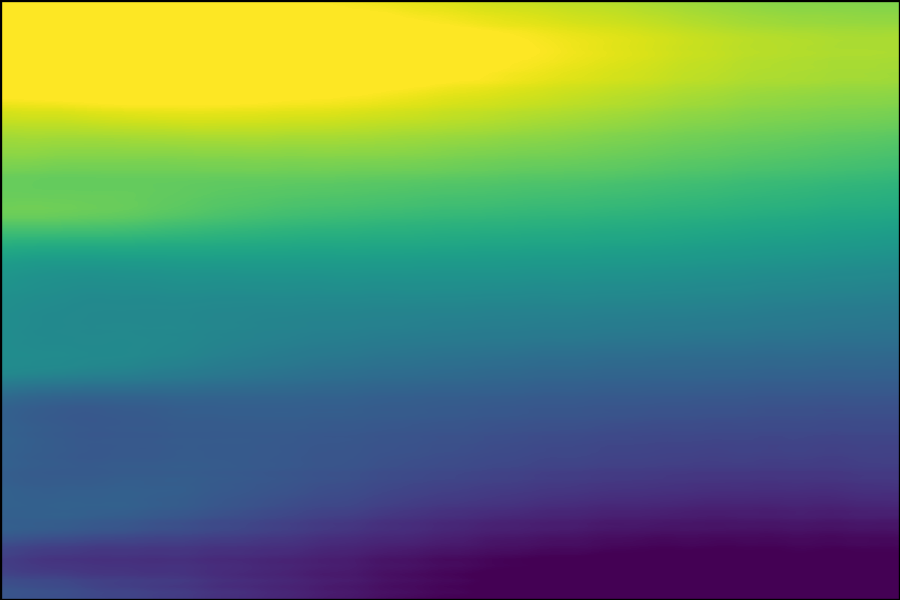}} &
        \raisebox{-0.5\height}{\includegraphics[width=\linewidth]{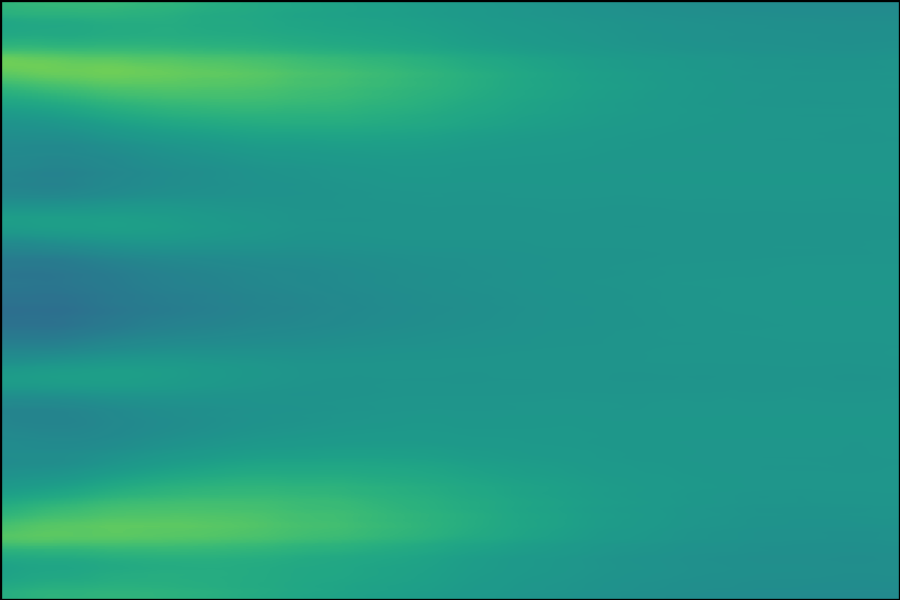}} &
        \raisebox{-0.5\height}{\includegraphics[width=\linewidth]{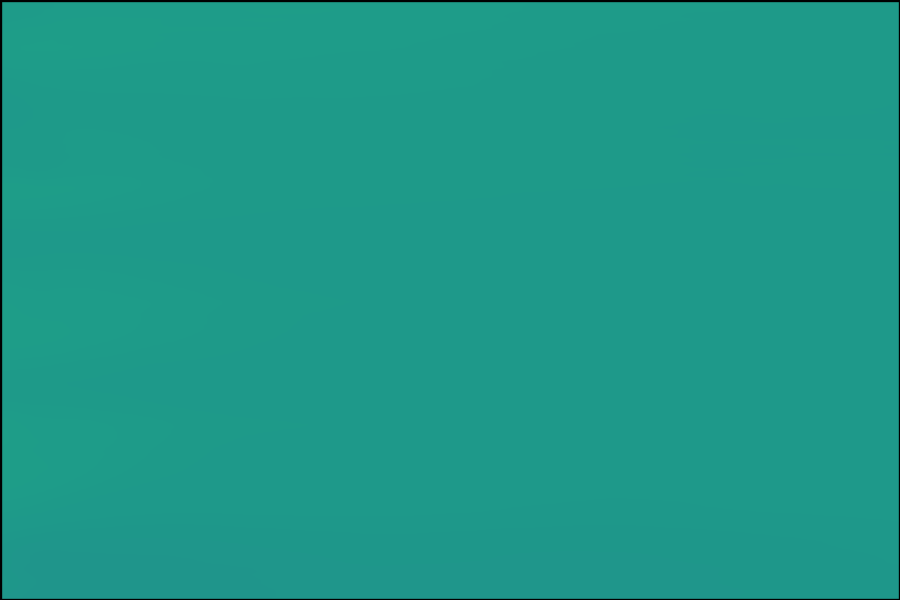}} \\\\
        & (a) & (b) & (c) & (d)
    \end{tabularx}
    \end{minipage}
    \caption{Visualization of the training dynamic. Top: learned $G(x_t, t)$ trajectories over the probability flow. Bottom: corresponding $-D(x_t, t)$ values at all $x_t$. $-D$ is taken for more intuitive visualization as the generation process runs backward in time. (a) shows if only training $D$ with $\bar{v}_t$ as positives without $G$ as negatives, $D$ degenerates to uniform gradient. (b,c) show how $D$ reacts to $G$ during training. (d) shows $D$ converges to 0 everywhere as $G$ converges to the ground-truth flow.}
    \label{fig:dis}
\end{figure}

%% file: algo/cafm.tex
\definecolor{codegreen}{rgb}{0,0.6,0}
\definecolor{codegray}{rgb}{0.5,0.5,0.5}

\captionsetup[algorithm]{font=small, labelfont=bf}

\lstset{
    language=Python,
    numbers=left,                
    numbersep=5pt,               
    numberstyle=\ttfamily\scriptsize,
    xleftmargin=2em,             
    basicstyle=\ttfamily\scriptsize,
    keywordstyle=\color{blue},
    commentstyle=\color{teal!70!black},
    stringstyle=\color{brown},
    showstringspaces=false,
}

\begin{algorithm}[h]
\caption{Continuous adversarial flow training} 
\label{algo:cafm}
\begin{lstlisting}
from functools import partial
from torch import mean, ones_like, stack, unbind
from torch.func import jvp, vmap

def step(G, D, x, z, t, c, mode, cp_scale, ot_scale):
    D.requires_grad_(mode == "dis")
    G.requires_grad_(mode == "gen")

    D = partial(D, condition=c)
    G = partial(G, condition=c)

    x_t = (1 - t) * x + t * z
    v_t = -x + z
    u_t = G(x_t, t)
    T = ones_like(t)

    if mode == "dis":
        o, do = vmap(lambda *tangents: jvp(D, (x_t, t), tangents))(
            stack([v_t, u_t]),
            stack([T, T])
        )
        dv, du = unbind(do)
        return (
            mean((dv - 1) ** 2) +
            mean((du + 1) ** 2) +
            mean(o ** 2) * cp_scale
        )
    else:
        _, du = jvp(D, (x_t, t), (u_t, T))
        return (
            mean((du - 1) ** 2) +
            mean(u_t ** 2) * ot_scale
        )
\end{lstlisting}
\end{algorithm}

%% file: sec/4.experiment.tex
\section{Experiment}

\subsection{ImageNet Generation Post-training}

On the class-conditional ImageNet~\cite{russakovsky2015imagenet} 256px generation task, we conduct experiments to post-train both latent-space flow-matching model SiT~\cite{ma2024sit} and pixel-space flow-matching model JiT~\cite{li2025back} with the CAFM objective and obtain significant performance gains in both guidance-free and guided settings, as measured by the Fréchet Inception Distance (FID)~\cite{heusel2017gans} and Inception Score (IS)~\cite{salimans2016improved}.

\begin{table}[h]
    \centering
    \begin{minipage}[t]{0.48\linewidth}
        \input{tab/sit}
    \end{minipage}
    \hfill
    \begin{minipage}[t]{0.48\linewidth}
        \input{tab/jit}
    \end{minipage}
\end{table}

\paragraph{SiT.} We adopt the officially pre-trained SiT-XL/2 model as the starting point for $G$ and keep the architecture completely unchanged. $D$ adopts the same architecture and weight initialization, except changing all LayerNorm to RMSNorm. We additionally follow the same modifications in AFM to prepend a learnable [CLS] token at input and add projection layers for the discriminator logit output. We use the same batch size of 256 as the original SiT. We set the learning rate to 1e-5 for both $G$ and $D$. We use Adam~\cite{kingma2014adam} optimizer with $\beta=(0,0.95)$. For the first 2 epochs, we freeze $G$ and only update $D$ for it to adapt to the new architecture. Then, we set $N=16$ to update $D$ 16 times per $G$'s update. Epochs are measured as the combined number of images seen by both $G$ and $D$ throughout our experiments. We use an exponential moving average (EMA) with a short decay of 0.99 on $G$. We set $\lambda_\mathrm{ot}=0$ for post-training. We use the exact inference and evaluation code provided by SiT, and use the Euler-Maruyama SDE sampler with 250 integration steps to match SiT's best setting. \Cref{tab:sit} shows that CAFM post-training significantly improves the FID from 8.26 to 3.63 in the guidance-free setting, and also improves the best FID from 2.06 to 1.53 in the guided setting under just 10 epochs of finetuning. For classifier-free guidance (CFG)~\cite{ho2021classifierfree}, the sweep finds that CAFMs achieve the best FID using CFG 1.3, which is lower than the original SiT at CFG 1.5. CAFMs improve generation at almost every swept CFG level. We also run a controlled trial by using the FM objective. It does not yield benefits compared to the SiT baseline, and the FID difference can be within the error margin of random evaluation sampling. This proves that the gains are a result of the CAFM objective. More ablation studies are provided in \cref{app:imagenet-post-train}.

\paragraph{JiT.} We adopt the officially pre-trained JiT-H/16 model as the starting point for $G$ and keep the architecture completely unchanged. $D$ adopts the same architecture and weight initialization as $G$. Because JiT already uses RMSNorm and has in-context class tokens, we simply take the first class token and add projection layers for the discriminator output. We convert the $x$-prediction result by $G$ to $v$ before giving it to $D$. We use the same batch size of 1024 as the original JiT. Follow SiT, we also set the learning rate to 1e-5, $\beta=(0,0.95)$, $N=16$, and EMA decay to 0.99. $D$ is warmed up for the first 4 epochs. We use the exact inference and evaluation code provided by JiT. We follow JiT to use the Heun ODE sampler with 50 steps. \Cref{tab:jit} shows that CAFMs also significantly improve the FID from 7.17 to 3.57 in the guidance-free setting, and improves the best FID from 1.86 to 1.80 in the guided setting. CAFMs improve performance in almost all swept CFG levels. We find that the FM control trial produces worse results than JiT's official checkpoint despite our best effort to reproduce. Regardless, it is sufficient to prove that the gains are originated from the CAFM objective.

\begin{table}[h]
    \centering
    \begin{minipage}[t]{0.48\linewidth}
        \input{tab/imagenet_latent}
    \end{minipage}
    \hfill
    \begin{minipage}[t]{0.48\linewidth}
        \input{tab/imagenet_pixel}
    \end{minipage}
\end{table}

\paragraph{Comparisons to the state of the arts.} \Cref{tab:imagenet_latent_space,tab:imagenet_pixel_space} compare our results to other methods. Under SD-VAE~\cite{rombach2022high} latent space and without using DINOv2~\cite{oquab2023dinov2}, our method achieves the best performance in both guided and guidance-free settings among the models compared. Since all works use the DiT architecture and similar training settings, it is easier to attribute the gain to our method. In pixel space, settings vary significantly, making it harder to pinpoint contributions by the method from architectural improvements. We suspect that SiD~\cite{hoogeboom2023simple} achieves better FID in the guidance-free setting because its 2B-parameter model can overfit the training data better. Overall, our method also achieves very competitive performance in the pixel space.

\subsection{Text-to-Image Generation Post-training}

\paragraph{Setup.} We experiment post-training with a text-to-image generation model, Z-Image~\cite{cai2025z}, using our CAFM objective. We first train the model with FM on our data for 10K iterations, then switch to CAFM for 20K, while keeping the FM trial running to match the iterations. Following common finetuning practice, FM training uses a batch size of 1024, AdamW optimizer~\cite{loshchilov2017decoupled} with a learning rate of 5e-5, $\beta=(0.9, 0.95)$, weight decay of 0.01, and an EMA decay of 0.999. Then, we switch to the CAFM objective while matching most of the FM settings. We lower $D$'s learning rate to 3e-5 to avoid loss spiking while keeping $G$ at 5e-5. We set $\beta=(0,0.95)$. We lower the EMA decay to 0.99 to account for $N=16$ discriminator update steps.

\paragraph{Evaluation.} Our models are evaluated by both GenEval~\cite{ghosh2023geneval} in \cref{tab:geneval} and by DPG-Bench~\cite{hu2024ella} in \cref{tab:dpg}. In GenEval, we use the prompt expansion (PE) provided by prior work~\cite{deng2025emerging,ai2026bitdance}. In both benchmarks, CAFM post-training significantly improves the performance of guidance-free generation, while also improving the guided setting.

\paragraph{Limitation.} Although CAFMs empirically achieve better performance in guidance-free generation, there is no guarantee that the models generalize to the true underlying data distribution, especially in the low-density regions containing outliers. Guidance can be used orthogonally to our method to improve benchmark scores as a low-temperature sampling technique.

\input{tab/geneval}
\input{tab/dpg}

\input{fig/t2i}

\clearpage
\subsection{ImageNet Generation Trained from Scratch}

Although CAFM is proposed primarily as a post-training method, for completeness, we also experiment with training from scratch using the CAFM objective on ImageNet 256px. We use the SiT-B/2 architecture with a batch size of 256, an optimizer learning rate of 1e-4 with $\beta=(0,0.95)$ for both $G$ and $D$, and an EMA decay of 0.9999, matching the original pre-training settings of SiT. The hyperparameters of the discriminator updates per generator update $N$ and the optimal transport loss weighting $\lambda_\mathrm{ot}$ are searched during training for the fastest convergence. In \cref{fig:pretrain_fid_curve}, we show that the CAFM objective can be used to train from scratch, but converges more slowly than FM under the same epochs, fitting our expectation in \cref{sec:pre-vs-post}.

\paragraph{Ablation studies on the hyperparameters.} Overall, we find that $\lambda_\mathrm{ot}$ should decrease over training and $N$ should increase over training for the best performance. In \cref{fig:pretrain_ablation_initial_N}, we first fix $\lambda_\mathrm{ot}=1$ and compare the hyperparameter of $N$. We find $N=4$ outperforms $N=1$ after the first 50 epochs, so we use $N=4$. Then in \cref{fig:pretrain_ablation_initial_ot}, we search for $\lambda_\mathrm{ot}$ and find that $\lambda_\mathrm{ot}=4$ converges the fastest in the first 50 epochs. Therefore, we use $N=4,\lambda_\mathrm{ot}=4$ as the initial settings. In \cref{fig:pretrain_ablation_reduce_ot}, we experiment with a lower $\lambda_\mathrm{ot}$ to 1 since 160 epochs and see further FID improvement, while $\lambda_\mathrm{ot}=4$ eventually plateaus. This shows the importance of decreasing $\lambda_\mathrm{ot}$ over training, concurring with the findings of AFM. In \cref{fig:pretrain_ablation_increase_N}, we further increase $N$ to 8 at 700 epochs and see faster convergence at the later stage. Note that we have swept other changes during training, including decreasing the learning rate, further decreasing $\lambda_\mathrm{ot}$, and further increasing $N$ to match the settings of post-training, but they yield worse performance. We suspect these changes are too early within our 1000-epoch pre-training budget. We leave further explorations on pre-training to future work.

\begin{figure}[h]
    \begin{minipage}[t]{0.47\linewidth}
        \input{fig/pretrain}
    \end{minipage}
    \hfill
    \begin{minipage}[t]{0.51\linewidth}
        \input{fig/pretrain_ablate}
    \end{minipage}
\end{figure}

%% file: tab/sit.tex
    \centering
    \scriptsize
    \caption{
    SiT-XL/2 on ImageNet 256px.\\
    \vspace{-5pt} \\
    \scriptsize{Full comparisons are in \cref{tab:sit_full} of the appendix.}
    }
    \label{tab:sit}
    \setlength{\tabcolsep}{3pt}
    \begin{tabular}{l|l|l|rr}
        \toprule
        CFG & Method & Epoch & FID$\downarrow$ & IS$\uparrow$ \\
        \midrule
        None & SiT & 1400 & 8.26 & 131.65 \\
        & SiT+FM & 1400+10 & 8.64 & 131.91 \\
        & SiT+CAFM & 1400+10 & \textcolor{red}{\textbf{3.63}} & \textbf{178.08} \\
        \midrule
        1.1 & SiT & 1400 & 5.55 & 161.77 \\
        & SiT+CAFM & 1400+10 & \textbf{2.27} & \textbf{212.06} \\
        \midrule
        1.2 & SiT & 1400 & 3.65 & 190.57 \\
        & SiT+CAFM & 1400+10 & \textbf{1.66} & \textbf{238.44} \\
        \midrule
        1.3 & SiT & 1400 & 2.57 & 220.52 \\
        & SiT+CAFM & 1400+10 & \textcolor{red}{\textbf{1.53}} & \textbf{263.52} \\
        \midrule
        1.4 & SiT & 1400 & 2.07 & 248.31 \\
        & SiT+CAFM & 1400+10 & \textbf{1.66} & \textbf{283.59} \\
        \midrule
        1.5 & SiT & 1400 & 2.06 & 277.50 \\
        & SiT+CAFM & 1400+10 & \textbf{1.97} & \textbf{301.91} \\
        \midrule
        1.6 & SiT & 1400 & \textbf{2.25} & 293.72 \\
        & SiT+CAFM & 1400+10 & 2.37 & \textbf{316.78} \\
        \bottomrule
    \end{tabular}

%% file: tab/jit.tex
    \centering
    \scriptsize
    \caption{
    JiT-H/16 on ImageNet 256px.\\
    \vspace{-5pt} \\
    \scriptsize{Full comparisons are in \cref{tab:jit_full} of the appendix.}
    }
    \label{tab:jit}
    \setlength{\tabcolsep}{3pt}
    \begin{tabular}{l|l|l|rr}
        \toprule
        CFG & Method & Epoch & FID$\downarrow$ & IS$\uparrow$ \\
        \midrule
        None & JiT & 600 & 7.17 & 151.54 \\
        & JiT+FM & 600+10 & 9.30 & 139.00 \\
        & JiT+CAFM & 600+10 & \textcolor{red}{\textbf{3.57}} & \textbf{198.08} \\
        \midrule
        1.4 & JiT & 600 & 3.24 & 219.52 \\
        & JiT+CAFM & 600+10 & \textbf{2.01} & \textbf{258.46} \\
        \midrule
        1.6 & JiT & 600 & 2.49 & 244.61 \\
        & JiT+CAFM & 600+10 & \textbf{1.84} & \textbf{275.96} \\
        \midrule
        1.8 & JiT & 600 & 2.12 & 265.43 \\
        & JiT+CAFM & 600+10 & \textcolor{red}{\textbf{1.80}} & \textbf{290.71} \\
        \midrule
        2.0 & JiT & 600 & 1.96 & 281.38 \\
        & JiT+CAFM & 600+10 & \textbf{1.83} & \textbf{301.96} \\
        \midrule
        2.2 & JiT & 600 & \textbf{1.86} & 303.40 \\
        & JiT+CAFM & 600+10 & 1.88 & \textbf{310.54} \\
        \midrule
        2.4 & JiT & 600 & 2.19 & 310.20 \\
        & JiT+CAFM & 600+10 & \textbf{1.95} & \textbf{319.23} \\
        \bottomrule
    \end{tabular}

%% file: tab/imagenet_latent.tex
    \centering
    \caption{SD-VAE latent-space continuous flow models on ImageNet 256px. Methods in gray use DINOv2~\cite{oquab2023dinov2}.}
    \label{tab:imagenet_latent_space}
    \scriptsize
    \setlength{\tabcolsep}{3pt}
    \begin{tabular}{l|l|r|rr}
        \toprule
        Guided & Method & Param & FID$\downarrow$ \\
        \midrule
        No & DiT-XL/2~\cite{peebles2023scalable} & 675M & 9.62 \\
        & SiT-XL/2~\cite{ma2024sit} & 675M & 8.26 \\
        & SiT-XL/2+Disperse~\cite{wang2025diffuse} & 675M & 7.43 \\
        & \textcolor[gray]{0.7}{DDT-XL~\cite{wang2025ddt}} & 675M & \textcolor[gray]{0.7}{6.27} \\
        & \textcolor[gray]{0.7}{SiT-XL/2+REPA~\cite{yu2024representation}} & 675M & \textcolor[gray]{0.7}{5.90} \\
        & SiT-XL/2+CAFM & 675M & \textcolor{red}{\textbf{3.63}} \\
        \midrule
        Yes & DiT-XL/2~\cite{peebles2023scalable} & 675M & 2.27 \\
        & SiT-XL/2~\cite{ma2024sit} & 675M & 2.06 \\
        & SiT-XL/2+Disperse~\cite{wang2025diffuse} & 675M & 1.97 \\
        & SiT-XL/2+CAFM & 675M & \textcolor{red}{\textbf{1.53}} \\
        & \textcolor[gray]{0.7}{SiT-XL/2+REPA~\cite{yu2024representation}} & 675M & \textcolor[gray]{0.7}{1.42} \\
        & \textcolor[gray]{0.7}{DDT-XL~\cite{wang2025ddt}} & 675M & \textcolor[gray]{0.7}{1.26} \\
        \bottomrule
    \end{tabular}

%% file: tab/imagenet_pixel.tex
    \centering
    \caption{Pixel-space continuous flow models on ImageNet 256px. Please consider that architectures and settings vary.}
    \label{tab:imagenet_pixel_space}
    \scriptsize
    \setlength{\tabcolsep}{3pt}
    \begin{tabular}{l|l|r|rr}
        \toprule
        Guided & Method & Param & FID$\downarrow$ \\
        \midrule
        No & ADM~\cite{dhariwal2021diffusion} & 554M & 10.94 \\
        & JiT-H/16~\cite{li2025back} & 956M & 7.17 \\
        & JiT-H/16+CAFM & 956M & \textcolor{red}{\textbf{3.57}} \\
        & SiD~\cite{hoogeboom2023simple} & 2B & \textbf{2.77} \\
        \midrule
        Yes & ADM-G~\cite{dhariwal2021diffusion} & 554M & 4.59 \\
        & SiD~\cite{hoogeboom2023simple} & 2B & 2.44 \\
        & PixNerd-XL/16~\cite{wang2025pixnerd} & 700M & 2.15 \\
        & PixelFlow-XL/4~\cite{chen2025pixelflow} & 677M & 1.98 \\
        & JiT-H/16~\cite{li2025back} & 956M & 1.86 \\
        & JiT-G/16~\cite{li2025back} & 2B & 1.82 \\
        & JiT-H/16+CAFM & 956M & \textcolor{red}{\textbf{1.80}} \\
        & SiD2~\cite{hoogeboom2025simpler} & 653M & \textbf{1.38} \\
        \bottomrule
    \end{tabular}

%% file: tab/geneval.tex
\begin{table}[h]
    \centering
    \caption{GenEval~\cite{ghosh2023geneval} on 512px text-to-image generation.}
    \label{tab:geneval}
    \scriptsize
    \setlength{\tabcolsep}{2pt}
    \begin{tabular}{l|cc|cccccc|c}
        \toprule
        Method & PE & CFG & Single Obj. & Two Obj. & Color Attr. & Position & 
        Counting & Colors. & Overall \\
        \midrule
        FM & \multirow{2}{*}{No} & \multirow{2}{*}{No} & 0.72 & 0.23 & 0.11 & 0.09 & 0.25 & 0.59 & 0.33\phantom{*} \\
        CAFM & & & 0.85 & 0.42 & 0.17 & 0.16 & 0.41 & 0.61 & 0.44\phantom{*} \\
        \midrule
        FM & \multirow{2}{*}{Yes} & \multirow{2}{*}{No} & 0.95 & 0.66 & 0.35 & 0.40 & 0.42 & 0.81 & 0.60\phantom{*} \\
        CAFM & & & 0.99 & 0.83 & 0.50 & 0.52 & 0.57 & 0.86 & 0.71\phantom{*} \\ 
        \midrule
        FM & \multirow{2}{*}{Yes} & \multirow{2}{*}{Yes} & 0.99 & 0.89 & 0.62 & 0.69 & 0.77 & 0.89 & 0.81\phantom{*} \\
        CAFM & & & 0.99 & 0.92 & 0.71 & 0.71 & 0.81 & 0.94 & 0.85\phantom{*} \\
        \bottomrule         
    \end{tabular}
    
\end{table}

%% file: tab/dpg.tex
\begin{table}[h]
    \centering
    \caption{DPG-Bench~\cite{hu2024ella} on 512px text-to-image generation.}
    \label{tab:dpg}
    \scriptsize
    \setlength{\tabcolsep}{7.5pt}
    \begin{tabular}{l|c|ccccc|c}
        \toprule
        Method & CFG & Global & Entity & Attribute & Relation & Other & Overall \\
        \midrule
        FM & \multirow{2}{*}{No} & 81.34 & 82.96 & 81.71 & 83.17 & 85.07 & 72.25 \\
        CAFM & & 87.82 & 86.65 & 86.33 & 86.49 & 84.85 & 77.21 \\
        \midrule
        FM & \multirow{2}{*}{Yes} & 90.34 & 90.56 & 88.98 & 88.17 & 90.71 & 83.67 \\
        CAFM & & 89.55 & 89.83 & 89.99 & 91.20 & 91.88 & 85.21 \\
        \bottomrule
    \end{tabular}
\end{table}

%% file: fig/t2i.tex
\begin{figure}[h!]
    \centering
    \setlength{\tabcolsep}{0pt}
    \captionsetup{justification=raggedright,singlelinecheck=false}
    \begin{subfigure}[t]{\linewidth}

        \begin{tabularx}{\linewidth}{XX@{\hspace{4pt}}XX}
            \includegraphics[width=\linewidth]{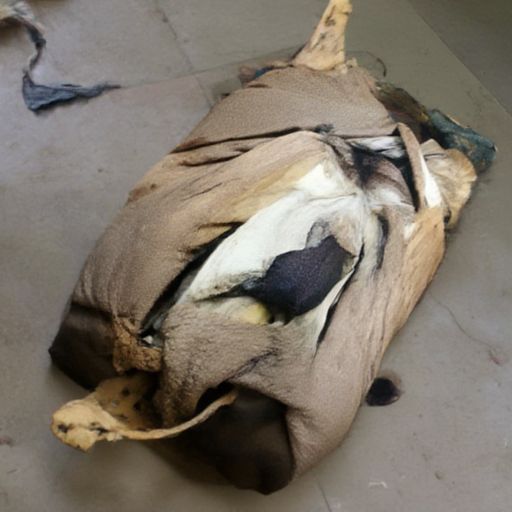} &
            \includegraphics[width=\linewidth]{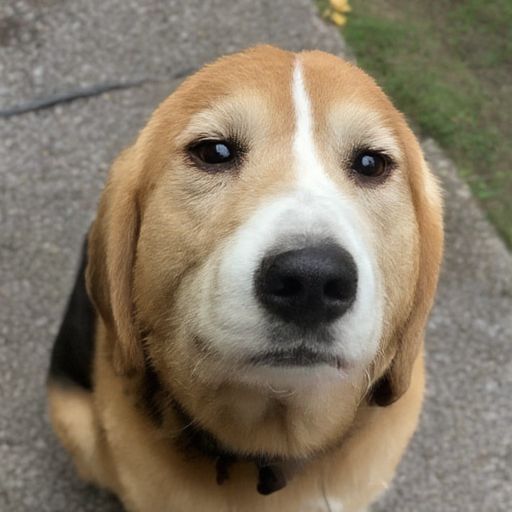} &
            \includegraphics[width=\linewidth]{img/t2i/dog_fm_4.jpg} &
            \includegraphics[width=\linewidth]{img/t2i/dog_ca_4.jpg} \\
        \end{tabularx}
        \vspace{-6pt}
        \caption{A photo of a dog.}
    \end{subfigure}

    \begin{subfigure}[t]{\linewidth}
        \begin{tabularx}{\linewidth}{XX@{\hspace{4pt}}XX@{\hspace{4pt}}XX}
            \includegraphics[width=\linewidth]{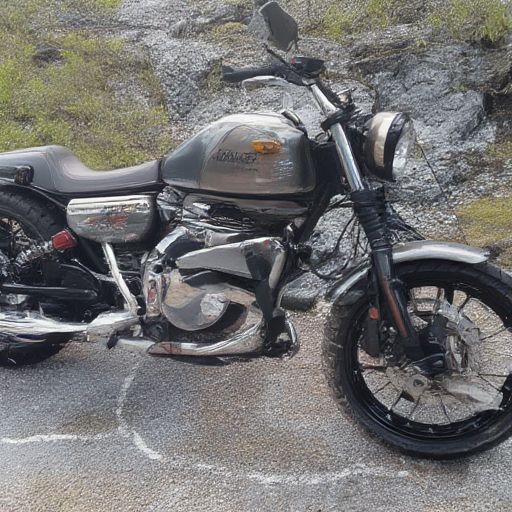} &
            \includegraphics[width=\linewidth]{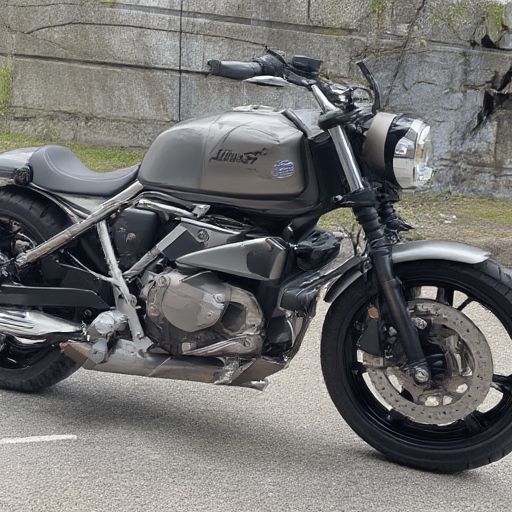} &
            \includegraphics[width=\linewidth]{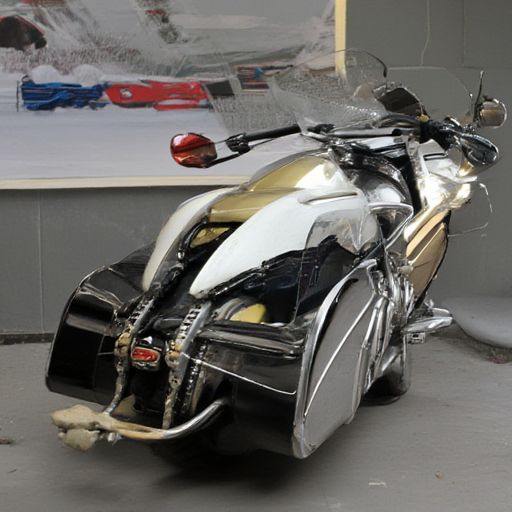} &
            \includegraphics[width=\linewidth]{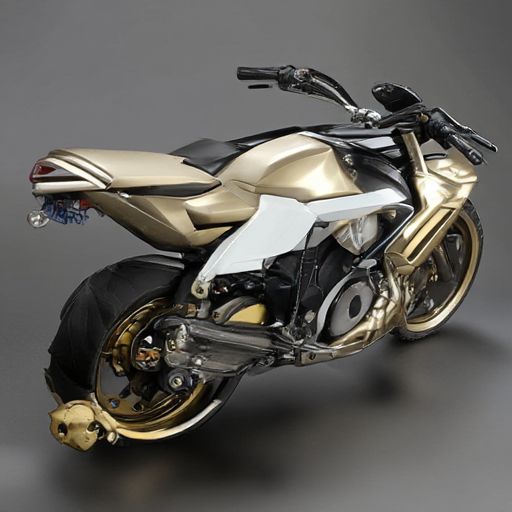} &
            \includegraphics[width=\linewidth]{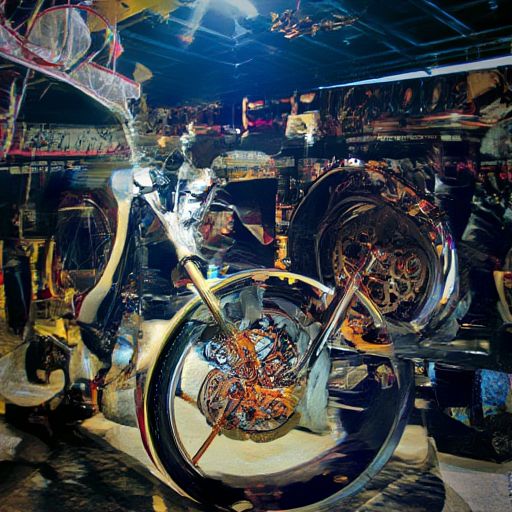} &
            \includegraphics[width=\linewidth]{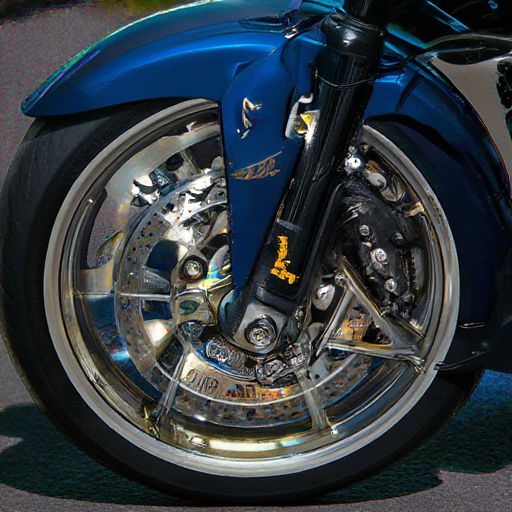} \\
        \end{tabularx}
        \vspace{-6pt}
        \caption{A photo of a motocycle.}
    \end{subfigure}

    \begin{subfigure}[t]{\linewidth}
        \begin{tabularx}{\linewidth}{XX@{\hspace{4pt}}XX@{\hspace{4pt}}XX}
            \includegraphics[width=\linewidth]{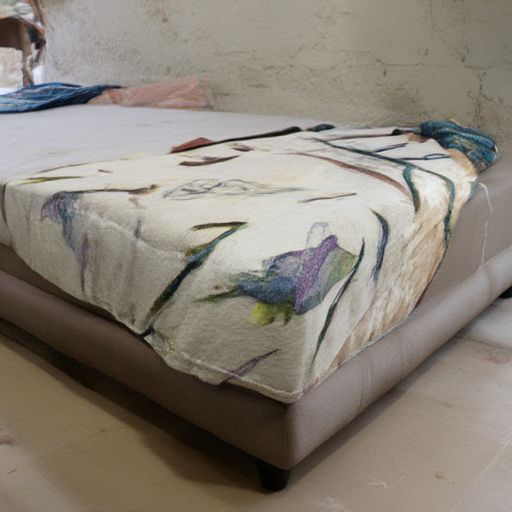} &
            \includegraphics[width=\linewidth]{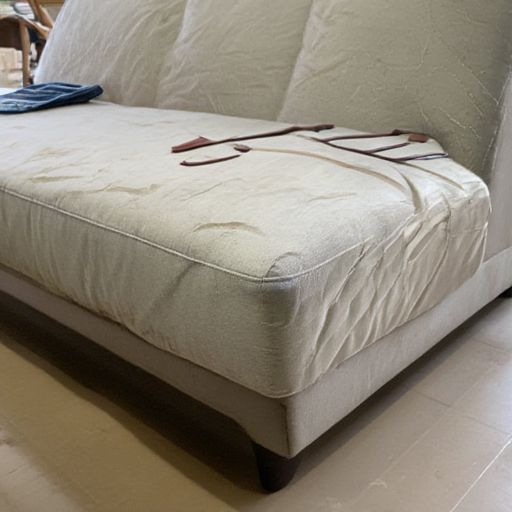} &
            \includegraphics[width=\linewidth]{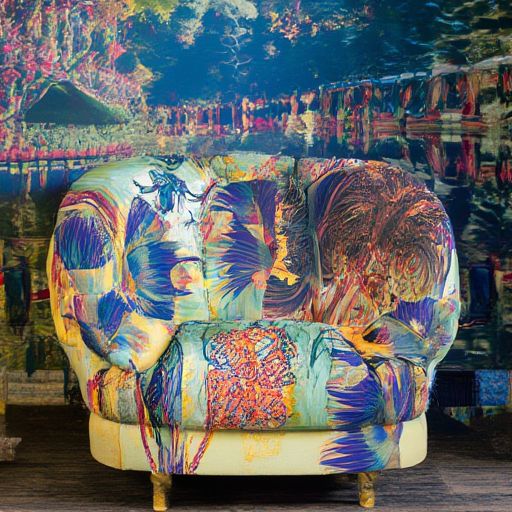} &
            \includegraphics[width=\linewidth]{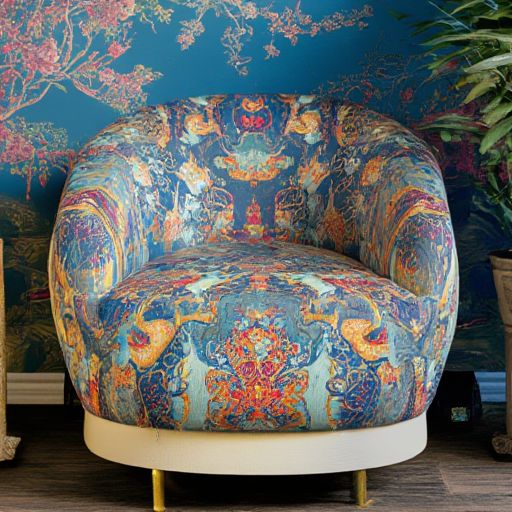} &
            \includegraphics[width=\linewidth]{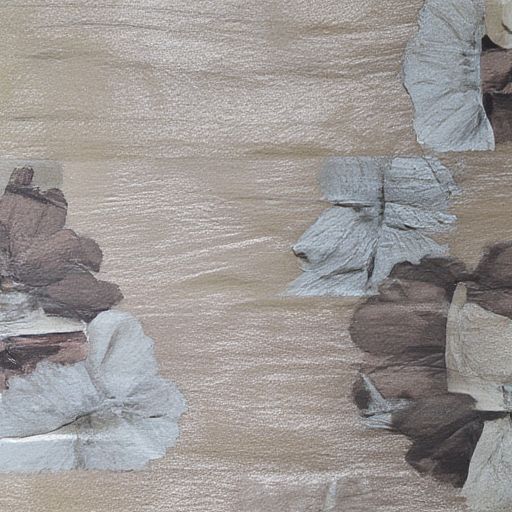} &
            \includegraphics[width=\linewidth]{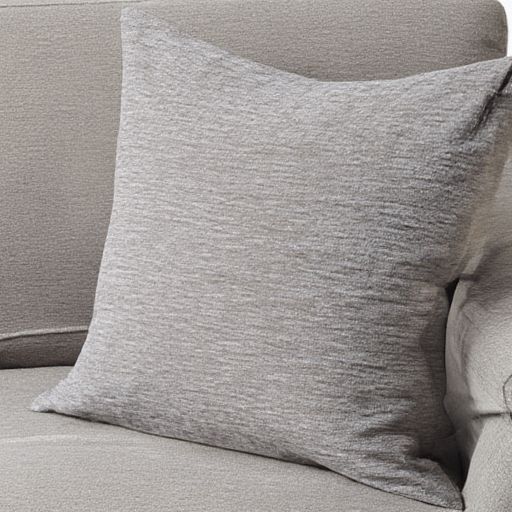}\\
        \end{tabularx}
        \vspace{-6pt}
        \caption{A photo of a couch.}
    \end{subfigure}

    \begin{subfigure}[t]{\linewidth}
        \begin{tabularx}{\linewidth}{XX@{\hspace{4pt}}XX@{\hspace{4pt}}XX}
            \includegraphics[width=\linewidth]{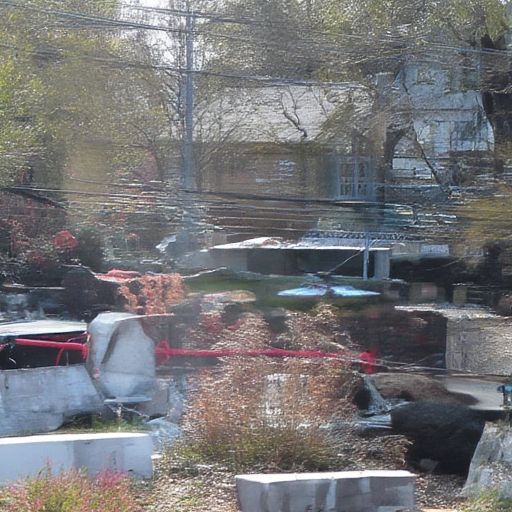} &
            \includegraphics[width=\linewidth]{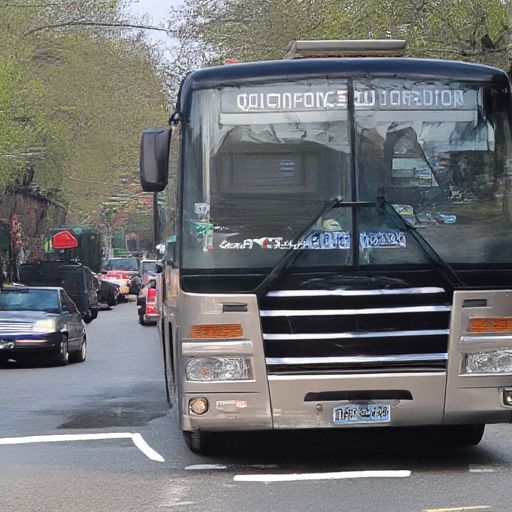} &
            \includegraphics[width=\linewidth]{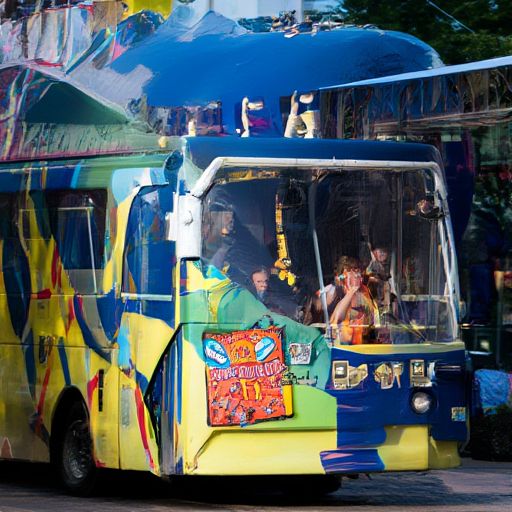} &
            \includegraphics[width=\linewidth]{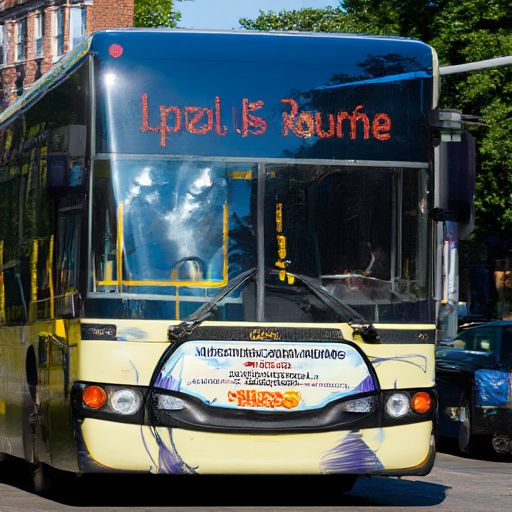} &
            \includegraphics[width=\linewidth]{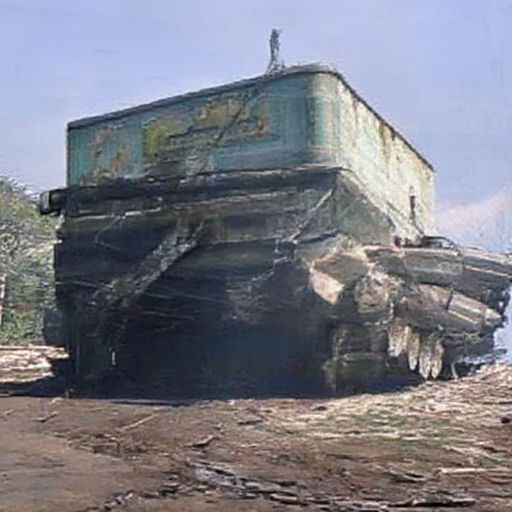} &
            \includegraphics[width=\linewidth]{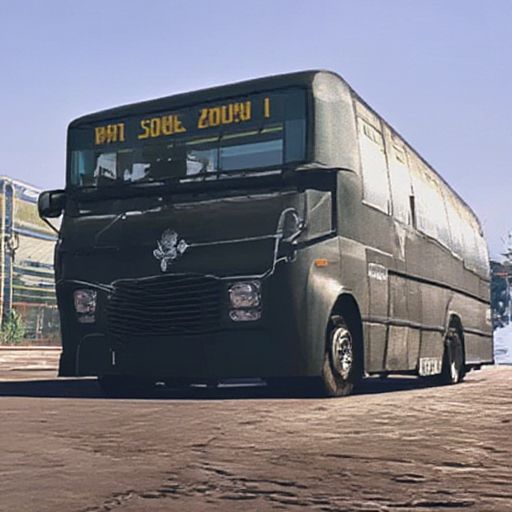}\\
        \end{tabularx}
        \vspace{-6pt}
        \caption{A photo of a bus.}
    \end{subfigure}

    \vspace{-5pt}
    \caption{
    Curated text-to-image samples on GenEval prompts. \\
    Without PE and CFG to show the most diverse range of samples. \\
    Left is FM. Right is CAFM. More visualizations are in \cref{fig:t2i_dpg} of the appendix.
    }
    \label{fig:placeholder}
\end{figure}

%% file: fig/pretrain.tex
    \vspace{0pt}
    \centering
    \includegraphics[width=\linewidth]{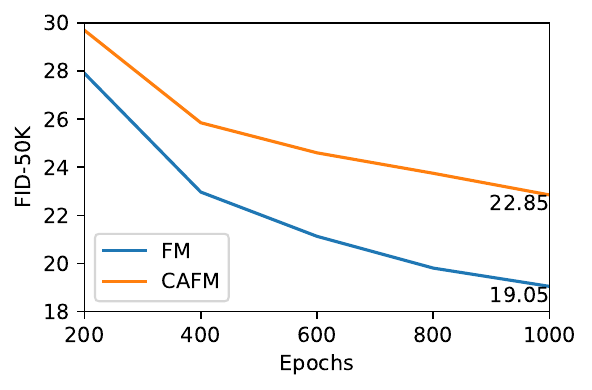}
    \caption{SiT-B/2 pre-training on ImageNet 256px generation. Although CAFM can be used for pre-training, it converges slower than FM under the same epochs, fitting our expectation that CAFM is more suitable for post-training.}
    \label{fig:pretrain_fid_curve}

%% file: fig/pretrain_ablate.tex
    \vspace{3pt}
    \centering
    \begin{subfigure}[t]{0.48\linewidth}
        \includegraphics[width=\linewidth]{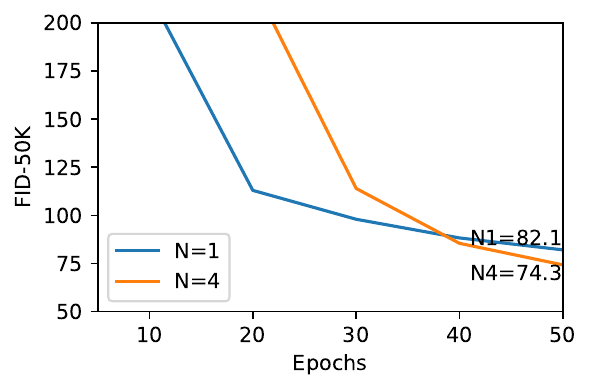}
        \caption{Initial $N$}
        \label{fig:pretrain_ablation_initial_N}
    \end{subfigure}
    \begin{subfigure}[t]{0.48\linewidth}
        \includegraphics[width=\linewidth]{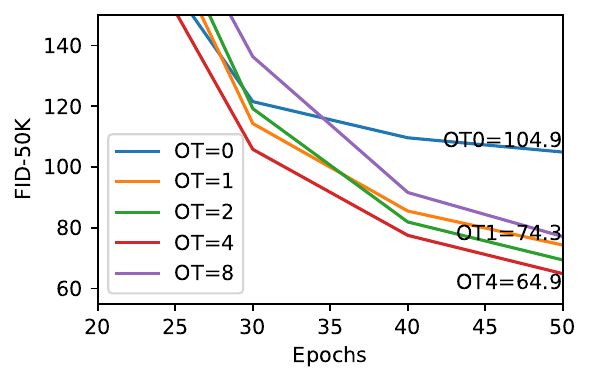}
        \caption{Initial $\lambda_\mathrm{ot}$}
        \label{fig:pretrain_ablation_initial_ot}
    \end{subfigure}
    \\ \vspace{7pt}
    \begin{subfigure}[t]{0.48\linewidth}
        \includegraphics[width=\linewidth]{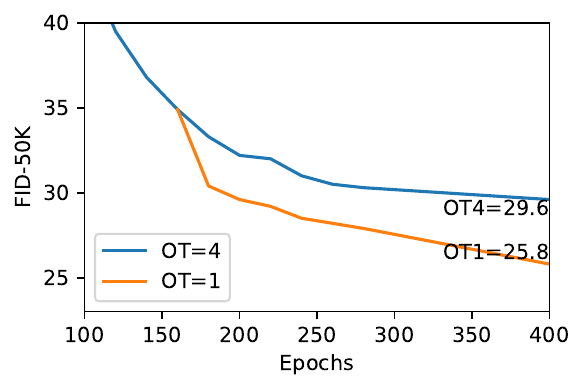}
        \caption{Reduce $\lambda_\mathrm{ot}$ at 160ep}
        \label{fig:pretrain_ablation_reduce_ot}
    \end{subfigure}
    \begin{subfigure}[t]{0.48\linewidth}
        \includegraphics[width=\linewidth]{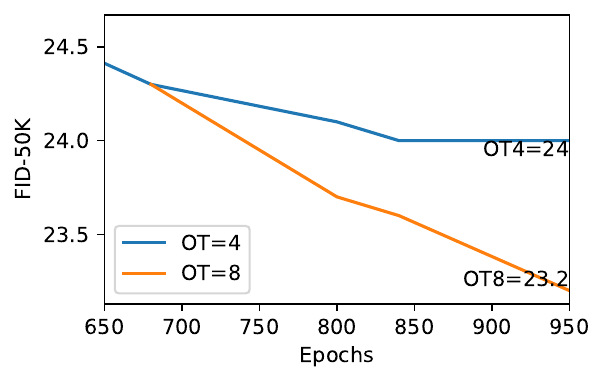}
        \caption{Increase $N$ at 700ep}
        \label{fig:pretrain_ablation_increase_N}
    \end{subfigure}
    \\
    \vspace{5pt}
    \caption{Ablation studies on the effect of different hyperparameters.}

%% file: sec/5.related.tex
\newpage

\section{Related Work}

\paragraph{Unifying Adversarial and Flow Modeling.} Adversarial training originates from generative adversarial networks (GANs)~\cite{goodfellow2014generative}. Recent work, adversarial flow models (AFMs), combines adversarial and discrete-time flow modeling. Our work on CAFMs is an extension of AFMs into continuous time.

\paragraph{Adversarial Post-Training.} Adversarial post-training of existing flow models has largely been researched as distillation methods to achieve few-step generation~\cite{lin2025diffusion,lin2024sdxl,lin2024animatediff,ren2024hyper,xu2024ufogen,yin2024improved,sauer2024adversarial,sauer2024fast,lin2025autoregressive,wang2025seedvr2,choudhury2025skipsr,kang2024distilling,wang2024phased}. Our work applies adversarial post-training on continuous-time flow models for inducing different model generalization instead.

\paragraph{Generalization Behavior.} Prior work~\cite{mathieu2020riemannian,de2022riemannian,chen2023flow} has explored lifting the flow models to custom manifolds, where the trajectories lie on the defined manifolds and hence also alter the model generalization. Our method still flows through the Euclidean space with the same ground-truth trajectories as standard flow matching but only induces different generalization through the loss objectives, related to prior perceptual loss research~\cite{lin2023diffusion}. Recent work has explored training on different latent spaces~\cite{rombach2022high,zheng2025diffusion,tong2026scaling,bfl2025representation}, which alters the space in which flow matching operates and implicitly changes generalization behavior. Our method is effective in both the generic latent space and the pixel space.

\paragraph{Guidance.} Guidance steers the sampling process of generative models toward a modified distribution. It can be derived from the gradient of an external classifier network~\cite{dhariwal2021diffusion,kim2023refining} or from an implicit classifier obtained from a pair of flow models through Bayes' rule~\cite{ho2021classifierfree,karras2024guiding,hu2023guided}. Guidance has effects similar to low-temperature sampling~\cite{xu2025temporal}, but we suspect that the improvement in sample quality also stems from the use of the explicit or implicit classifier that better captures the data manifold. Unlike guidance which can produce canonical and out-of-distribution samples~\cite{lin2024common}, our method converges to the ground-truth flow and remains faithful to the original distribution. Guidance can be applied orthogonally, and our experiments show that improving the base model also improves guided results.

\paragraph{Divergence Measures.} Flow matching, through its connection to score matching~\cite{song2021scorebased}, minimizes forward KL divergence. GANs can minimize different divergences~\cite{nowozin2016f}, which also influences generalization. We compare different objectives in \cref{tab:sit_f} of the appendix and leave further investigation to future work.

%% file: sec/6.conclusion.tex
\section{Conclusion}

We have introduced continuous adversarial flow models (CAFMs), a type of continuous-time flow model trained with the adversarial objective. We have empirically demonstrated that our objective can be efficiently used as a post-training method on flow-matching models and provides performance improvement on ImageNet generation and on text-conditional image generation. Our work offers exciting prospects for future research.

%% file: sec/7.acknowledgment.tex
\section*{Acknowledgment}

We thank Kunchang Li and Yuwei Guo for their valuable discussions and assistance.

%% file: sup/sup.tex
\section{Additional Results on ImageNet Post-training}
\label{app:imagenet-post-train}

\Cref{tab:sit_full} provides the full evaluation metrics of SiT-XL/2 following ADM~\cite{dhariwal2021diffusion}.

\vspace{-15pt}
\input{tab/sit_full}
\vspace{-15pt}

\noindent
\Cref{tab:sit_n} shows that $N=8$ leads to divergence, while $N=32$ leads to slower learning, so we pick $N=16$. \Cref{tab:sit_ot} shows that $\lambda_\mathrm{ot}=0$ yields the best result for post-training. \Cref{tab:sit_lr} shows that increasing the learning rate for both $G,D$ causes grad norm spikes and divergence. \Cref{tab:sit_epoch} shows that FID stays the same after training longer. \Cref{tab:sit_f} shows that the least square loss~\cite{mao2017least} produces stronger results than non-saturating loss~\cite{goodfellow2014generative}.

\begin{table}[h]
    \centering
    \begin{minipage}[t]{0.48\linewidth}
        \input{tab/sit_N}
    \end{minipage}
    \hfill
    \begin{minipage}[t]{0.48\linewidth}
        \input{tab/sit_ot}
    \end{minipage}
    \\
    \vspace{20pt}
    \begin{minipage}[t]{0.48\linewidth}
        \input{tab/sit_lr}
    \end{minipage}
    \hfill
    \begin{minipage}[t]{0.48\linewidth}
        \input{tab/sit_iter}
    \end{minipage}
\end{table}

\clearpage

\input{tab/sit_f}

\noindent
\Cref{tab:jit_full} shows the full evaluation metrics of JiT-H/16. The FID and IS metrics are computed using the code provided by JiT, and the other metrics are computed using the code provided by ADM. The JiT settings follow those of SiT, and we do not conduct separate ablation studies on JiT.

\input{tab/jit_full}

\clearpage

\noindent
\Cref{tab:imagenet hyperparameter} enumerates the hyperparameters used for our CAFM post-training.
\vspace{-15pt}
\input{tab/imagenet_hyperparams}

\noindent
Qualitative comparisons are provided in \cref{fig:sit,fig:sit_cfg,fig:jit,fig:jit_cfg} in the following pages.

\input{fig/sit}
\input{fig/sit_cfg}

\input{fig/jit}
\input{fig/jit_cfg}

\clearpage
\section{Additional Results on Text-to-Image Post-training}

\paragraph{Architecture.} Our text-to-image generation experiments are conducted on Z-Image~\cite{cai2025z} model, an open-source, 6B-parameter, single-stream diffusion transformer. We use the pre-distillation checkpoint, which is suitable for our continuous flow experiments. For the generator, we adopt the exact architecture without changes. For the discriminator, we follow APT~\cite{lin2025diffusion} to add a cross-attention layer on the visual features at the last layer to project the discriminator logit. Compared to inserting [CLS] at input, this design allows most parts of the transformer to stay intact. Because Z-Image already uses RMSNorm, no changes are made to the normalization layers.

\paragraph{Dataset and training.} Z-Image is trained on proprietary supervised finetuning (SFT) data, which are inaccessible to us. Also, the SFT data likely contain high-quality images generated by prior text-to-image models using CFG, which is implicitly a form of CFG distillation. For our experiments, we use open-source image datasets that contain only natural images, filtered and recaptioned. To eliminate the dataset from being an influencing factor, we first finetune Z-Image on our data using the FM objective for 10k iterations, and we find that the model quickly adapts. Then we run CAFM finetuning, while keeping the FM trial running with equivalent iterations for comparison. The CAFM finetuning is run for a total of 20k iterations including both $G,D$ updates. The hyperparameters are listed in \cref{tab:t2i_hyperparams}. 
\vspace{-10pt}
\input{tab/t2i_hyperparams}

\paragraph{Additional results.} \Cref{tab:geneval_full,tab:dpg_full} shows the metrics including the original Z-Image model for reference. In GenEval, our CAFM-finetuned model beats both the original and our FM-finetuned baseline. But in the DPG benchmark, our model performs worse than the original ZImage model, which we believe is due to the use of different datasets. These tables are provided only for reference. Only the FM-finetuned model is the fair comparison baseline, and for this reason, we removed the original Z-Image model from the tables in the main text. More qualitative comparisons are provided in \cref{fig:t2i_dpg}.

\clearpage
\vspace*{\fill}
\input{tab/geneval_full}
\vspace*{\fill}
\input{tab/dpg_full}
\vspace*{\fill}
\clearpage

\input{fig/t2i_dpg1}
\input{fig/t2i_dpg2}
\input{fig/t2i_dpg3}
\input{fig/t2i_dpg4}

\clearpage
\paragraph{Limitation.} Despite CAFM improving the generalization of the model, \Cref{fig:t2i_failure} shows that guidance-free generation can still yield incorrect images sometimes, especially in the low-density regions containing
outliers. Our research does not claim that our method can reach production quality in the guidance-free setting, but only to demonstrate that our method can improve the generalization of the model, which yields gains in both the guidance-free and guided settings.

\vspace{-10pt}
\input{fig/t2i_failure}

\section{On Flow Matching Objective}
\label{app:flow-matching-objective}

This section shows that criteria other than the squared $L_2$ metric can also be valid for flow matching.

Consider the criterion:
\begin{equation}
    d(a, b) = (a - b)^\top M (a - b),
\end{equation}
where the squared $L_2$ criterion corresponds to the special case $M = I$.

We show that any strictly positive definite (SPD) matrix $M \in \mathbb{R}^{n \times n}$, defined by:
\begin{equation}
    d(a, b) = (a - b)^\top M (a - b) > 0, \quad \forall a\ne b,
\end{equation}
also satisfies:
\begin{equation}
    \arg\min_a \mathbb{E}_b[d(a, b)] = \mathbb{E}[b],
\end{equation}
and hence converges to the marginal velocity $v_t = \mathbb{E}[\bar{v}_t \mid x_t]$ under the conditional flow matching objective:
\begin{equation}
    \mathcal{L}_\mathrm{FM} = \mathbb{E}_{x,z,t} [d(G(x_t, t), \bar{v}_t)].
\end{equation}

First, we expand the expectation, where the expectation is taken over $b$:
\begin{align}
    \mathbb{E}[d(a, b)] &= \mathbb{E}[(a - b)^\top M (a - b)] \\
    &= a^\top M a - 2a^\top M \mathbb{E}[b] + \mathbb{E}[b^\top M b].
\end{align}

Then, we take the derivative with respect to $a$. The last term, $\mathbb{E}[b^\top M b]$, is constant with respect to $a$ and is therefore dropped:
\begin{equation}
    \nabla_a \mathbb{E}[d(a, b)] = 2Ma - 2M\mathbb{E}[b].
\end{equation}

Finally, we set the gradient to zero:
\begin{align}
    2Ma - 2M\mathbb{E}[b] &= 0, \\
    M(a - \mathbb{E}[b]) &= 0.
\end{align}

Since $M$ is strictly positive definite, it has no zero eigenvalues, so the only solution is:
\begin{align}
    a - \mathbb{E}[b] = 0, \\
    a = \mathbb{E}[b].
\end{align}
This concludes that the minimizer in $a$ is exactly $\mathbb{E}[b]$, just as in the standard $L_2$ case. In theory, the choice of strictly positive definite $M$ is irrelevant, since all such choices converge to the ground-truth marginal velocity. In practice, however, different choices of $M$ can affect model generalization under finite capacity.

\section{On Discriminator JVP Designs}
\label{app:jvp-design}

One may ask why we formulate $D$ using JVP instead of simply defining it as:
\begin{equation}
    D(x_t, t, v_t) : (\mathbb{R}^n \times [0,1] \times \mathbb{R}^n) \rightarrow \mathbb{R},
\end{equation}
and define the adversarial training objectives as follows:
\begin{align}
    \mathcal{L}^D_\textrm{adv'} = \mathbb{E}_{x,z,t} \left[ f\big(D(x_t, t, v_t), D(x_t, t, G(x_t, t))\big) \right],\label{eq:naive_D} \\
    \mathcal{L}^G_\textrm{adv'} = \mathbb{E}_{x,z,t} \left[ f\big(D(x_t, t, G(x_t, t)), D(x_t, t, v_t)\big) \right].\label{eq:naive_G}
\end{align}
This naive formulation has at least two fundamental problems:

First, the ground-truth marginal velocity $v_t$ used in~\cref{eq:naive_D,eq:naive_G} is generally inaccessible during training. If we replace it with the conditional velocity $\bar{v}_t$, the objective no longer enforces learning of the true marginal velocity field. The key issue is that $D$ is nonlinear, so in general
\begin{equation}
    \mathbb{E}[D(\bar{v})] \neq D(\mathbb{E}[\bar{v}]).
\end{equation}
Therefore, matching discriminator responses to conditional targets does not imply matching the marginal target. At equilibrium, $G$ would need to represent the full conditional-velocity distribution at each $x_t$; however, for a fixed $(x_t,t)$, the generator outputs only a single deterministic velocity. Consequently, $G$ is repeatedly pulled toward incompatible conditional targets $\bar{v}_t$, leading to oscillatory updates.

Second, even in the idealized setting where the marginal velocity $v_t$ were accessible, this direct formulation can suffer from vanishing or uninformative gradients. The real target at each $x_t$ is effectively a point mass (Dirac-like) in velocity space. When the supports overlap poorly, the discriminator can separate real and generated samples too easily, saturate, and provide weak learning signals to $G$.

Our work instead formulates the discriminator as $D(x_t, t)$ and performs discrimination in JVP space:
\begin{equation}
    D_\mathrm{jvp}(x_t, t, \dot{x}_t, \dot{t}) = \frac{\partial D(x_t,t)}{\partial x_t}\dot{x}_t + \frac{\partial D(x_t,t)}{\partial t}\dot{t},
    \label{eq:d_jvp_appendix}
\end{equation}
with adversarial objectives:
\begin{align}
    \mathcal{L}_\mathrm{adv'}^D = \mathbb{E}_{x,z,t} [ f(&D_\mathrm{jvp}(x_t, t, \bar{v}_t, T),
    D_\mathrm{jvp}(x_t, t, G(x_t,t), T)) ], \\
    \mathcal{L}_\mathrm{adv'}^G = \mathbb{E}_{x,z,t} [ f(&D_\mathrm{jvp}(x_t, t, G(x_t,t), T),
    D_\mathrm{jvp}(x_t, t, \bar{v}_t, T)) ],
\end{align}
where
\begin{equation}
    f(a, b) = (a - 1)^2 + (b + 1)^2
\end{equation}
is an LSGAN-like~\cite{mao2017least} contrastive function.

This encourages:
\begin{align}
    D_\mathrm{jvp}(x_t, t, v_t, 1) &= +1, \label{eq:dis_pos_appendix}\\
    D_\mathrm{jvp}(x_t, t, G(x_t, t), 1) &= -1 \label{eq:dis_neg_appendix}.
\end{align}

By~\cref{eq:d_jvp_appendix} and linearity of expectation,
\begin{align}
    \mathbb{E} [&D_\mathrm{jvp}(x_t, t, \bar{v}_t, 1) - D_\mathrm{jvp}(x_t, t, G(x_t, t), 1)] \\
    &= \mathbb{E} \left[\frac{\partial D(x_t,t)}{\partial x_t} (\bar{v}_t - G(x_t, t))\right] \\
    &= \frac{\partial D(x_t,t)}{\partial x_t} (\mathbb{E}[\bar{v}_t] - G(x_t, t)) \\
    &= \frac{\partial D(x_t,t)}{\partial x_t} (v_t - G(x_t, t)).
\end{align}

When $G(x_t, t) \neq v_t$, $D_\mathrm{jvp}$ can be optimized toward~\cref{eq:dis_pos_appendix,eq:dis_neg_appendix}. Equilibrium is reached only when $G(x_t, t) = v_t$ for all $x_t$, in which case
\begin{equation}
    \frac{\partial D(x_t,t)}{\partial x_t} (v_t - G(x_t, t)) = 0.
\end{equation}

However, the above derivation relies only on the linearity of JVP. An alternative is to parameterize $D$ using separate networks:
\begin{equation}
    D(x_t, t, v_t) = A(x_t, t)^\top v_t + B(x_t, t),
\end{equation}
where
\begin{align}
    &A(x_t, t) : \mathbb{R}^n \times [0, 1] \rightarrow \mathbb{R}^n,\\
    &B(x_t, t) : \mathbb{R}^n \times [0, 1] \rightarrow \mathbb{R}^1.
\end{align}

This approach differs from JVP because JVP additionally enforces:
\begin{align}
    A(x_t, t) &= \nabla_x D(x_t, t), \\
    B(x_t, t) &= \partial_t D(x_t, t).
\end{align}
Because of these constraints,
\begin{equation}
    D(x_1, 1) - D(x_0, 0) = \int_0^1 D_\mathrm{jvp}(x_t, t, v_t, 1)\ dt.
\end{equation}
The discriminator is globally consistent along trajectories. Empirically, we find that these constraints improve optimization, and the separate formulation does not yield good results on high-dimensional data.

\clearpage
\section{On the Vanishing-Gradient Problem}
\label{app:vanishing-gradient}

The vanishing-gradient problem in adversarial training arises when the real data distribution and the generator distribution have non-overlapping support. Formally, the support of a distribution $p(x)$ is the set on which it assigns positive probability mass:
\begin{equation}
    \mathrm{supp}(p) = \{x \mid p(x) > 0\}.
\end{equation}

In high-dimensional spaces, data distributions (e.g., images) often concentrate near a tiny, low-dimensional manifold. When the support of the real data distribution $S_{\text{data}}$ and that of the generator distribution $S_G$ do not overlap,
\begin{equation}
    S_{\text{data}} \cap S_G = \emptyset,
\end{equation}
an optimal discriminator can separate the two distributions perfectly. As a result, it forms a steep decision boundary in regions between the supports while producing nearly flat gradients around generated samples, yielding little useful learning signal for the generator.

There are two approaches to theoretically mitigating the vanishing-gradient issue. WGAN~\cite{arjovsky2017wasserstein} proposes using the Wasserstein-1 distance, but it requires the discriminator network to be 1-Lipschitz. Enforcing this condition is difficult in practice, so it is often replaced by gradient penalties~\cite{gulrajani2017improved,roth2017stabilizing}, which provide a softer constraint at the cost of weaker theoretical guarantees. Another approach is to add instance noise~\cite{mescheder2018training}:
\begin{equation}
    \tilde{x} = x + \epsilon, \quad \epsilon \sim \mathcal{N}(0, \sigma^2\mathbf{I}).
\end{equation}
This is equivalent to convolving the data distribution with a Gaussian distribution, which has support everywhere. Therefore, the resulting distribution also has support everywhere.

Although $\sigma > 0$ can be arbitrarily small and still ensures that the convolved distribution has support everywhere in theory, the off-manifold region carries only tiny probability mass. In practice, the discriminator can still form sharp decision boundaries in these regions. Increasing $\sigma$ further mitigates gradient vanishing, but it also causes the model to learn a noisy distribution rather than the intended data distribution. The optimal $\sigma$ depends on the distance between the real and generator manifolds and on the capacity of the discriminator.

Adversarial flow models~\cite{lin2025adversarial} allow piecewise learning on the probability flow $G(x_s, s, t) \rightarrow x_t$. The vanishing-gradient problem should be further mitigated as the discretization interval $|t-s|$ decreases. Continuous adversarial flow models extend this formulation to continuous time, where $|t-s| \rightarrow 0$. From this perspective, they should mitigate the vanishing-gradient problem.

From a mathematical perspective, we also show that, for CAFMs, the gradient with respect to $G$ does not vanish under an optimal $D$ because of the linearization effect of JVP. Specifically, let
\begin{equation}
    D_\mathrm{jvp}(x_t, t, G(x_t, t), 1) = J_x(x_t, t) G(x_t, t) + J_t(x_t, t),
    \label{eq:d_jvp_appendix}
\end{equation}
where
\begin{equation}
    J_x(x_t, t)=\frac{\partial D(x_t,t)}{\partial x_t}, \quad J_t(x_t, t)=\frac{\partial D(x_t,t)}{\partial t}
\end{equation}
are the Jacobian matrices of the discriminator network. In backpropagation, the gradient with respect to $G(x_t, t)$ is simply

\begin{equation}
    \frac{\partial\mathcal{L}}{\partial G(x_t, t)} = J_x(x_t, t)^\top g,
\end{equation}
where
\begin{equation}
    g := \frac{\partial \mathcal{L}}{ \partial D_\mathrm{jvp}}
\end{equation}
is the gradient propagated from the loss.

Therefore, $G(x_t, t)$ receives a nonzero gradient whenever $g \ne 0$ and $J_x(x_t, t) \ne 0$. We previously showed in~\cref{app:jvp-design} that whenever $G(x_t, t) \ne v_t$, an optimal $D$ learns a nonzero $J_x(x_t, t)$ for discrimination.

For the least squares contrastive function
\begin{equation}
    f(a, b) = (a - 1)^2 + (b + 1)^2.
\end{equation}
The derivative with respect to $a$ is:
\begin{equation}
    \frac{df(a,b)}{da} = 2(a-1).
\end{equation}
Therefore, it has a non-zero gradient with respect to input $a$ for any $a \ne 1$. In the case of an optimal $D$, it outputs $-1$ for the $v_t$ prediction by $G$.

In practice, we also find that our method can train properly without gradient penalties and discriminator augmentation~\cite{karras2020training}.

\newpage
\section{On Implementation of JVP}
\label{app:implementation}

Our experiment is conducted in PyTorch, where we use \texttt{torch.func.jvp} and \texttt{torch.func.vmap} for forward-mode Jacobian-vector product (JVP) and vectorizing map (Vmap). Both functions are compatible with DDP, FSDP, and gradient checkpointing with a special arrangement. 

Specifically, it is important to implement it as \texttt{ddp(jvp(D))}, instead of \texttt{jvp(ddp(D))}, so that JVP (and Vmap) is only wrapped around the network as a regular PyTorch module, instead of applying JVP to DDP, which includes incompatible gradient synchronization logic. DDP is used for the ImageNet experiments.

For FSDP and gradient checkpointing, an example implementation can be found in prior work, rCM~\cite{zheng2025large}. rCM wraps JVP on every \texttt{nn.Module}, which we find unnecessary and excessive. It is sufficient to only wrap JVP (and Vmap) to the top-level submodules for FSDP sharding and gradient checkpointing. FSDP and gradient checkpointing are used for the text-to-image generation experiments.

For attention, we use the \texttt{math} fused kernel of PyTorch's scaled dot product attention, which supports both JVP and Vmap natively and is sufficient for image generation training.

\section{On LayerNorm and RMSNorm}

We empirically find that switching the discriminator LayerNorm~\cite{ba2016layer} to RMSNorm~\cite{zhang2019root} significantly improves training stability. \Cref{fig:norm} shows the discriminator gradient norm when pre-training the SiT-B/2 model on ImageNet 256px. LayerNorm causes large spikes in discriminator gradient norm, whereas RMSNorm does not under equal settings. Prior work involving JVP has also found that RMSNorm provides better stability~\cite{zhou2025terminal}.  

\input{fig/norm}

\clearpage
\section{On Computational Efficiency}

\paragraph{Discriminator update count $N$.} Our method performs $N$ discriminator updates for each generator update, with the goal of keeping the discriminator close to its local optimum throughout training. Importantly, this schedule does \emph{not} imply that the generator converges $N$ times more slowly than in standard flow matching. In classical flow matching, the model estimates the marginal velocity field by taking an expectation over Monte Carlo samples of conditional velocities. Analogously, in our framework, the discriminator estimates flow potentials using Monte Carlo samples of conditional velocities. The key distinction is that the generator is then updated using these learned potentials, moving in the direction that increases the potential most strongly. This coupling enables stable and informative generator updates even when discriminator optimization is emphasized.

\paragraph{JVP computation efficiency.} For ImageNet SiT-XL/2 post-training, CAFM requires approximately $4.8\times$ more wall-clock time per epoch than FM. This overhead arises from introducing an additional discriminator network, along with its forward pass and backward JVP computations. We consider this additional computation acceptable for post-training applications.

%% file: tab/sit_full.tex
\begin{table}[h]
    \centering
    \scriptsize
    \caption{SiT-XL/2 full metrics on ImageNet 256px.}
    \label{tab:sit_full}
    \vspace{-5pt}
    \setlength{\tabcolsep}{3pt}
    \begin{tabular}{l|l|l|rrrrr}
        \toprule
        CFG & Method & Epoch & FID$\downarrow$ & IS$\uparrow$ & sFID $\downarrow$ & Prec.$\uparrow$ & Rec$\uparrow$ \\
        \midrule
        None & SiT & 1400 & 8.26 & 131.65 & 6.32 & 0.68 & 0.67 \\
        & SiT+FM & 1400+10 & 8.64 & 131.91 & 6.36 & 0.68 & 0.67 \\
        & SiT+CAFM & 1400+10 & \textcolor{red}{\textbf{3.63}} & \textbf{178.08} & \textbf{4.72} & \textbf{0.71} & \textbf{0.69} \\
        \midrule
        1.1 & SiT & 1400 & 5.55 & 161.77 & 5.67 & 0.72 & 0.66 \\
        & SiT+FM & 1400+10 & 5.53 & 161.72 & 5.68 & 0.72 & 0.65 \\
        & SiT+CAFM & 1400+10 & \textbf{2.27} & \textbf{212.06} & \textbf{4.55} & \textbf{0.74} & \textbf{0.67} \\
        \midrule
        1.2 & SiT & 1400 & 3.65 & 190.57 & 5.19 & 0.75 & 0.64 \\
        & SiT+FM & 1400+10 & 3.62 & 191.56 & 5.19 & 0.75 & 0.64 \\
        & SiT+CAFM & 1400+10 & \textbf{1.66} & \textbf{238.44} & \textbf{4.49} & \textbf{0.77} & \textbf{0.66} \\
        \midrule
        1.3 & SiT & 1400 & 2.57 & 220.52 & 4.83 & \textbf{0.78} & 0.63 \\
        & SiT+FM & 1400+10 & 2.55 & 219.89 & 4.85 & 0.78 & 0.62 \\
        & SiT+CAFM & 1400+10 & \textcolor{red}{\textbf{1.53}} & \textbf{263.52} & \textbf{4.59} & \textbf{0.78} & \textbf{0.64} \\
        \midrule
        1.4 & SiT & 1400 & 2.07 & 248.31 & \textbf{4.61} & \textbf{0.80} & 0.61 \\
        & SiT+FM & 1400+10 & 2.09 & 247.98 & 4.63 & \textbf{0.80} & 0.61 \\
        & SiT+CAFM & 1400+10 & \textbf{1.66} & \textbf{283.59} & 4.75 & 0.79 & \textbf{0.63} \\
        \midrule
        1.5 & SiT & 1400 & 2.06 & 277.50 & \textbf{4.49} & \textbf{0.83} & 0.59 \\
        & SiT+FM & 1400+10 & 2.02 & 270.94 & 4.53 & 0.82 & 0.59 \\
        & SiT+CAFM & 1400+10 & \textbf{1.97} & \textbf{301.91} & 5.03 & 0.81 & \textbf{0.62} \\
        \midrule
        1.6 & SiT & 1400 & \textbf{2.25} & 293.72 & \textbf{4.51} & \textbf{0.84} & 0.58 \\
        & SiT+FM & 1400+10 & 2.26 & 292.53 & 4.53 & \textbf{0.84} & 0.57 \\
        & SiT+CAFM & 1400+10 & 2.37 & \textbf{316.78} & 5.35 & 0.81 & 0.60 \\
        \bottomrule
    \end{tabular}
\end{table}

%% file: tab/sit_N.tex
    \centering
    \scriptsize
    \caption{SiT-XL/2 CAFM post-train ablation study on $N$.}
    \label{tab:sit_n}
    \vspace{-5pt}
    \setlength{\tabcolsep}{8pt}
    \begin{tabular}{c|ccc}
        \toprule
        $N$ & 8 & 16 & 32 \\
        \midrule
        FID$\downarrow$ & 294.91 & \textbf{3.63} & 3.68 \\
        \bottomrule
    \end{tabular}

%% file: tab/sit_ot.tex
    \centering
    \scriptsize
    \caption{SiT-XL/2 CAFM post-train ablation study on $\lambda_\mathrm{ot}$.}
    \label{tab:sit_ot}
    \vspace{-5pt}
    \setlength{\tabcolsep}{8pt}
    \begin{tabular}{c|cc}
        \toprule
        $\lambda_\mathrm{ot}$ & 0 & 0.01 \\
        \midrule
        FID$\downarrow$ & \textbf{3.63} & 4.50 \\
        \bottomrule
    \end{tabular}

%% file: tab/sit_lr.tex
    \centering
    \scriptsize
    \caption{SiT-XL/2 CAFM post-train ablation study on learning rate.}
    \label{tab:sit_lr}
    \vspace{-5pt}
    \setlength{\tabcolsep}{8pt}
    \begin{tabular}{c|cc}
        \toprule
        LR & 1e-5 & 5e-5 \\
        \midrule
        FID$\downarrow$ & \textbf{3.63} & 283.96 \\
        \bottomrule
    \end{tabular}

%% file: tab/sit_iter.tex
    \centering
    \scriptsize
    \caption{SiT-XL/2 CAFM post-train ablation study on epochs.}
    \label{tab:sit_epoch}
    \vspace{-5pt}
    \setlength{\tabcolsep}{8pt}
    \begin{tabular}{c|cc}
        \toprule
        Epoch & 10 & 20 \\
        \midrule
        FID$\downarrow$ & \textbf{3.63} & 3.64 \\
        \bottomrule
    \end{tabular}

%% file: tab/sit_f.tex
\begin{table}[h]
    \centering
    \scriptsize
    \caption{
    SiT-XL/2 CAFM post-train ablation study on $f(a,b)$.\\
    Non-saturating~\cite{goodfellow2014generative}: $f(a,b)=-\log(\sigma(a))-\log(1-\sigma(b))$.\\
    Hinge~\cite{lim2017geometric}: $f^D(a,b)=\max(0,1-a)+\max(0,1+b),\ f^G(a,b)=-a+b$. \\
    Least squares~\cite{mao2017least}: $f(a,b)=(a-1)^2+(b+1)^2$.\\
    CFG is swept separately and the best result is reported for each.
    }
    \label{tab:sit_f}
    \setlength{\tabcolsep}{3pt}
    \begin{tabular}{l|l|l|rrrrr}
        \toprule
        CFG & $f(a,b)$ & Epoch & FID$\downarrow$ & IS$\uparrow$ & sFID $\downarrow$ & Prec.$\uparrow$ & Rec$\uparrow$ \\
        \midrule
        None & Non-saturating & 1400+10 & \textbf{3.54} & 167.45 & 5.20 & \textbf{0.72} & 0.68 \\
        & Hinge & 1400+10 & 4.00 & 175.21 & 5.36 & 0.71 & 0.68 \\
        & \textbf{Least squares} & 1400+10 & 3.63 & \textbf{178.08} & \textbf{4.72} & 0.71 & \textbf{0.69} \\
        \midrule
        1.3 & Non-saturating & 1400+10 & 1.58 & 245.10 & 4.84 & \textbf{0.79} & \textbf{0.64} \\
        & Hinge & 1400+10 & 1.57 & 258.46 & 4.86 & \textbf{0.79} & \textbf{0.64} \\
        & \textbf{Least squares} & 1400+10 & \textbf{1.53} & \textbf{263.52} & \textbf{4.59} & 0.78 & \textbf{0.64} \\
        \bottomrule
    \end{tabular}
\end{table}

%% file: tab/jit_full.tex
\begin{table}[H]
    \centering
    \scriptsize
    \caption{JiT-H/16 full metrics on ImageNet 256px.}
    \label{tab:jit_full}
    \setlength{\tabcolsep}{3pt}
    \begin{tabular}{l|l|l|rrrrr}
        \toprule
        CFG & Method & Epoch & FID$\downarrow$ & IS$\uparrow$ & sFID $\downarrow$ & Prec.$\uparrow$ & Rec$\uparrow$ \\
        \midrule
        None & JiT & 600 & 7.17 & 151.54 & 5.51 & 0.68 & \textbf{0.67} \\
        & JiT+FM & 600+10 & 9.30 & 139.00 & 6.16 & 0.67 & 0.66 \\
        & JiT+CAFM & 600+10 & \textcolor{red}{\textbf{3.57}} & \textbf{198.08} & \textbf{4.77} & \textbf{0.74} & 0.65 \\
        \midrule
        1.2 & JiT & 600 & 4.60 & 188.88 & 5.34 & 0.71 & \textbf{0.66} \\
        & JiT+FM & 600+10 & 5.97 & 176.99 & 5.88 & 0.71 & 0.65 \\
        & JiT+CAFM & 600+10 & \textbf{2.47} & \textbf{232.69} & \textbf{4.76} & \textbf{0.76} & 0.64 \\
        \midrule
        1.4 & JiT & 600 & 3.24 & 219.52 & 5.25 & 0.74 & \textbf{0.66} \\
        & JiT+FM & 600+10 & 4.07 & 209.78 & 5.66 & 0.74 & 0.64 \\
        & JiT+CAFM & 600+10 & \textbf{2.01} & \textbf{258.46} & \textbf{4.81} & \textbf{0.77} & 0.64 \\
        \midrule
        1.6 & JiT & 600 & 2.49 & 244.61 & 5.22 & 0.76 & \textbf{0.65} \\
        & JiT+FM & 600+10 & 3.01 & 237.46 & 5.51 & 0.75 & 0.63 \\
        & JiT+CAFM & 600+10 & \textbf{1.84} & \textbf{275.96} & \textbf{4.91} & \textbf{0.78} & 0.63 \\
        \midrule
        1.8 & JiT & 600 & 2.12 & 265.43 & 5.20 & 0.77 & \textbf{0.65} \\
        & JiT+FM & 600+10 & 2.43 & 261.26 & 5.41 & 0.77 & 0.63 \\
        & JiT+CAFM & 600+10 & \textcolor{red}{\textbf{1.80}} & \textbf{290.71} & \textbf{5.03} & \textbf{0.79} & 0.63 \\
        \midrule
        2.0 & JiT & 600 & 1.96 & 281.38 & 5.22 & 0.78 & \textbf{0.64} \\
        & JiT+FM & 600+10 & 2.12 & 280.24 & 5.34 & 0.78 & 0.62 \\
        & JiT+CAFM & 600+10 & \textbf{1.83} & \textbf{301.96} & \textbf{5.16} & \textbf{0.79} & 0.63 \\
        \midrule
        2.2 & JiT & 600 & \textbf{1.86} & 303.40 & \textbf{5.27} & 0.78 & \textbf{0.64} \\
        & JiT+FM & 600+10 & 1.98 & 296.68 & 5.31 & 0.79 & 0.62 \\
        & JiT+CAFM & 600+10 & 1.88 & \textbf{310.54} & \textbf{5.27} & \textbf{0.80} & 0.62 \\
        \midrule
        2.4 & JiT & 600 & 2.19 & 310.20 & 5.32 & 0.79 & \textbf{0.63} \\
        & JiT+FM & 600+10 & \textbf{1.94} & 310.06 & \textbf{5.29} & \textbf{0.80} & 0.61 \\
        & JiT+CAFM & 600+10 & 1.95 & \textbf{319.23} & 5.40 & 0.79 & 0.62 \\
        \bottomrule
    \end{tabular}
\end{table}

%% file: tab/imagenet_hyperparams.tex
\begin{table}[H]
    \centering
    \caption{
    ImageNet CAFM post-training hyperparameters \\
    \scriptsize{*We formulate $x_0$ as image while JiT formulates $x_0$ as noise, so the timestep is reversed in writing.}
    }
    \label{tab:imagenet hyperparameter}
    \setlength{\tabcolsep}{3pt}
    \scriptsize
    \begin{tabular}{r|cc}
        \toprule
         & SiT & JiT \\
        \midrule
        $G,D$ learning rate & 1e-5 & 1e-5 \\
        batch size & 256 & 1024 \\
        total epoch & 10 & 10 \\
        $D$ warm-up epoch & 2 & 4 \\
        timesteps & uniform(0, 1) & lognormal(0.8, 0.8)* \\
        CFG interval & [0, 1] & [0, 0.9]* \\
        Sampler & SDE 250 steps & Heun 50 steps \\
        Adam $\beta$ & \multicolumn{2}{c}{(0.0, 0.95)} \\
        weight decay & \multicolumn{2}{c}{0} \\
        $N$ & \multicolumn{2}{c}{16} \\
        $\lambda_\mathrm{ot}$ & \multicolumn{2}{c}{0} \\
        EMA decay & \multicolumn{2}{c}{0.99} \\
        precision & \multicolumn{2}{c}{TF32} \\
        \bottomrule
    \end{tabular}
    
\end{table}

%% file: fig/sit.tex
\begin{figure}[h!]
    \centering
    \captionsetup{justification=raggedright,singlelinecheck=false}
    \begin{subfigure}[t]{\linewidth}
        \includegraphics[width=\linewidth]{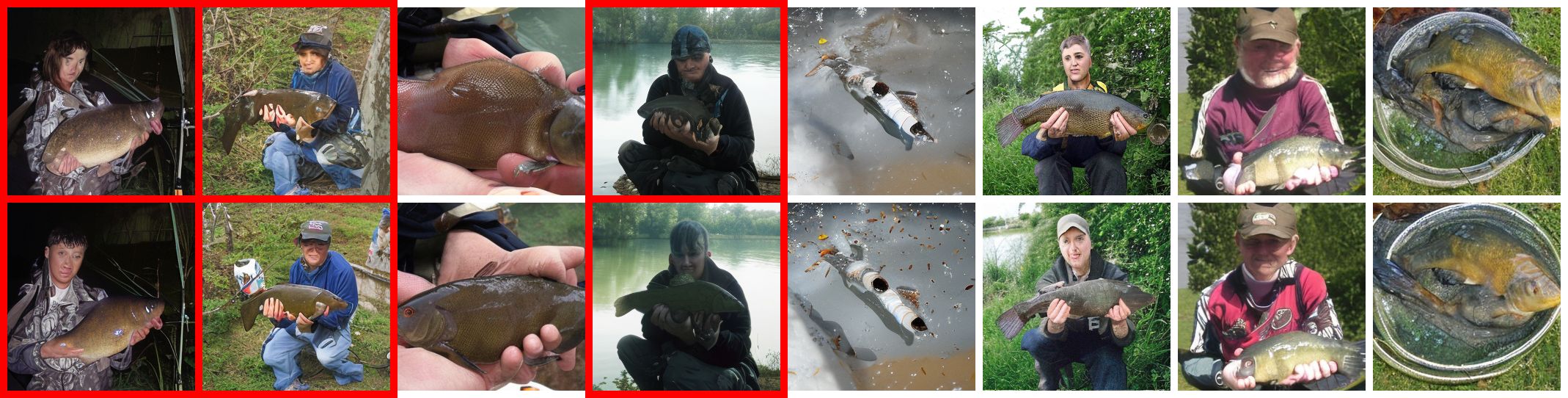}
        \caption{Class 0: tench.}
    \end{subfigure}
    \begin{subfigure}[t]{\linewidth}
        \includegraphics[width=\linewidth]{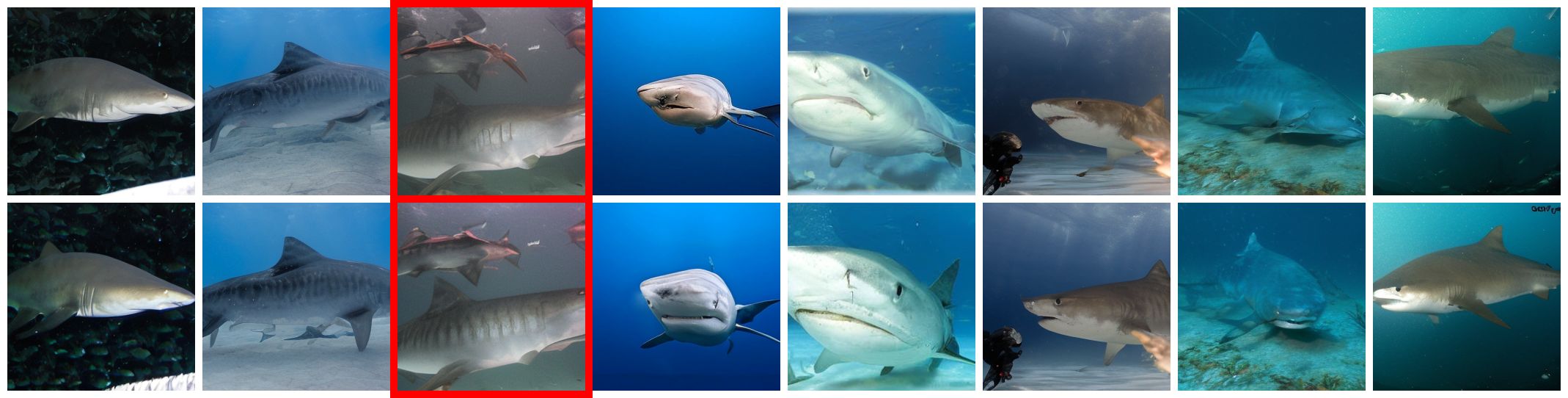}
        \caption{Class 3: tiger shark.}
    \end{subfigure}
    \begin{subfigure}[t]{\linewidth}
        \includegraphics[width=\linewidth]{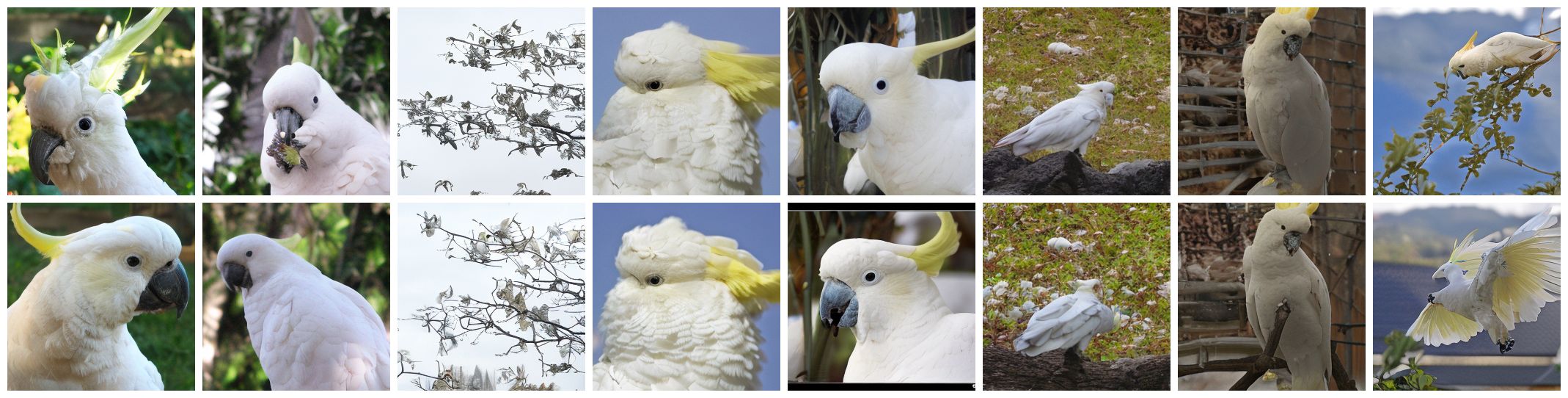}
        \caption{Class 89: cockatoo.}
    \end{subfigure}
    \begin{subfigure}[t]{\linewidth}
        \includegraphics[width=\linewidth]{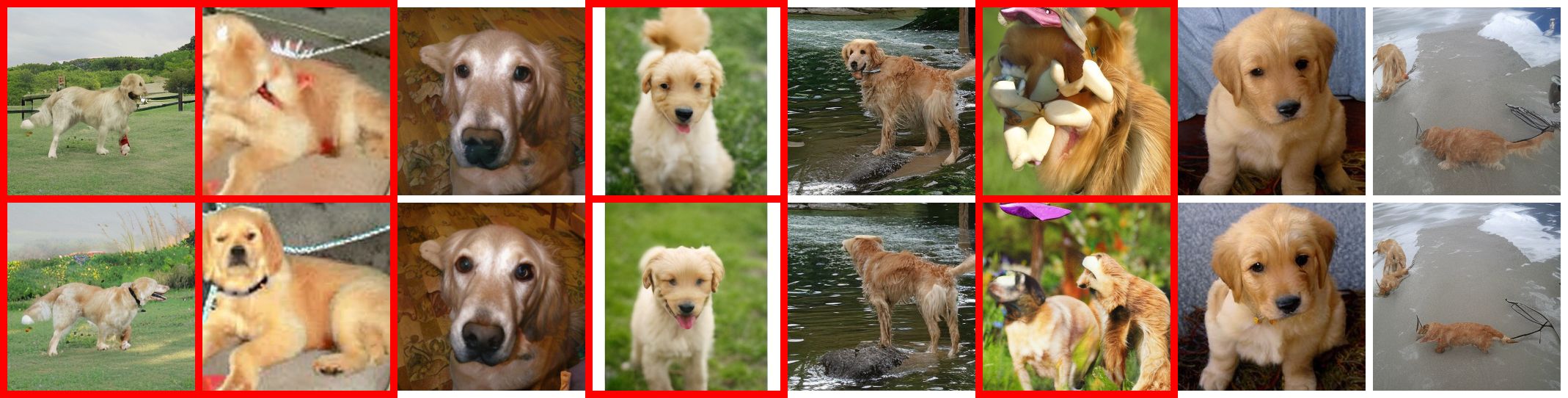}
        \caption{Class 207: golden retriever.}
    \end{subfigure}
    \begin{subfigure}[t]{\linewidth}
        \includegraphics[width=\linewidth]{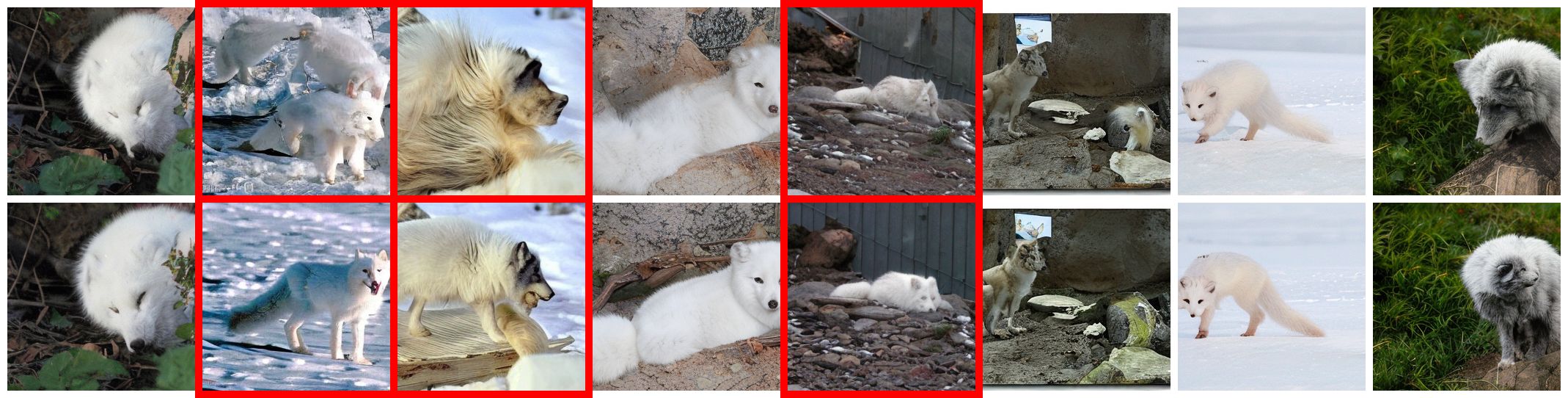}
        \caption{Class 279: white fox.}
    \end{subfigure}
    \caption{
        SiT-XL/2 guidance-free, latent-space ImageNet 256px generation. \\
        Top is FM (FID 8.26). Bottom is CAFM (FID 3.63). \\
        Uncurated. We highlight samples with visible improvements in red.
    }
    \label{fig:sit}
\end{figure}

%% file: fig/sit_cfg.tex
\begin{figure}[h!]
    \centering
    \captionsetup{justification=raggedright,singlelinecheck=false}
    \begin{subfigure}[t]{\linewidth}
        \includegraphics[width=\linewidth]{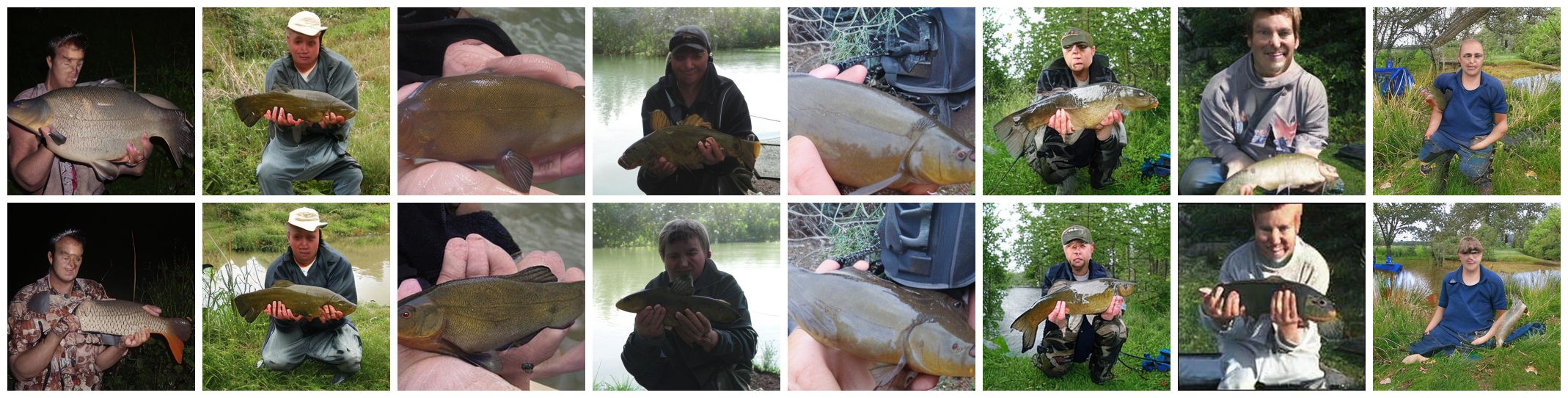}
        \caption{Class 0: tench.}
    \end{subfigure}
    \begin{subfigure}[t]{\linewidth}
        \includegraphics[width=\linewidth]{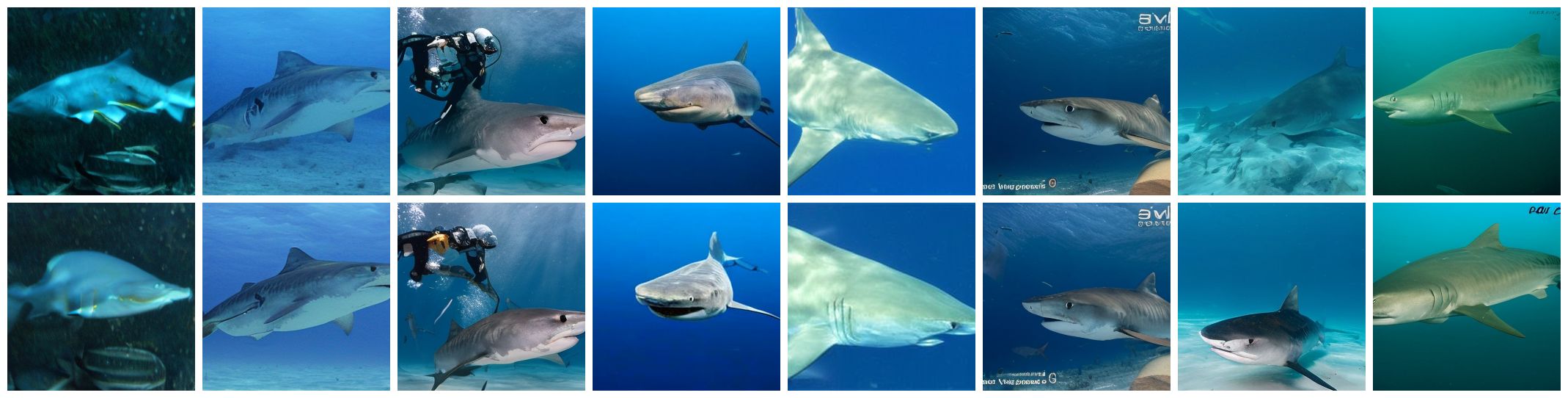}
        \caption{Class 3: tiger shark.}
    \end{subfigure}
    \begin{subfigure}[t]{\linewidth}
        \includegraphics[width=\linewidth]{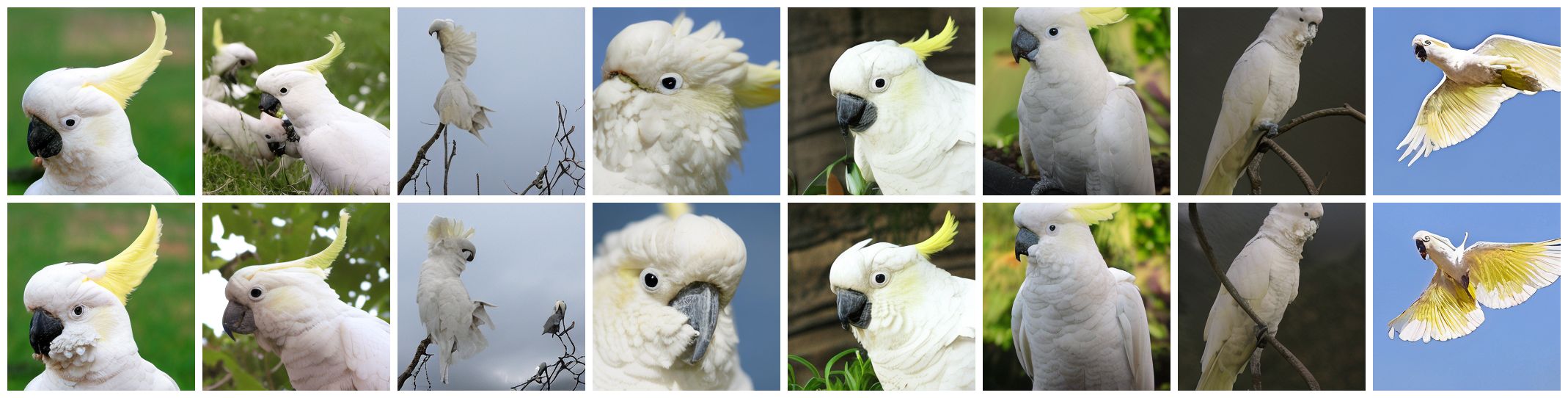}
        \caption{Class 89: cockatoo.}
    \end{subfigure}
    \begin{subfigure}[t]{\linewidth}
        \includegraphics[width=\linewidth]{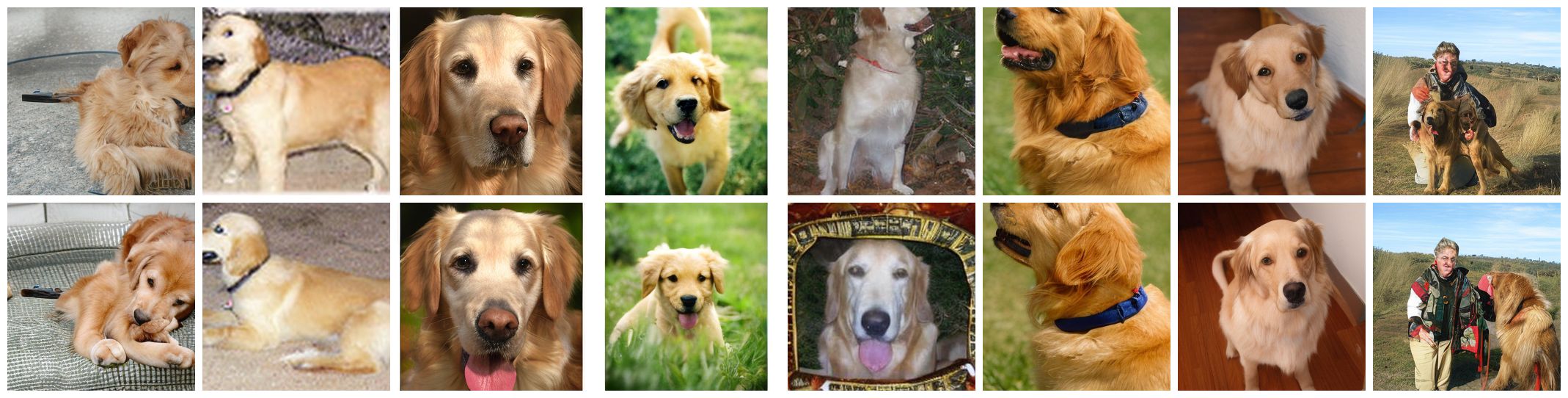}
        \caption{Class 207: golden retriever.}
    \end{subfigure}
    \begin{subfigure}[t]{\linewidth}
        \includegraphics[width=\linewidth]{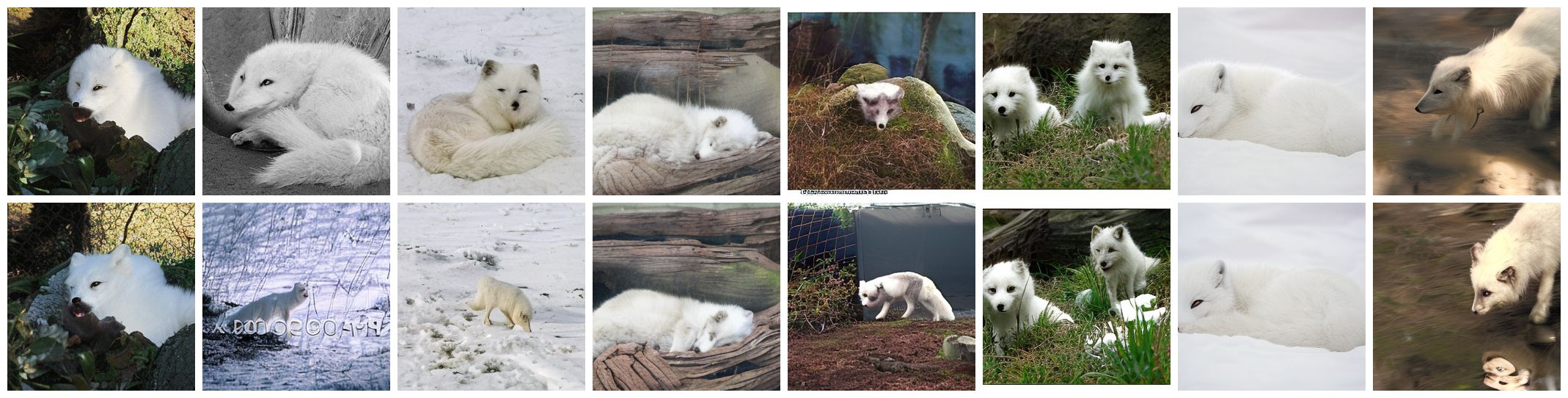}
        \caption{Class 279: white fox.}
    \end{subfigure}
    \caption{
        SiT-XL/2 guided, latent-space ImageNet 256px generation. \\
        Top is FM (CFG 1.5, FID 2.06). Bottom is CAFM (CFG 1.3, FID 1.53). \\
        Uncurated.
    }
    \label{fig:sit_cfg}
\end{figure}

%% file: fig/jit.tex
\begin{figure}[h!]
    \centering
    \captionsetup{justification=raggedright,singlelinecheck=false}
    \begin{subfigure}[t]{\linewidth}
        \includegraphics[width=\linewidth]{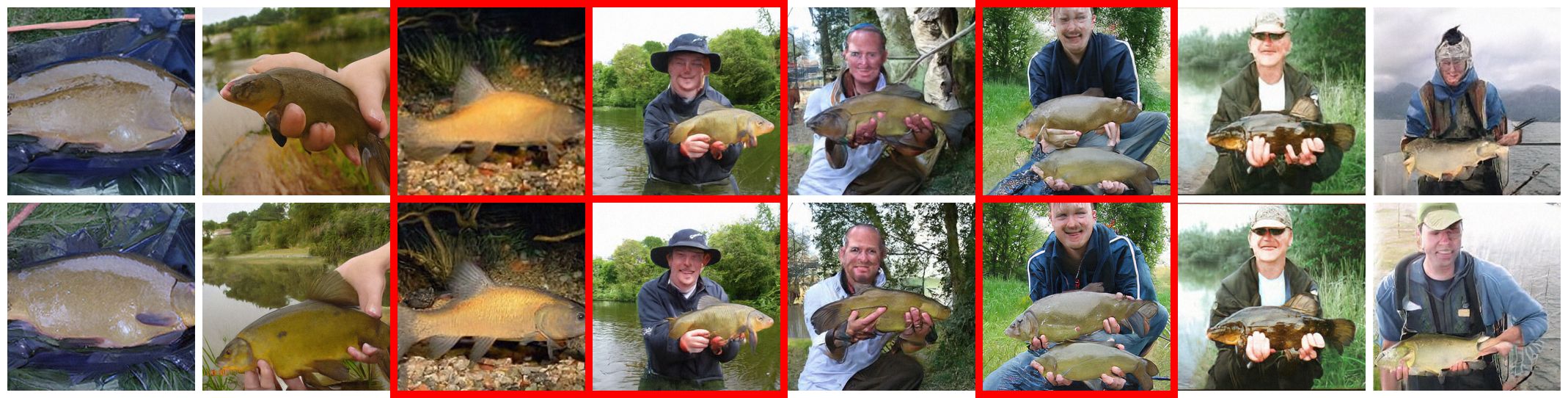}
        \caption{Class 0: tench.}
    \end{subfigure}
    \begin{subfigure}[t]{\linewidth}
        \includegraphics[width=\linewidth]{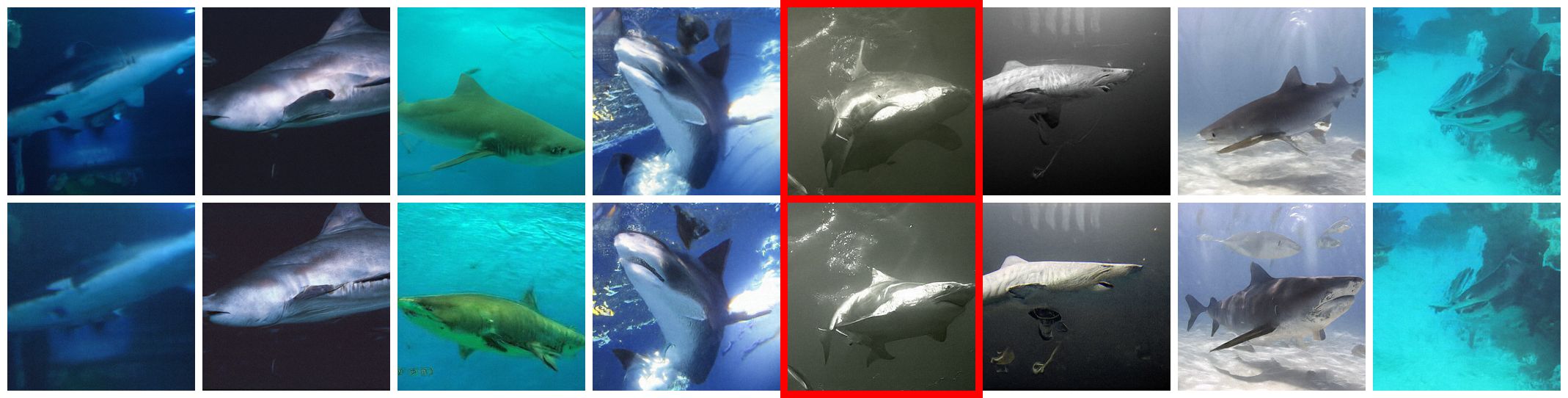}
        \caption{Class 3: tiger shark.}
    \end{subfigure}
    \begin{subfigure}[t]{\linewidth}
        \includegraphics[width=\linewidth]{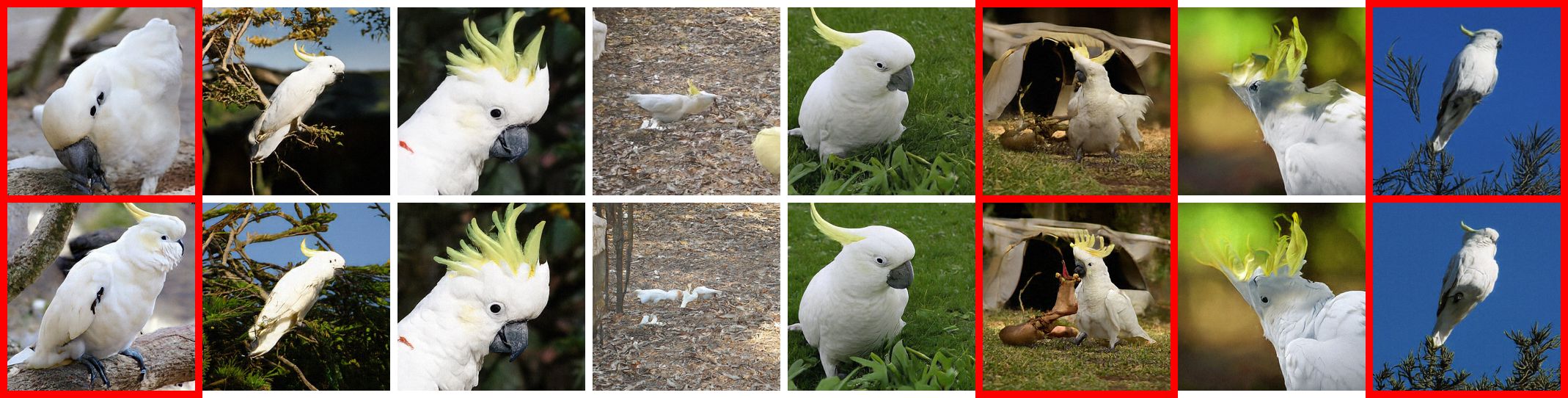}
        \caption{Class 89: cockatoo.}
    \end{subfigure}
    \begin{subfigure}[t]{\linewidth}
        \includegraphics[width=\linewidth]{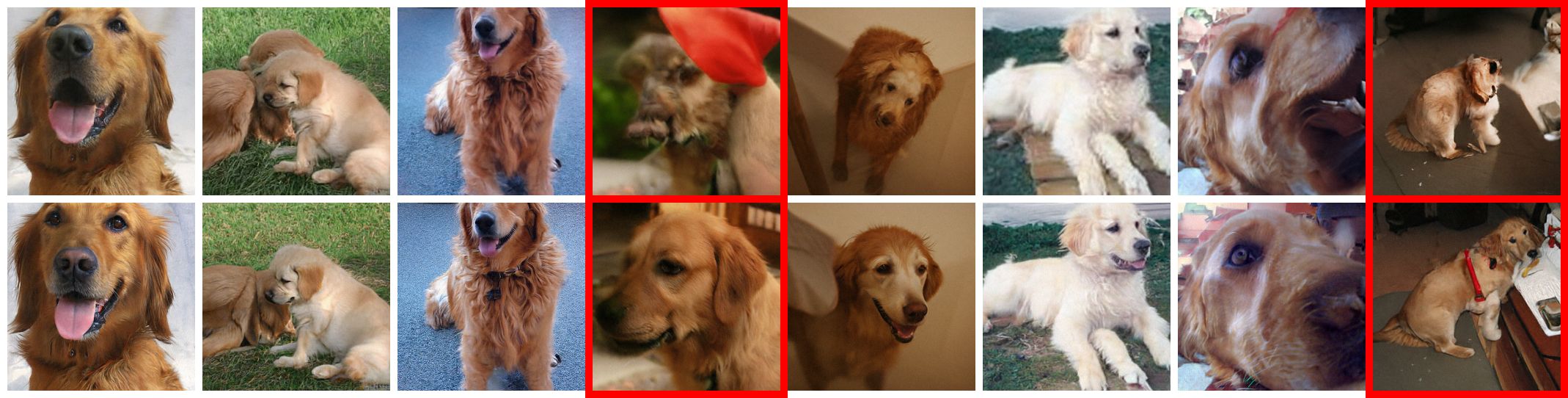}
        \caption{Class 207: golden retriever.}
    \end{subfigure}
    \begin{subfigure}[t]{\linewidth}
        \includegraphics[width=\linewidth]{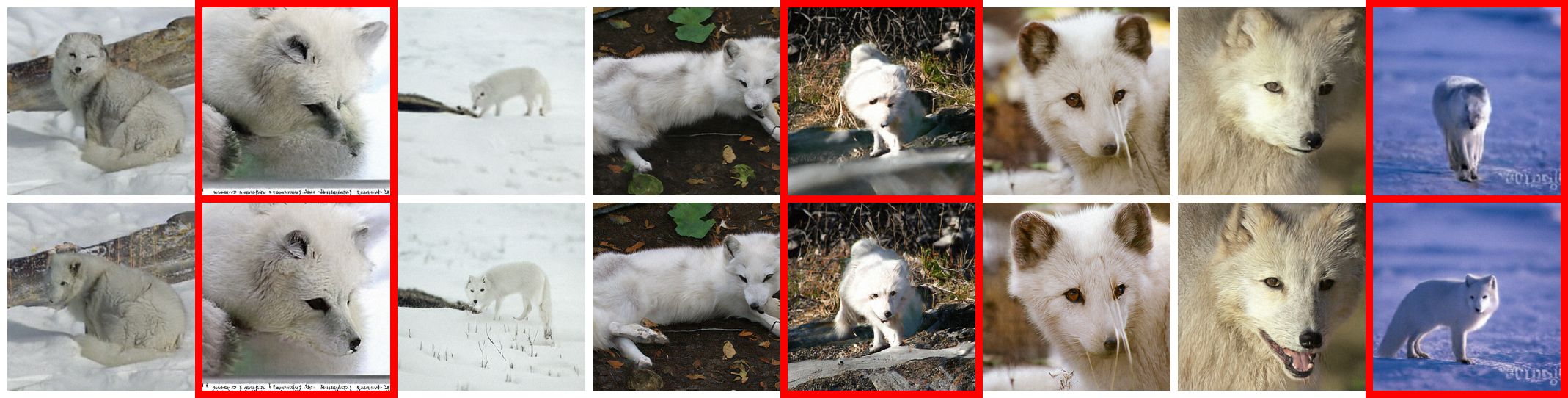}
        \caption{Class 279: white fox.}
    \end{subfigure}
    \caption{
        JiT-H/16 guidance-free, pixel-space ImageNet 256px generation. \\
        Top is FM (FID 7.17). Bottom is CAFM (FID 3.57). \\
        Uncurated. We highlight samples with visible improvements in red.
    }
    \label{fig:jit}
\end{figure}

%% file: fig/jit_cfg.tex
\begin{figure}[h!]
    \centering
    \captionsetup{justification=raggedright,singlelinecheck=false}
    \begin{subfigure}[t]{\linewidth}
        \includegraphics[width=\linewidth]{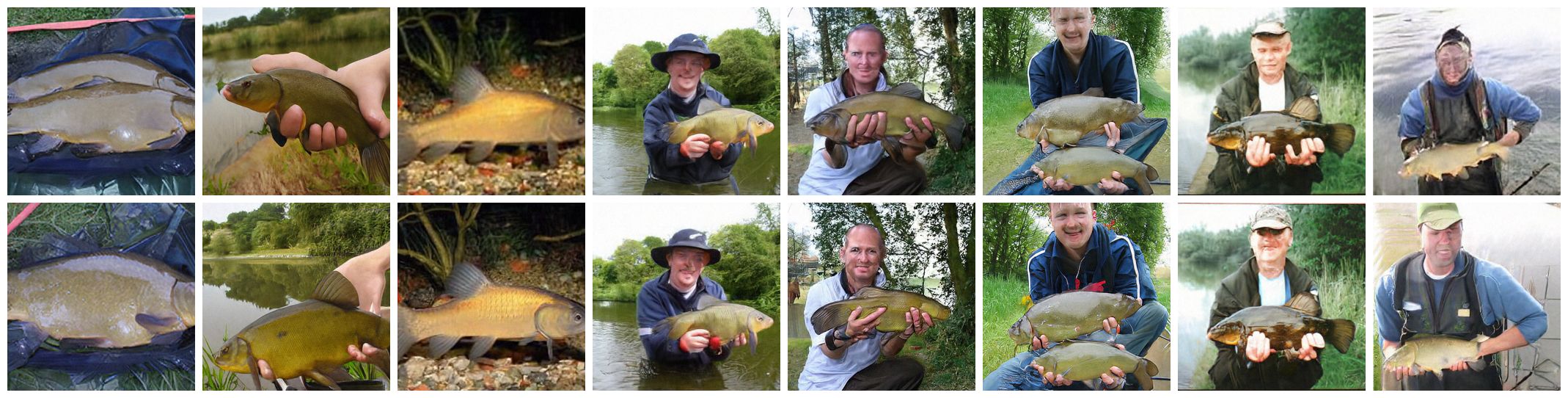}
        \caption{Class 0: tench.}
    \end{subfigure}
    \begin{subfigure}[t]{\linewidth}
        \includegraphics[width=\linewidth]{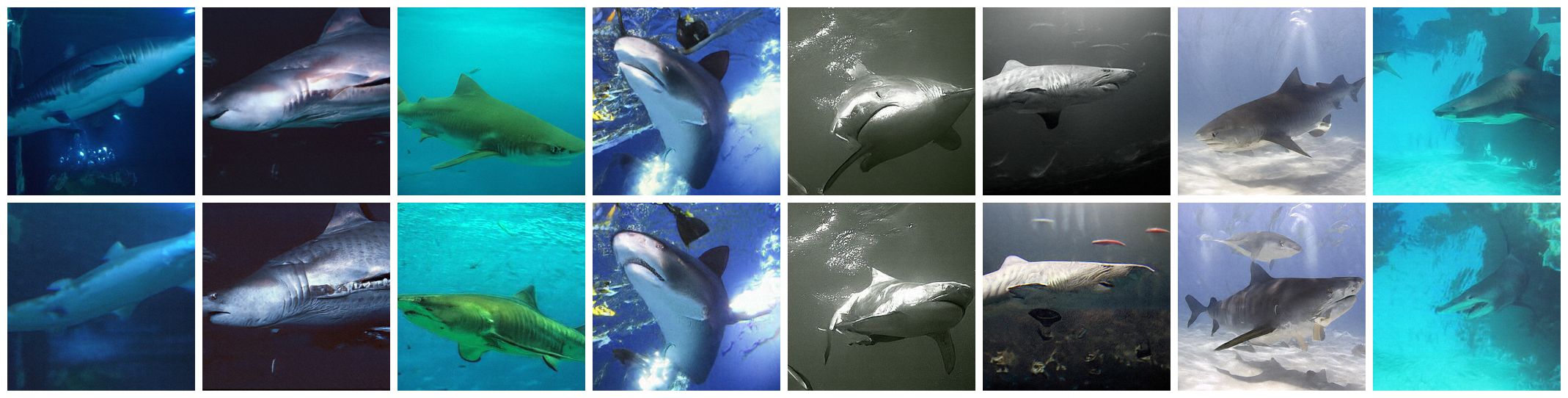}
        \caption{Class 3: tiger shark.}
    \end{subfigure}
    \begin{subfigure}[t]{\linewidth}
        \includegraphics[width=\linewidth]{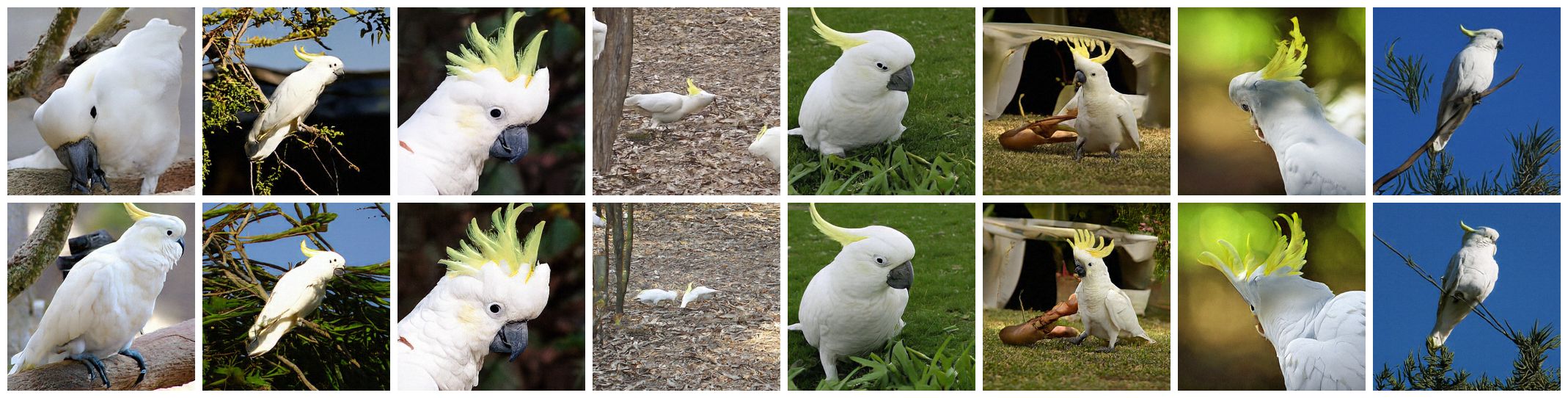}
        \caption{Class 89: cockatoo.}
    \end{subfigure}
    \begin{subfigure}[t]{\linewidth}
        \includegraphics[width=\linewidth]{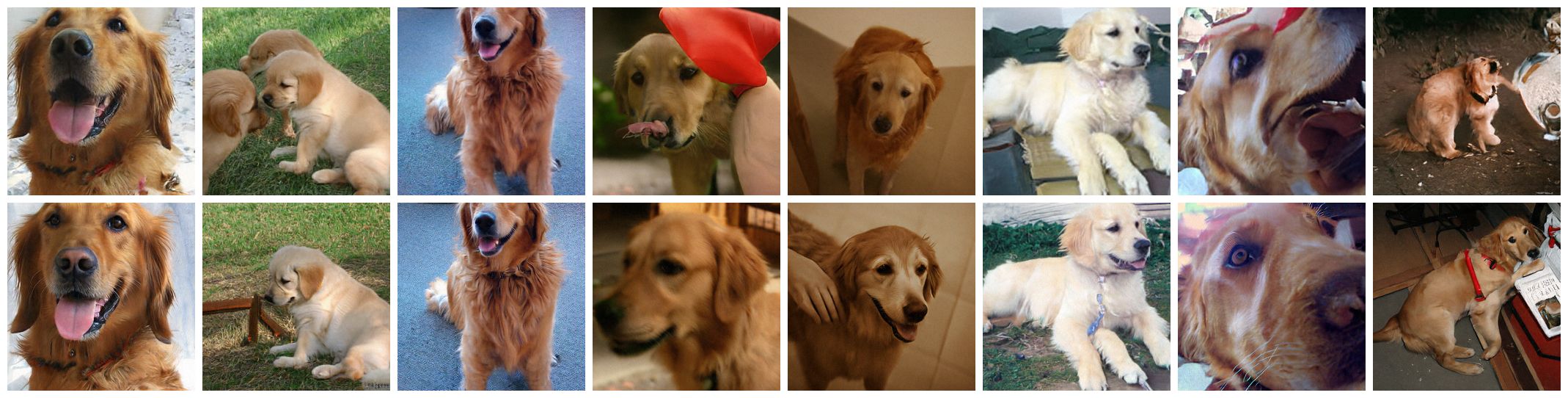}
        \caption{Class 207: golden retriever.}
    \end{subfigure}
    \begin{subfigure}[t]{\linewidth}
        \includegraphics[width=\linewidth]{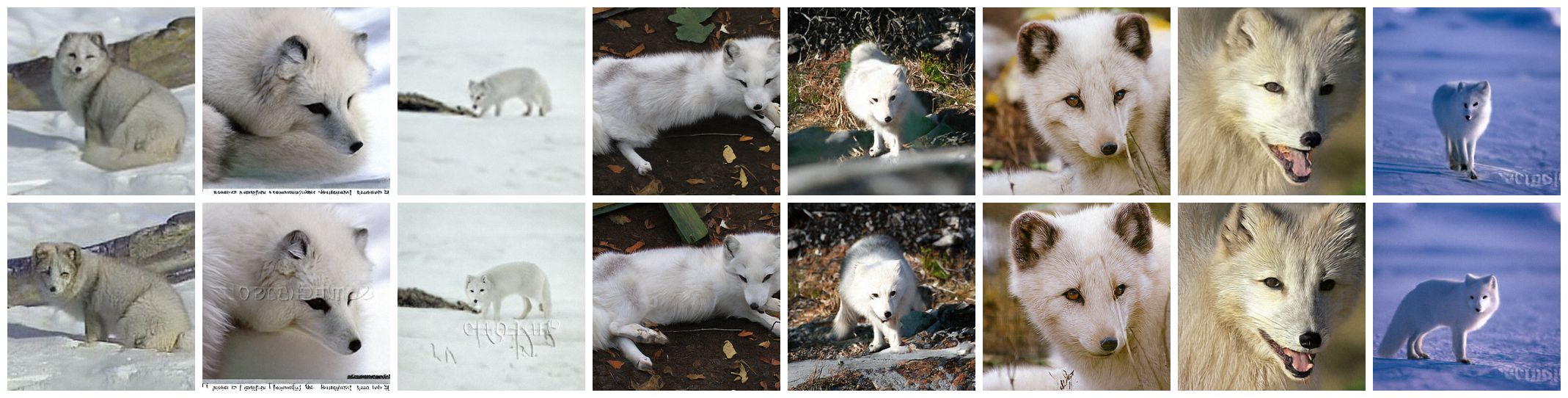}
        \caption{Class 279: white fox.}
    \end{subfigure}
    \caption{
        JiT-H/16 guided, pixel-space ImageNet 256px generation. \\
        Top is FM (CFG2.2, FID 1.86). Bottom is CAFM (CFG1.8, FID 1.80). \\
        Uncurated.
    }
    \label{fig:jit_cfg}
\end{figure}

%% file: tab/t2i_hyperparams.tex
\begin{table}[H]
    \centering
    \caption{Text-to-image CAFM post-training hyperparameters}
    \label{tab:t2i_hyperparams}
    \setlength{\tabcolsep}{3pt}
    \scriptsize
    \begin{tabular}{r|cc}
        \toprule
         & FM & CAFM \\
        \midrule
        $G$ learning rate & 5e-5 & 5e-5 \\
        $D$ learning rate &  & 3e-5 \\
        $N$ & & 16 \\
        $\lambda_\mathrm{ot}$ & & 0 \\
        AdamW $\beta$ & (0.9, 0.95) & (0, 0.95) \\
        iterations & 30k & FM10k+20k \\
        resolution & \multicolumn{2}{c}{512px} \\
        batch size & \multicolumn{2}{c}{1024} \\
        weight decay & \multicolumn{2}{c}{0.01} \\
        timesteps & \multicolumn{2}{c}{uniform(0,1), shift 3} \\
        sampler & \multicolumn{2}{c}{Euler 50 steps, shift 6} \\
        CFG dropout & \multicolumn{2}{c}{0.1} \\
        CFG scale & \multicolumn{2}{c}{4 when used} \\
        precision & \multicolumn{2}{c}{BF16} \\
        \bottomrule
    \end{tabular}
    
\end{table}

%% file: tab/geneval_full.tex
\begin{table}[h]
    \centering
    \caption{
    GenEval~\cite{ghosh2023geneval} on 512px T2I generation including original ZImage.\\
    \vspace{1pt}
    \\
    \scriptsize{*Trained on different datasets. Not directly comparable.}
    }
    \label{tab:geneval_full}
    \scriptsize
    \setlength{\tabcolsep}{2pt}
    \begin{tabular}{l|cc|cccccc|c}
        \toprule
        Method & PE & CFG & Single Obj. & Two Obj. & Color Attr. & Position & 
        Counting & Colors. & Overall \\
        \midrule
        \textcolor[gray]{0.7}{ZImage} & \multirow{3}{*}{No} & \multirow{3}{*}{No} & \textcolor[gray]{0.7}{0.78} & \textcolor[gray]{0.7}{0.37} & \textcolor[gray]{0.7}{0.16} & \textcolor[gray]{0.7}{0.11} & \textcolor[gray]{0.7}{0.32} & \textcolor[gray]{0.7}{0.59} & \textcolor[gray]{0.7}{0.39*} \\
        ZImage+FM & & & 0.72 & 0.23 & 0.11 & 0.09 & 0.25 & 0.59 & 0.33\phantom{*} \\
        ZImage+CAFM & & & 0.85 & 0.42 & 0.17 & 0.16 & 0.41 & 0.61 & 0.44\phantom{*} \\
        \midrule
        \textcolor[gray]{0.7}{ZImage} & \multirow{3}{*}{Yes} & \multirow{3}{*}{No} & \textcolor[gray]{0.7}{0.96} & \textcolor[gray]{0.7}{0.83} & \textcolor[gray]{0.7}{0.43} & \textcolor[gray]{0.7}{0.47} & \textcolor[gray]{0.7}{0.51} & \textcolor[gray]{0.7}{0.85} & \textcolor[gray]{0.7}{0.68*} \\
        ZImage+FM & & & 0.95 & 0.66 & 0.35 & 0.40 & 0.42 & 0.81 & 0.60\phantom{*} \\
        ZImage+CAFM & & & 0.99 & 0.83 & 0.50 & 0.52 & 0.57 & 0.86 & 0.71\phantom{*} \\
        \midrule
        \textcolor[gray]{0.7}{ZImage} & \multirow{3}{*}{Yes} & \multirow{3}{*}{Yes} & \textcolor[gray]{0.7}{0.98} & \textcolor[gray]{0.7}{0.95} & \textcolor[gray]{0.7}{0.70} & \textcolor[gray]{0.7}{0.67} & \textcolor[gray]{0.7}{0.67} & \textcolor[gray]{0.7}{0.95} & \textcolor[gray]{0.7}{0.82*} \\
        ZImage+FM & & & 0.99 & 0.89 & 0.62 & 0.69 & 0.77 & 0.89 & 0.81\phantom{*} \\
        ZImage+CAFM & & & 0.99 & 0.92 & 0.71 & 0.71 & 0.81 & 0.94 & 0.85\phantom{*} \\
        \bottomrule         
    \end{tabular}
    
\end{table}

%% file: tab/dpg_full.tex
\begin{table}[h]
    \centering
    \caption{
    DPG-Bench~\cite{hu2024ella} on 512px T2I generation including original ZImage.\\
    \vspace{1pt}
    \\
    \scriptsize{*Trained on different datasets. Not directly comparable.}
    }
    \label{tab:dpg_full}
    \scriptsize
    \setlength{\tabcolsep}{7.5pt}
    \begin{tabular}{l|c|ccccc|c}
        \toprule
        Method & CFG & Global & Entity & Attribute & Relation & Other & Overall \\
        \midrule
        \textcolor[gray]{0.7}{ZImage} & \multirow{3}{*}{No} & \textcolor[gray]{0.7}{87.27} & \textcolor[gray]{0.7}{87.22} & \textcolor[gray]{0.7}{86.44} & \textcolor[gray]{0.7}{88.97} & \textcolor[gray]{0.7}{89.21} & \textcolor[gray]{0.7}{79.83*} \\
        ZImage+FM & & 81.34 & 82.96 & 81.71 & 83.17 & 85.07 & 72.25\phantom{*} \\
        ZImage+CAFM & & 87.82 & 86.65 & 86.33 & 86.49 & 84.85 & 77.21\phantom{*} \\
        \midrule
        \textcolor[gray]{0.7}{ZImage} & \multirow{3}{*}{Yes} & \textcolor[gray]{0.7}{90.55} & \textcolor[gray]{0.7}{91.71} & \textcolor[gray]{0.7}{91.49} & \textcolor[gray]{0.7}{92.43} & \textcolor[gray]{0.7}{87.94} & \textcolor[gray]{0.7}{86.35*} \\
        ZImage+FM & & 90.34 & 90.56 & 88.98 & 88.17 & 90.71 & 83.67\phantom{*} \\
        ZImage+CAFM & & 89.55 & 89.83 & 89.99 & 91.20 & 91.88 & 85.21\phantom{*} \\
        \bottomrule
    \end{tabular}
\end{table}

%% file: fig/t2i_dpg1.tex
\begin{figure}
    \centering
    \captionsetup{justification=raggedright,singlelinecheck=false}
    \setlength{\tabcolsep}{5pt}
    \scriptsize
    \begin{tabularx}{\linewidth}{|X|X|X|X|}
        FM & CAFM & FM+CFG4 & CAFM+CFG4    
    \end{tabularx}
    \setlength{\tabcolsep}{1pt}
    \begin{subfigure}[t]{1\linewidth}
        \begin{tabularx}{\linewidth}{XXXX}
            \includegraphics[width=\linewidth]{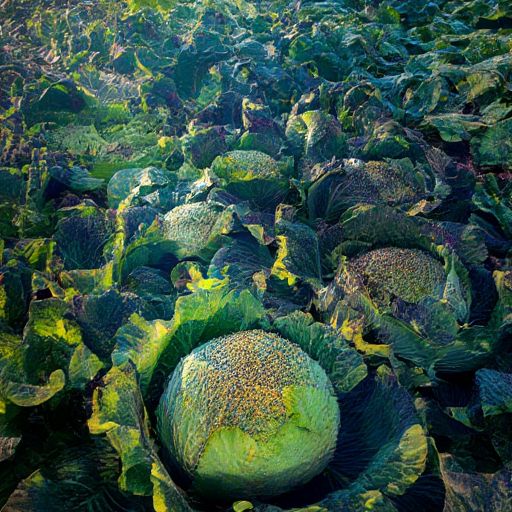} &
            \includegraphics[width=\linewidth]{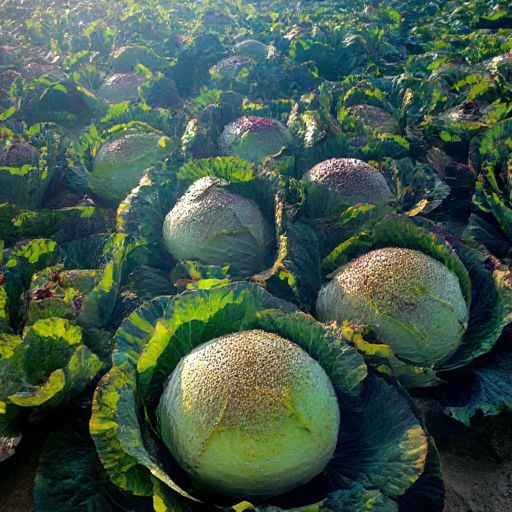} &
            \includegraphics[width=\linewidth]{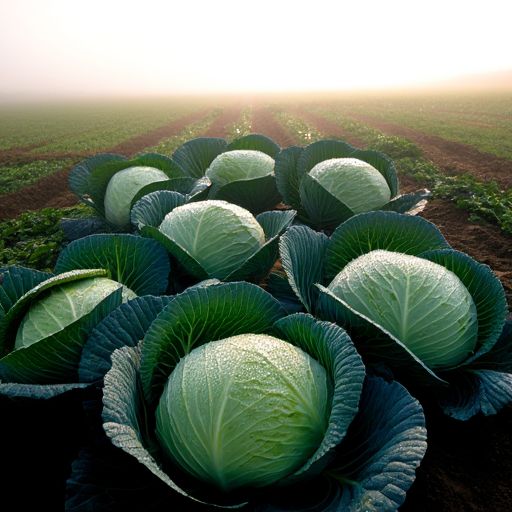} &
            \includegraphics[width=\linewidth]{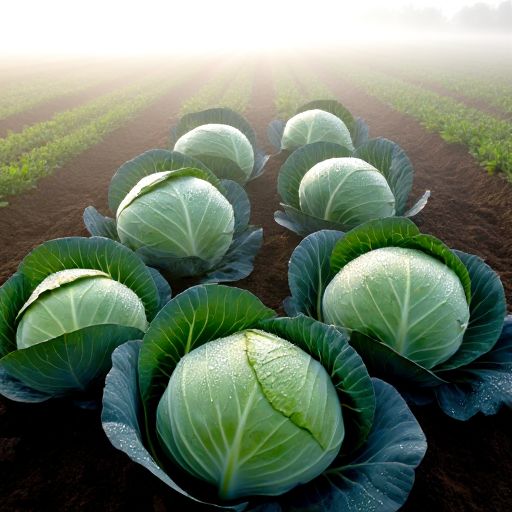}
        \end{tabularx}
        \vspace{-4pt}
        \caption{\textbf{DPG prompt 0:} Eight cabbages in a dewy morning field}
        \vspace{4pt}
    \end{subfigure}
    
    \begin{subfigure}[t]{1\linewidth}
        \begin{tabularx}{\linewidth}{XXXX}
            \includegraphics[width=\linewidth]{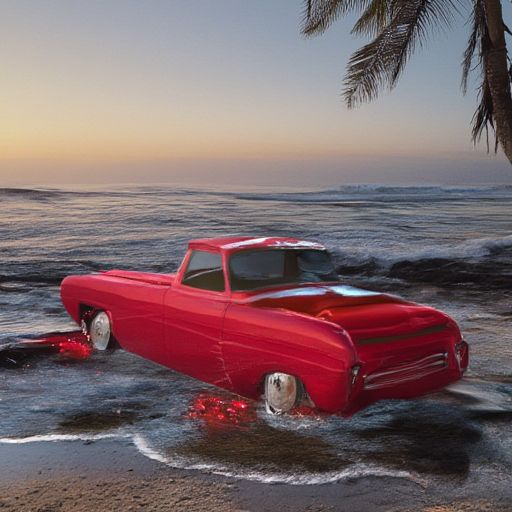} &
            \includegraphics[width=\linewidth]{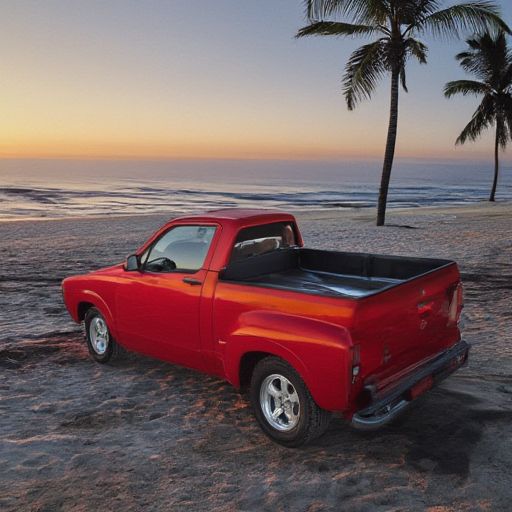} &
            \includegraphics[width=\linewidth]{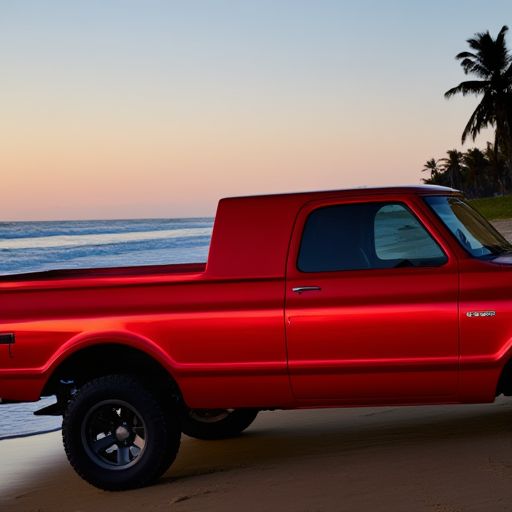} &
            \includegraphics[width=\linewidth]{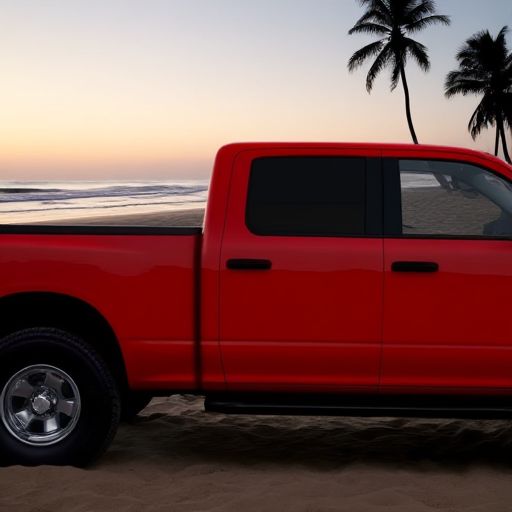}
        \end{tabularx}
        \vspace{-4pt}
        \caption{\textbf{DPG prompt 1:} A red pickup truck parked on a beach at dusk, palm trees in the back}
        \vspace{4pt}
    \end{subfigure}

    \begin{subfigure}[t]{1\linewidth}
        \begin{tabularx}{\linewidth}{XXXX}
            \includegraphics[width=\linewidth]{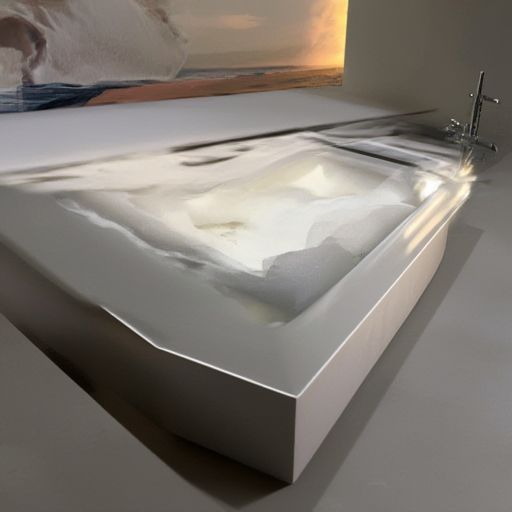} &
            \includegraphics[width=\linewidth]{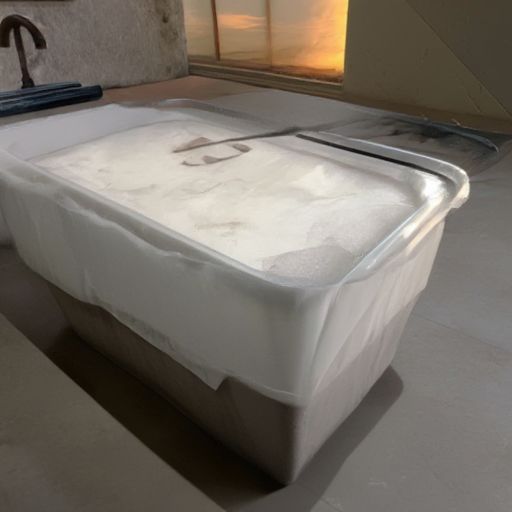} &
            \includegraphics[width=\linewidth]{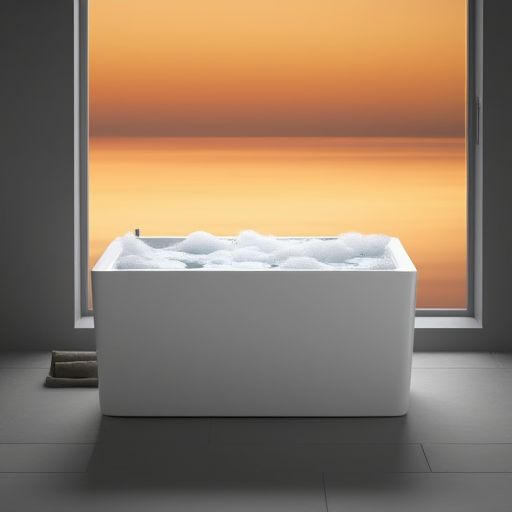} &
            \includegraphics[width=\linewidth]{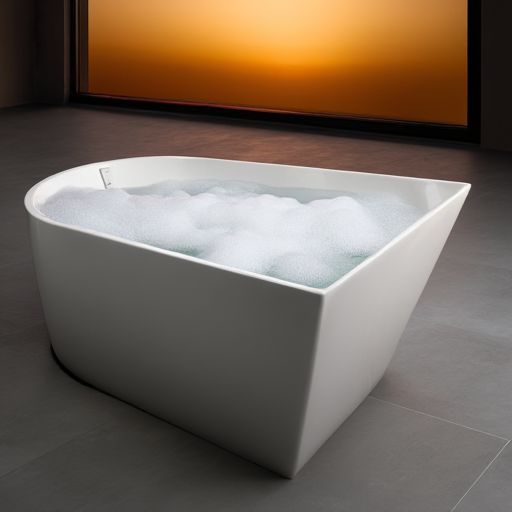}
        \end{tabularx}
        \vspace{-4pt}
        \caption{\textbf{DPG prompt 2:} A bathroom with a white rectangular bathtub full of bubbles}
        \vspace{4pt}
    \end{subfigure}

    \begin{subfigure}[t]{1\linewidth}
        \begin{tabularx}{\linewidth}{XXXX}
            \includegraphics[width=\linewidth]{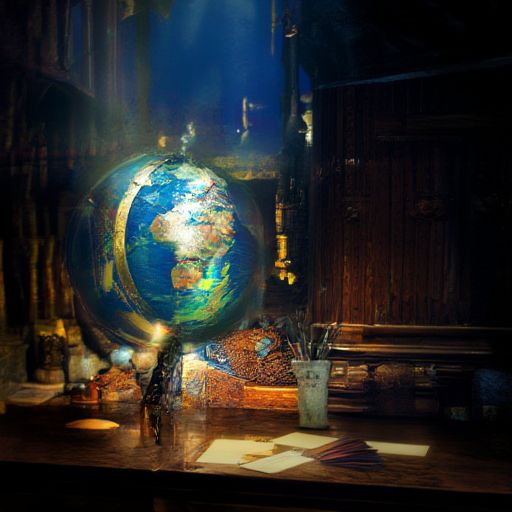} &
            \includegraphics[width=\linewidth]{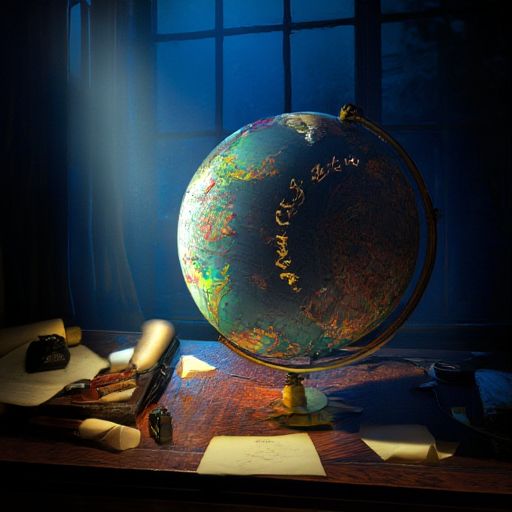} &
            \includegraphics[width=\linewidth]{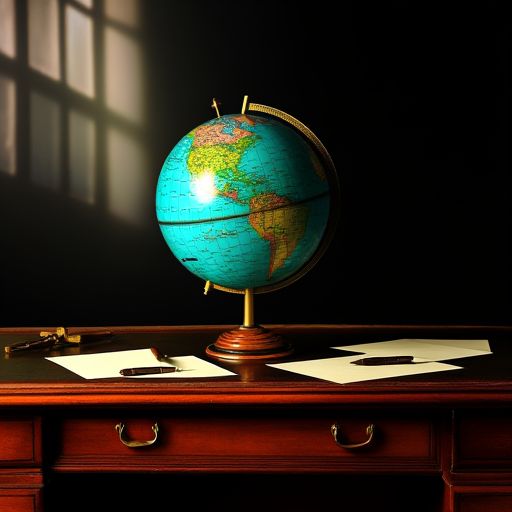} &
            \includegraphics[width=\linewidth]{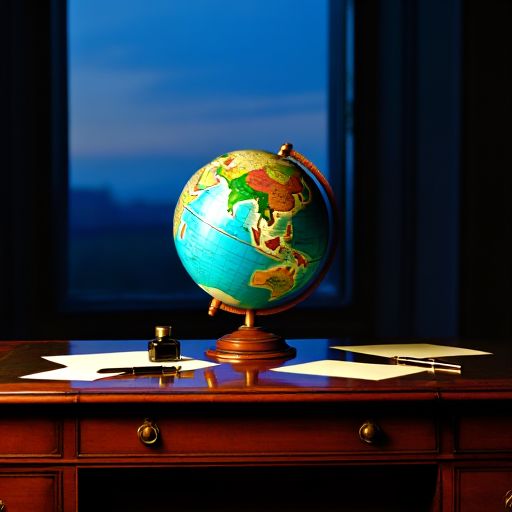}
        \end{tabularx}
        \vspace{-4pt}
        \caption{\textbf{DPG prompt 3:} An antique mahogany desk with a detailed globe and vintage pens}
        \vspace{4pt}
    \end{subfigure}

    \begin{subfigure}[t]{1\linewidth}
        \begin{tabularx}{\linewidth}{XXXX}
            \includegraphics[width=\linewidth]{img/t2i_dpg/dpg5_fm.jpg} &
            \includegraphics[width=\linewidth]{img/t2i_dpg/dpg5_ca.jpg} &
            \includegraphics[width=\linewidth]{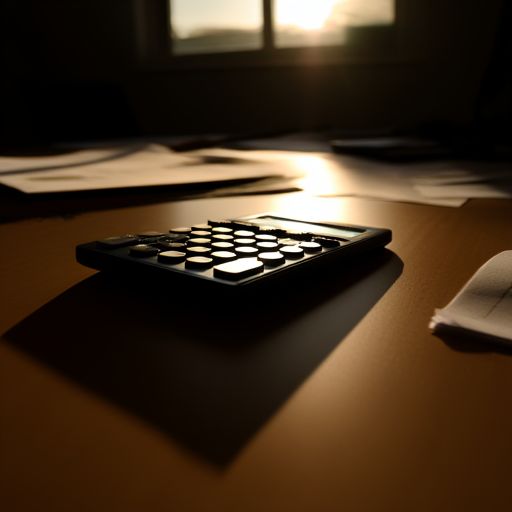} &
            \includegraphics[width=\linewidth]{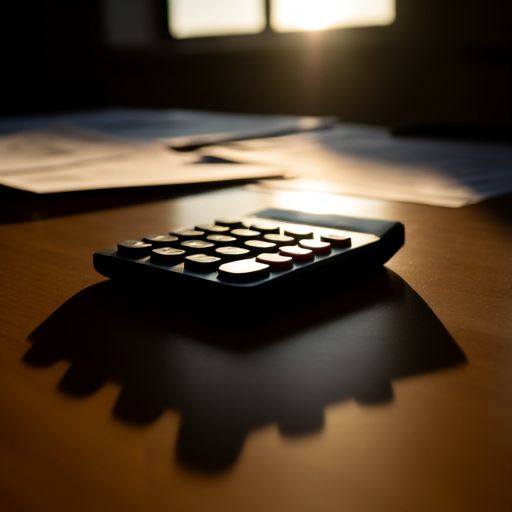}
        \end{tabularx}
        \vspace{-4pt}
        \caption{\textbf{DPG prompt 5:} A calculator on a wooden desk with scattered papers}
        \vspace{4pt}
    \end{subfigure}
    
    \captionsetup{justification=centering}
    \caption{
    Curated text-to-image comparisons on DPG benchmark prompts. \\
    Prompts are shortened for paper presentation. \\
    (part 1 of 4)
    }
    \label{fig:t2i_dpg}
\end{figure}

%% file: fig/t2i_dpg2.tex
\begin{figure}
    \ContinuedFloat
    \centering
    \captionsetup{justification=raggedright,singlelinecheck=false}
    \setlength{\tabcolsep}{5pt}
    \scriptsize
    \begin{tabularx}{\linewidth}{|X|X|X|X|}
        FM & CAFM & FM+CFG4 & CAFM+CFG4    
    \end{tabularx}
    \setlength{\tabcolsep}{1pt}
    \begin{subfigure}[t]{1\linewidth}
        \begin{tabularx}{\linewidth}{XXXX}
            \includegraphics[width=\linewidth]{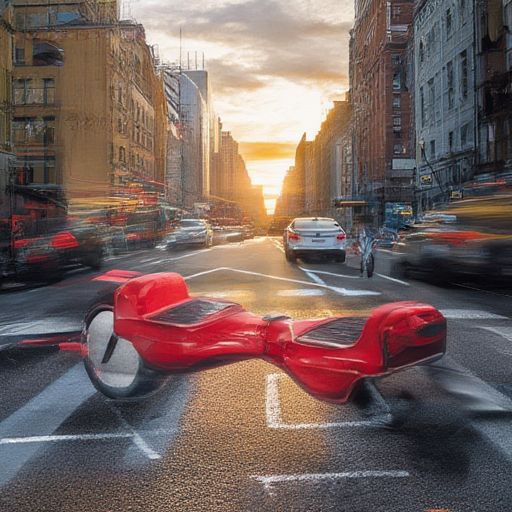} &
            \includegraphics[width=\linewidth]{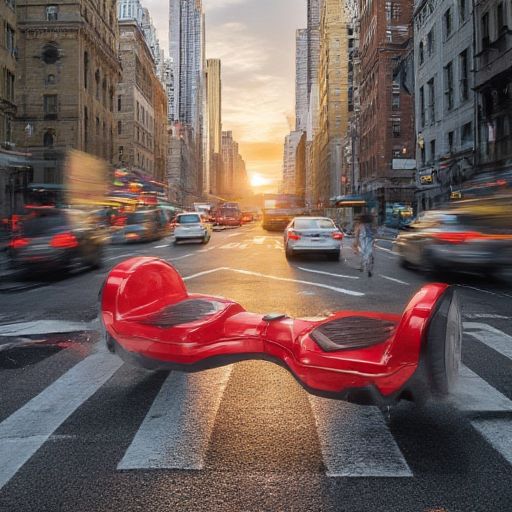} &
            \includegraphics[width=\linewidth]{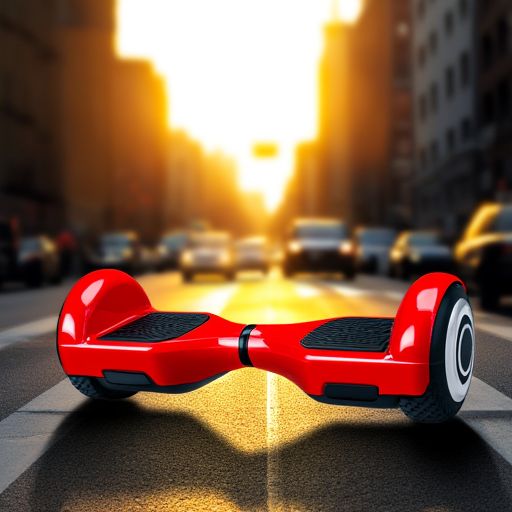} &
            \includegraphics[width=\linewidth]{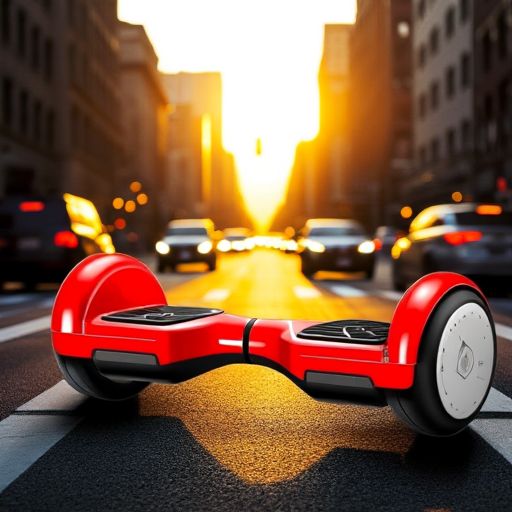}
        \end{tabularx}
        \vspace{-4pt}
        \caption{\textbf{DPG prompt 8:} A red hoverboard on a city street at sunset, surrounded by tall buildings}
        \vspace{4pt}
    \end{subfigure}
    
    \begin{subfigure}[t]{1\linewidth}
        \begin{tabularx}{\linewidth}{XXXX}
            \includegraphics[width=\linewidth]{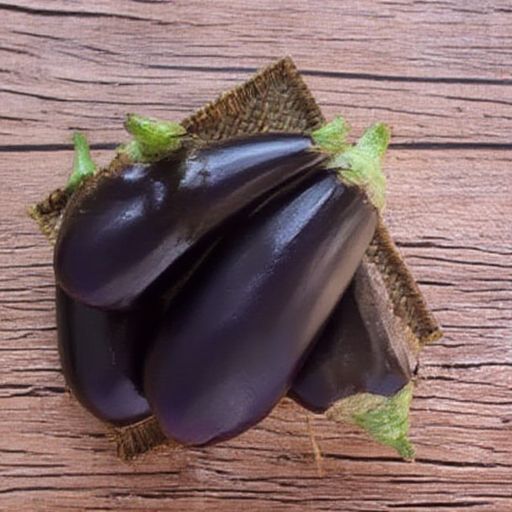} &
            \includegraphics[width=\linewidth]{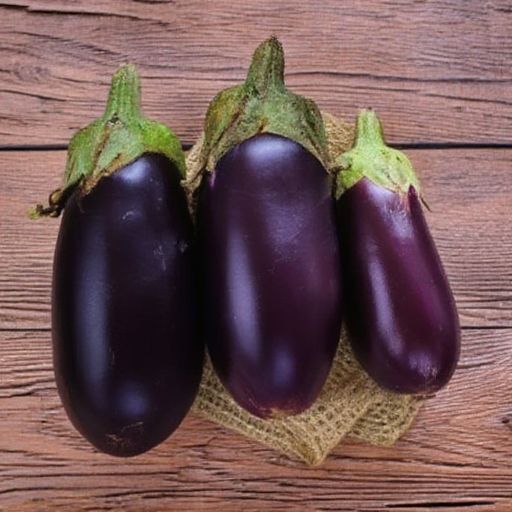} &
            \includegraphics[width=\linewidth]{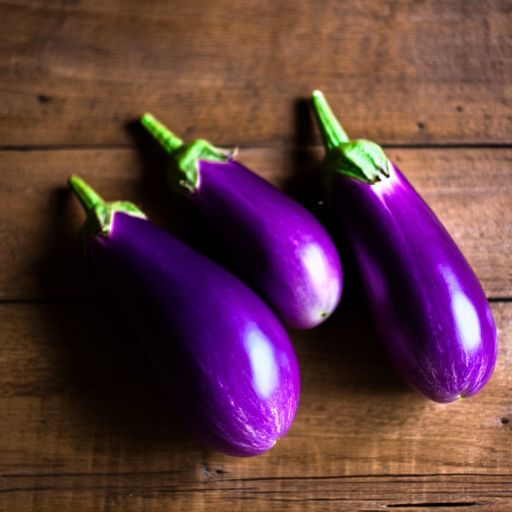} &
            \includegraphics[width=\linewidth]{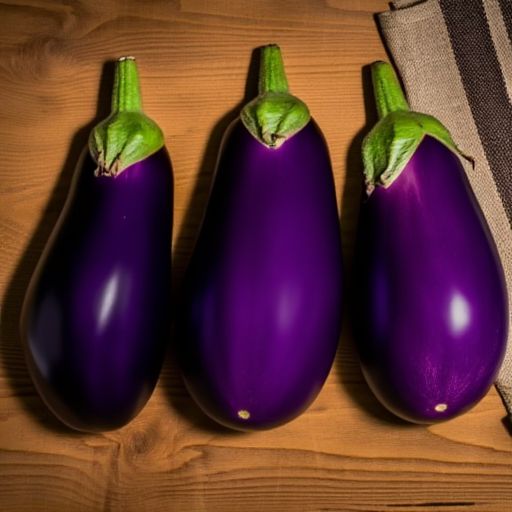}
        \end{tabularx}
        \vspace{-4pt}
        \caption{\textbf{DPG prompt 10:} Three purple eggplants on a rustic wooden table with a napkin}
        \vspace{4pt}
    \end{subfigure}

    \begin{subfigure}[t]{1\linewidth}
        \begin{tabularx}{\linewidth}{XXXX}
            \includegraphics[width=\linewidth]{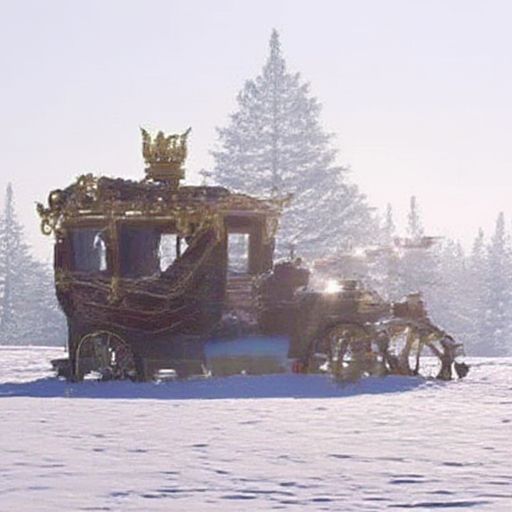} &
            \includegraphics[width=\linewidth]{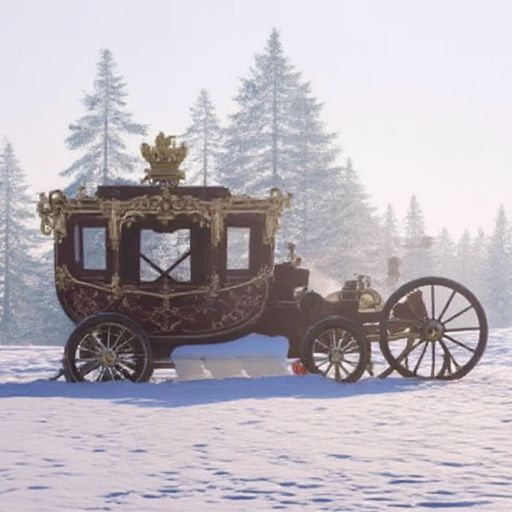} &
            \includegraphics[width=\linewidth]{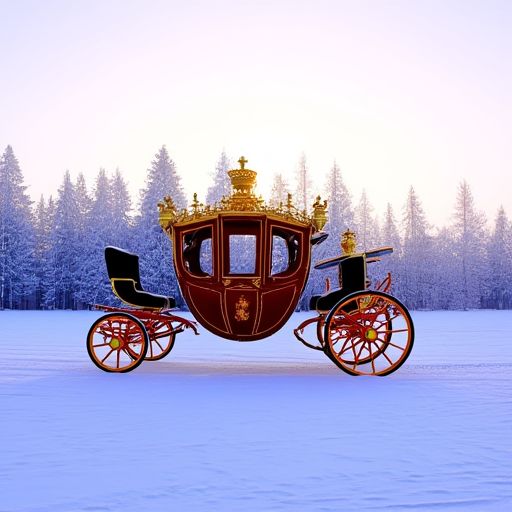} &
            \includegraphics[width=\linewidth]{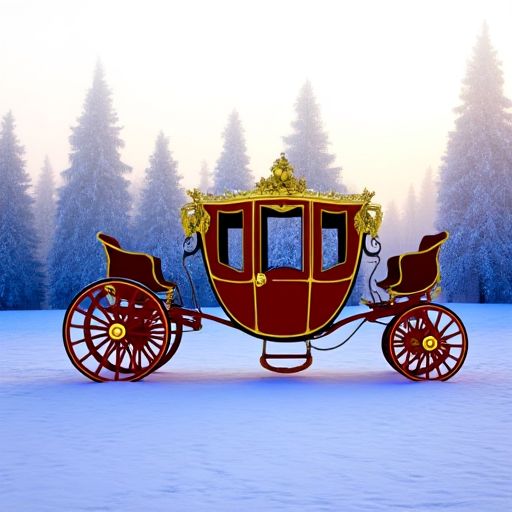}
        \end{tabularx}
        \vspace{-4pt}
        \caption{\textbf{DPG prompt 20:} A red and gold royal carriage in a snowy landscape with pine trees}
        \vspace{4pt}
    \end{subfigure}

    \begin{subfigure}[t]{1\linewidth}
        \begin{tabularx}{\linewidth}{XXXX}
            \includegraphics[width=\linewidth]{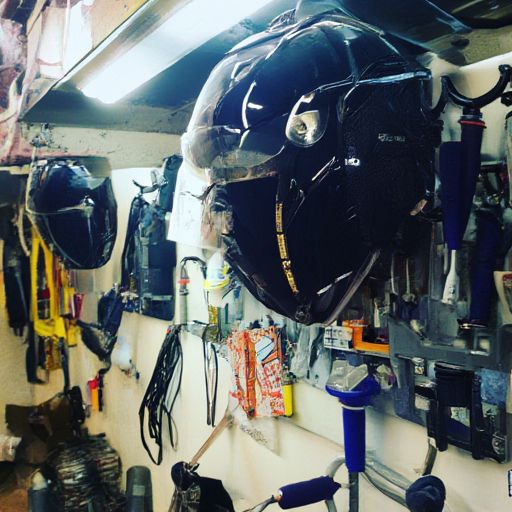} &
            \includegraphics[width=\linewidth]{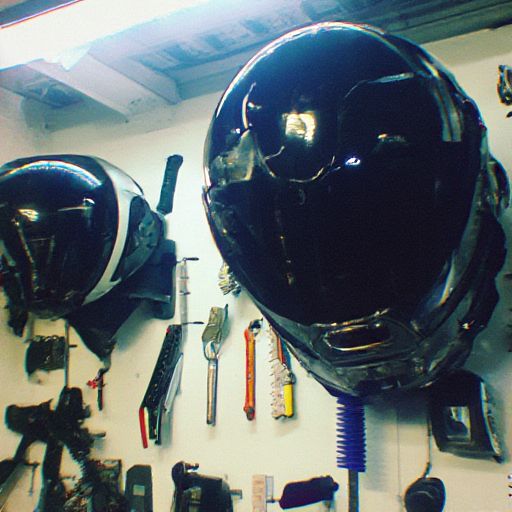} &
            \includegraphics[width=\linewidth]{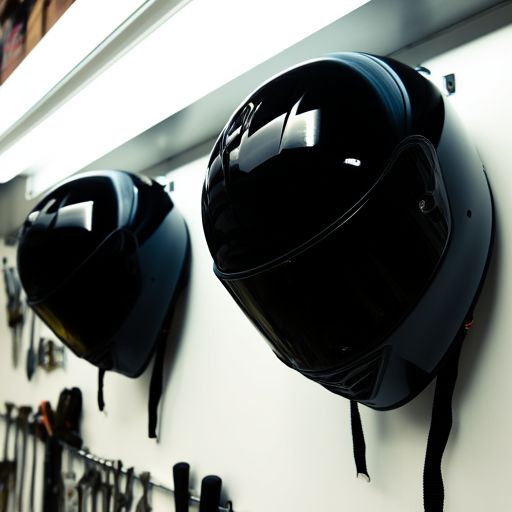} &
            \includegraphics[width=\linewidth]{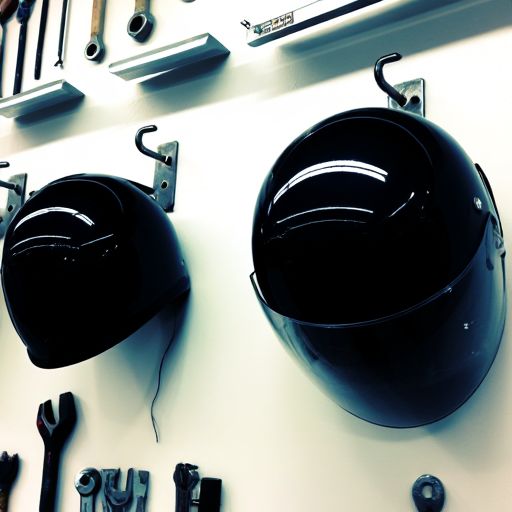}
        \end{tabularx}
        \vspace{-4pt}
        \caption{\textbf{DPG prompt 40: } Two black motorcycle helmets hanging on a white wall with tools}
        \vspace{4pt}
    \end{subfigure}

    \begin{subfigure}[t]{1\linewidth}
        \begin{tabularx}{\linewidth}{XXXX}
            \includegraphics[width=\linewidth]{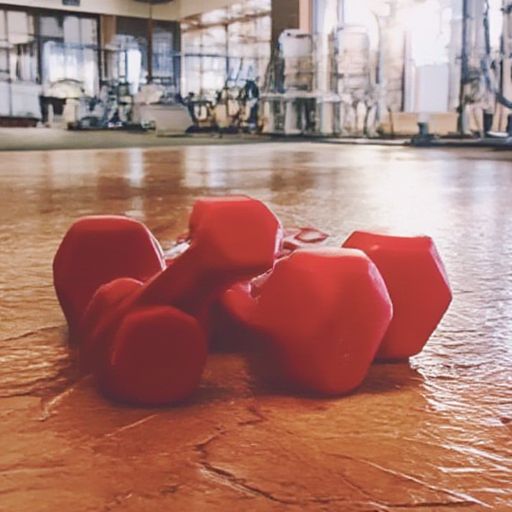} &
            \includegraphics[width=\linewidth]{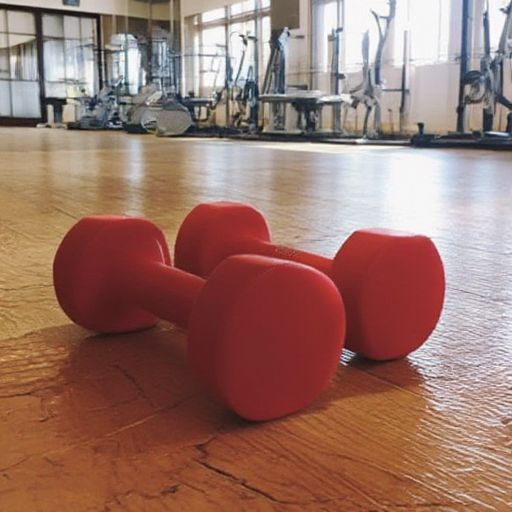} &
            \includegraphics[width=\linewidth]{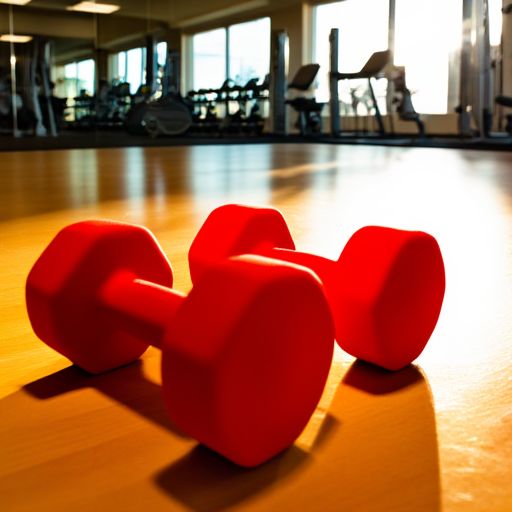} &
            \includegraphics[width=\linewidth]{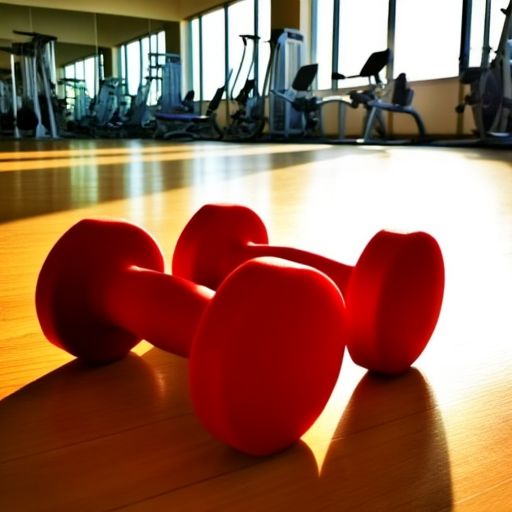}
        \end{tabularx}
        \vspace{-4pt}
        \caption{\textbf{DPG prompt 50: } Three red dumbbells on a wooden gym floor}
        \vspace{4pt}
    \end{subfigure}
    
    \captionsetup{justification=centering}
    \caption{
    Curated text-to-image comparisons on DPG benchmark prompts. \\
    Prompts are shortened for paper presentation. \\
    (part 2 of 4)
    }
    \label{fig:t2i_dpg}
\end{figure}

%% file: fig/t2i_dpg3.tex
\begin{figure}
    \ContinuedFloat
    \centering
    \captionsetup{justification=raggedright,singlelinecheck=false}
    \setlength{\tabcolsep}{5pt}
    \scriptsize
    \begin{tabularx}{\linewidth}{|X|X|X|X|}
        FM & CAFM & FM+CFG4 & CAFM+CFG4    
    \end{tabularx}
    \setlength{\tabcolsep}{1pt}
    \begin{subfigure}[t]{1\linewidth}
        \begin{tabularx}{\linewidth}{XXXX}
            \includegraphics[width=\linewidth]{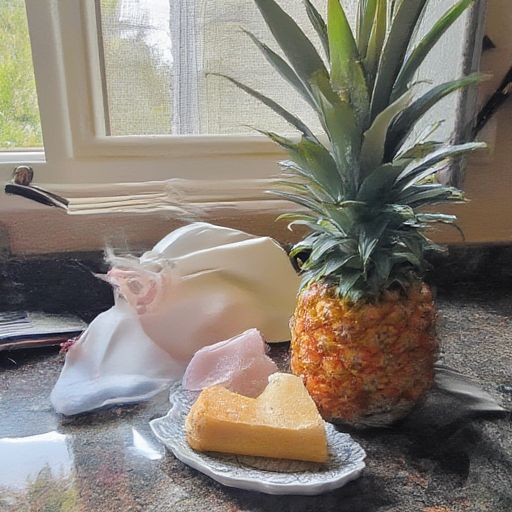} &
            \includegraphics[width=\linewidth]{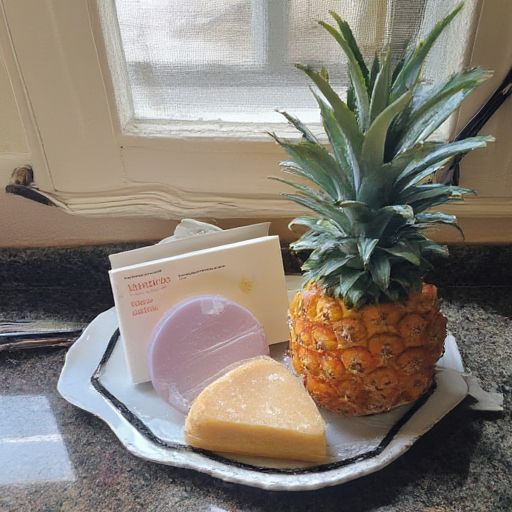} &
            \includegraphics[width=\linewidth]{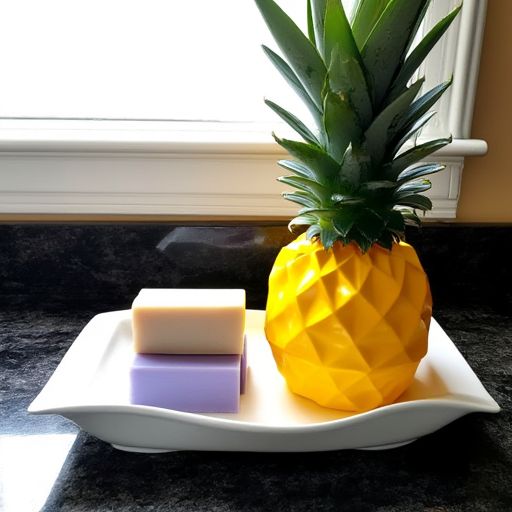} &
            \includegraphics[width=\linewidth]{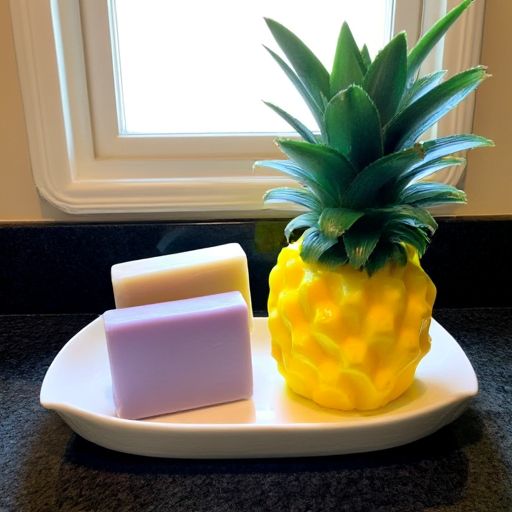}
        \end{tabularx}
        \vspace{-4pt}
        \caption{\textbf{DPG prompt 80:} A lavender and an oatmeal soap beside a yellow pineapple on a white dish}
        \vspace{4pt}
    \end{subfigure}
    
    \begin{subfigure}[t]{1\linewidth}
        \begin{tabularx}{\linewidth}{XXXX}
            \includegraphics[width=\linewidth]{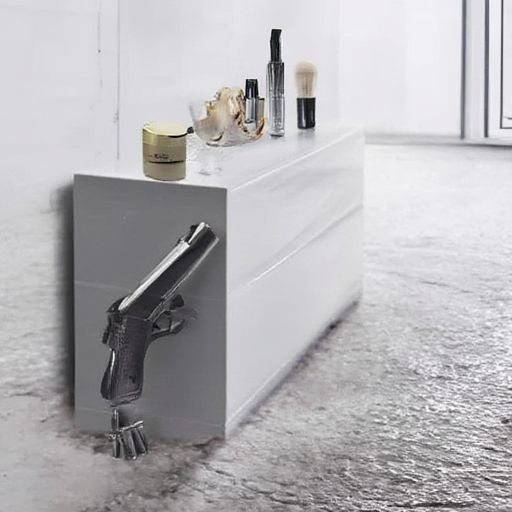} &
            \includegraphics[width=\linewidth]{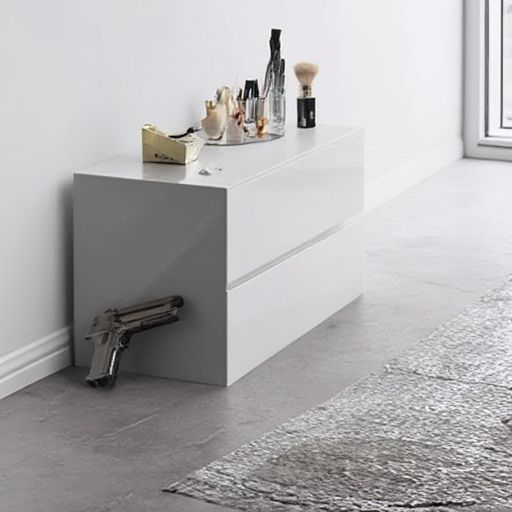} &
            \includegraphics[width=\linewidth]{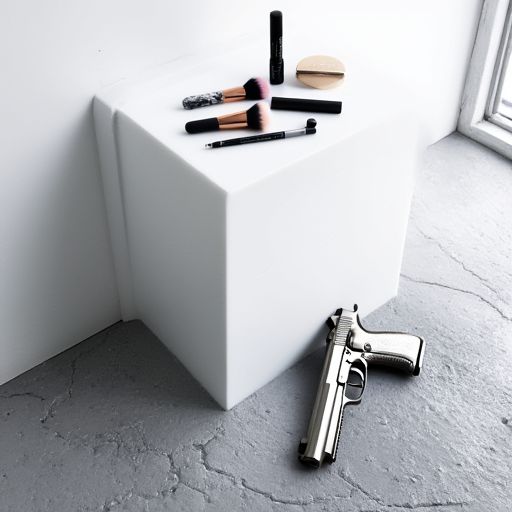} &
            \includegraphics[width=\linewidth]{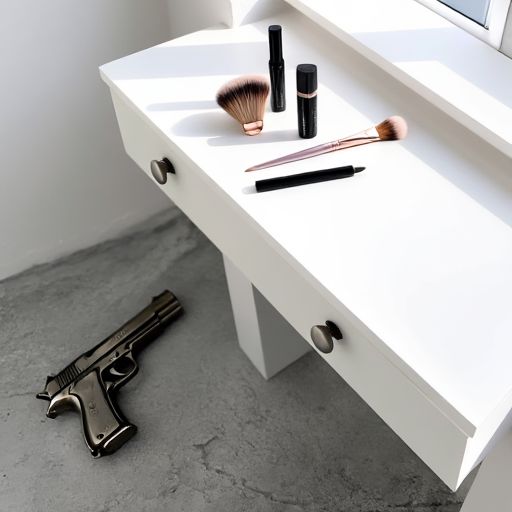}
        \end{tabularx}
        \vspace{-4pt}
        \caption{\textbf{DPG prompt 100:} A white desk with beauty products next to a handgun on the floor}
        \vspace{4pt}
    \end{subfigure}

    \begin{subfigure}[t]{1\linewidth}
        \begin{tabularx}{\linewidth}{XXXX}
            \includegraphics[width=\linewidth]{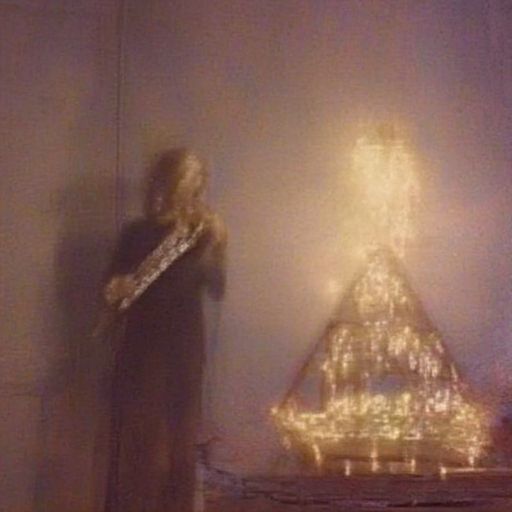} &
            \includegraphics[width=\linewidth]{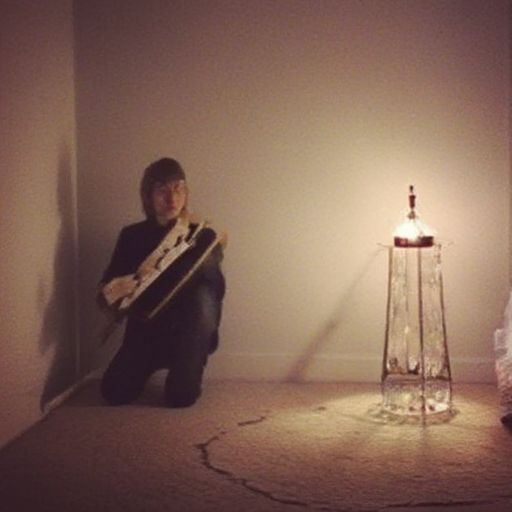} &
            \includegraphics[width=\linewidth]{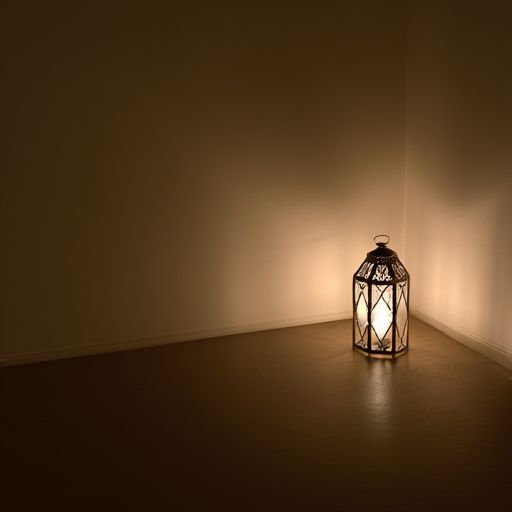} &
            \includegraphics[width=\linewidth]{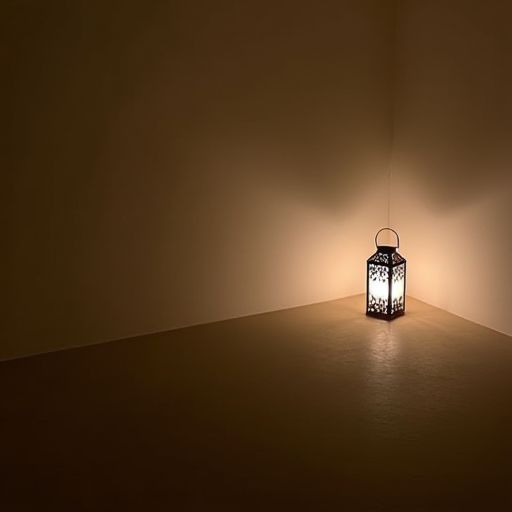}
        \end{tabularx}
        \vspace{-4pt}
        \caption{\textbf{DPG prompt 120:} A musician playing a recorder in a quiet room, lit by a glowing lantern}
        \vspace{4pt}
    \end{subfigure}

    \begin{subfigure}[t]{1\linewidth}
        \begin{tabularx}{\linewidth}{XXXX}
            \includegraphics[width=\linewidth]{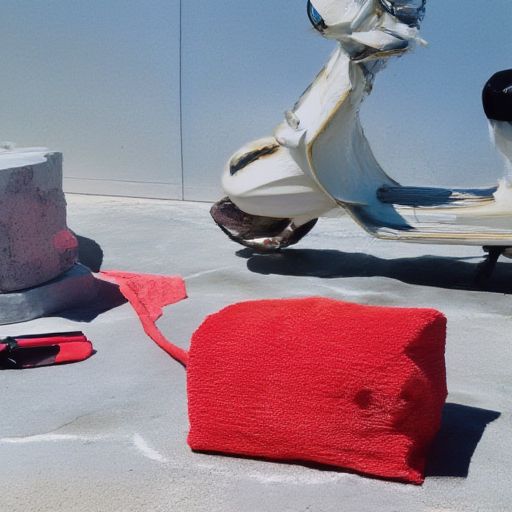} &
            \includegraphics[width=\linewidth]{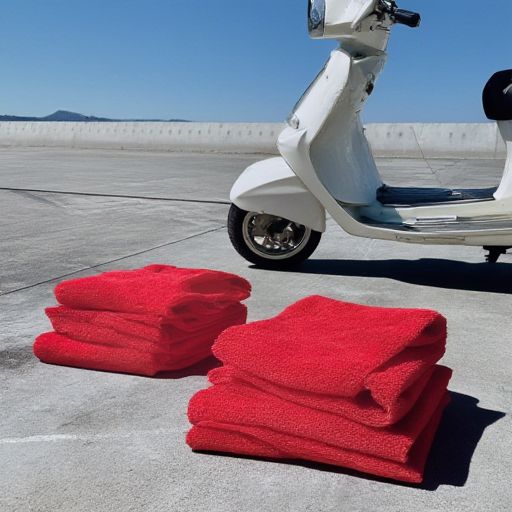} &
            \includegraphics[width=\linewidth]{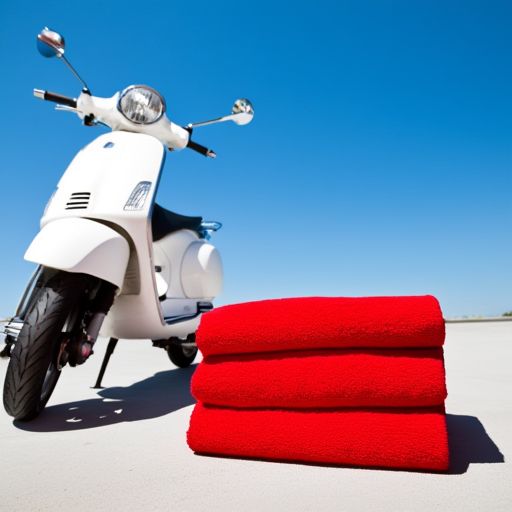} &
            \includegraphics[width=\linewidth]{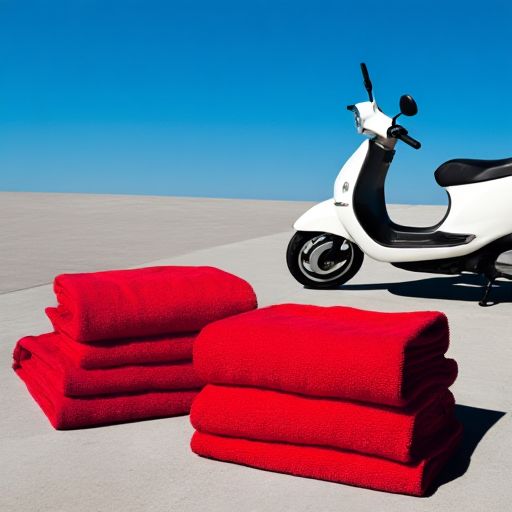}
        \end{tabularx}
        \vspace{-4pt}
        \caption{\textbf{DPG prompt 140:} Three folded red towels on concrete beside a white scooter under sky}
        \vspace{4pt}
    \end{subfigure}

    \begin{subfigure}[t]{1\linewidth}
        \begin{tabularx}{\linewidth}{XXXX}
            \includegraphics[width=\linewidth]{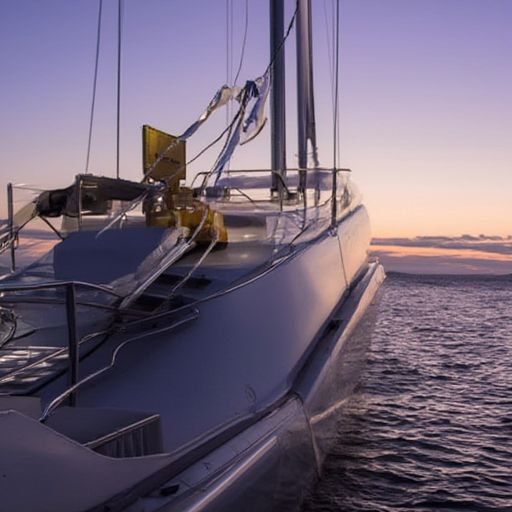} &
            \includegraphics[width=\linewidth]{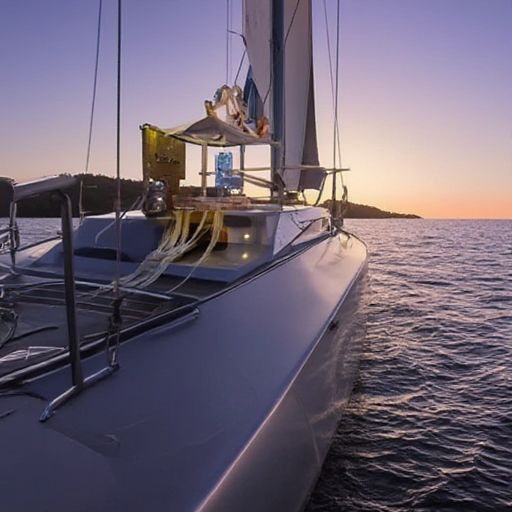} &
            \includegraphics[width=\linewidth]{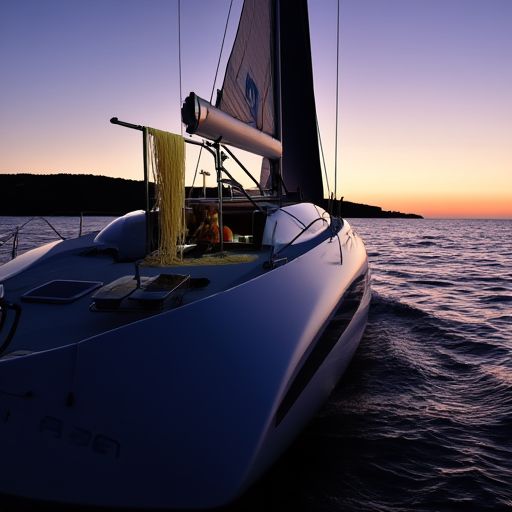} &
            \includegraphics[width=\linewidth]{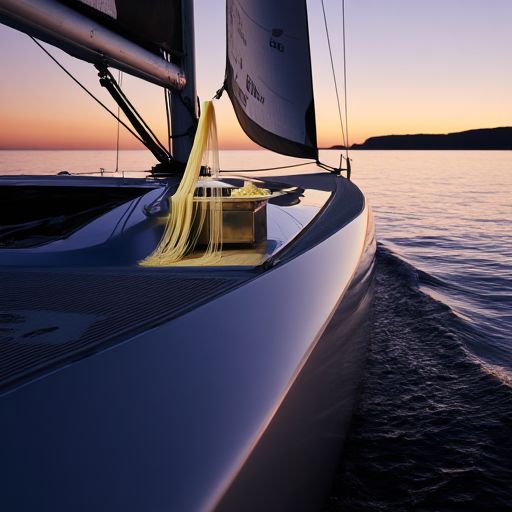}
        \end{tabularx}
        \vspace{-4pt}
        \caption{\textbf{DPG prompt 160:} A silver sailboat at twilight with someone cooking noodles on deck}
        \vspace{4pt}
    \end{subfigure}
    
    \captionsetup{justification=centering}
    \caption{
    Curated text-to-image comparisons on DPG benchmark prompts. \\
    Prompts are shortened for paper presentation. \\
    (part 3 of 4)
    }
    \label{fig:t2i_dpg}
\end{figure}

%% file: fig/t2i_dpg4.tex
\begin{figure}
    \ContinuedFloat
    \centering
    \captionsetup{justification=raggedright,singlelinecheck=false}
    \setlength{\tabcolsep}{5pt}
    \scriptsize
    \begin{tabularx}{\linewidth}{|X|X|X|X|}
        FM & CAFM & FM+CFG4 & CAFM+CFG4    
    \end{tabularx}
    \setlength{\tabcolsep}{1pt}
    \begin{subfigure}[t]{1\linewidth}
        \begin{tabularx}{\linewidth}{XXXX}
            \includegraphics[width=\linewidth]{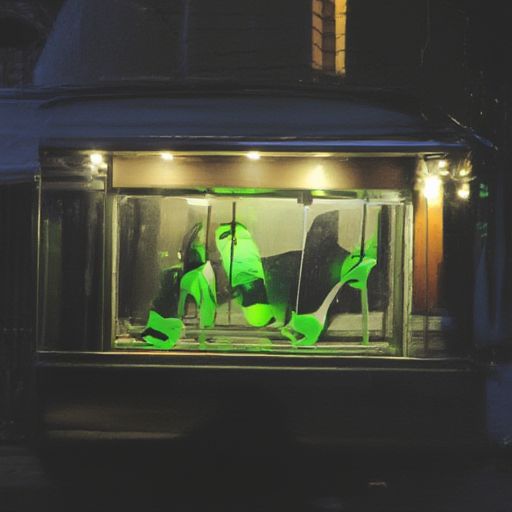} &
            \includegraphics[width=\linewidth]{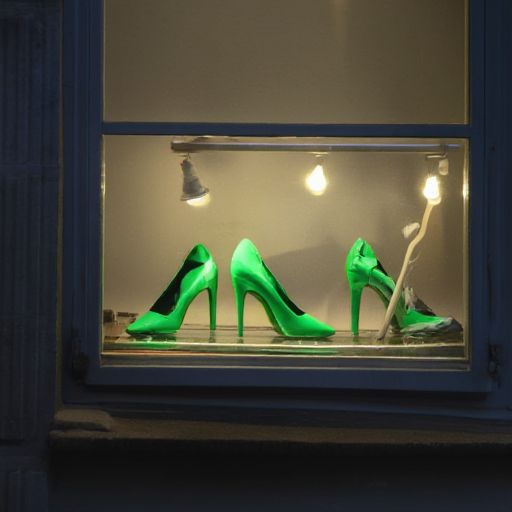} &
            \includegraphics[width=\linewidth]{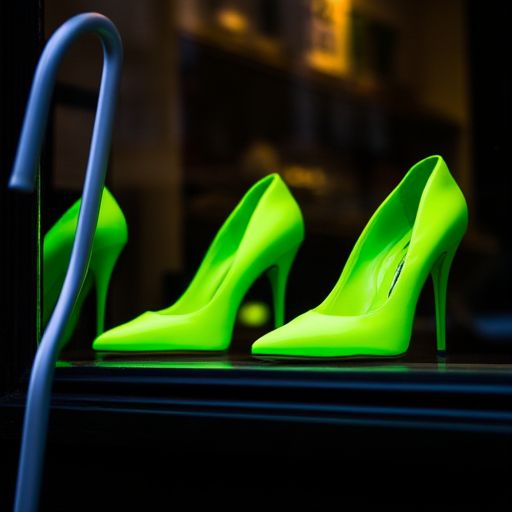} &
            \includegraphics[width=\linewidth]{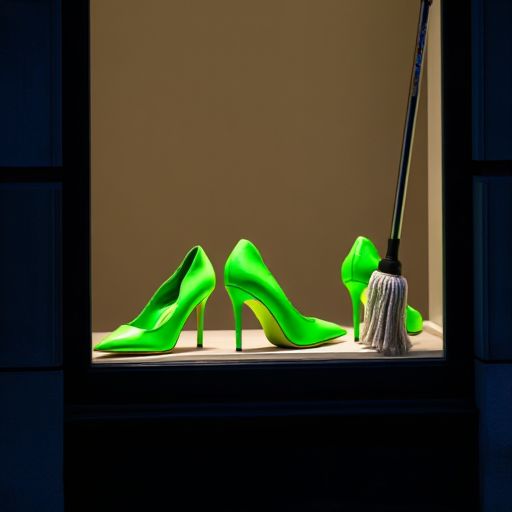}
        \end{tabularx}
        \vspace{-4pt}
        \caption{\textbf{DPG prompt 170:} Three green high heels in a storefront display. A mop leans on the window}
        \vspace{4pt}
    \end{subfigure}
    
    \begin{subfigure}[t]{1\linewidth}
        \begin{tabularx}{\linewidth}{XXXX}
            \includegraphics[width=\linewidth]{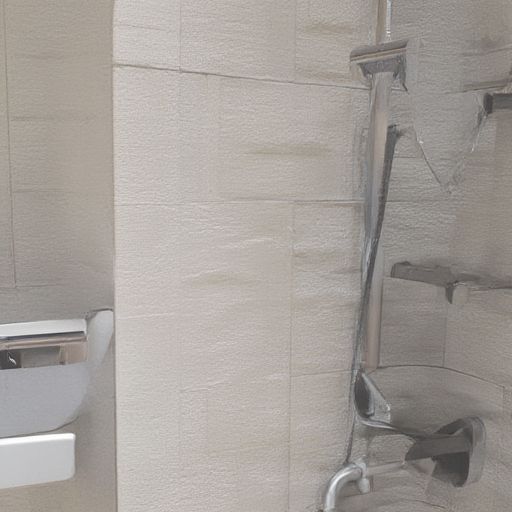} &
            \includegraphics[width=\linewidth]{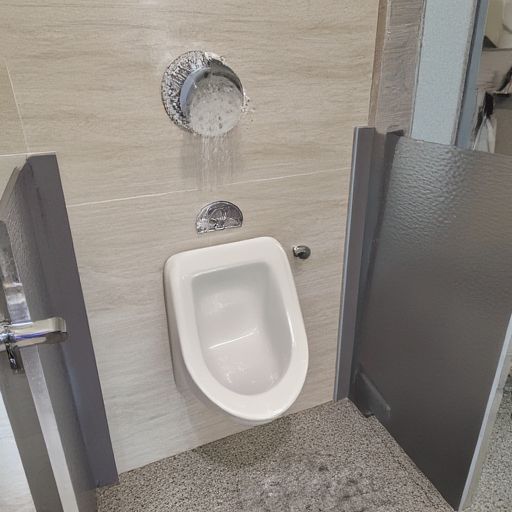} &
            \includegraphics[width=\linewidth]{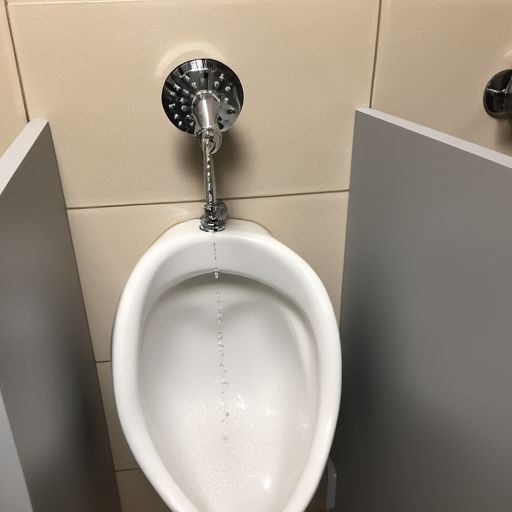} &
            \includegraphics[width=\linewidth]{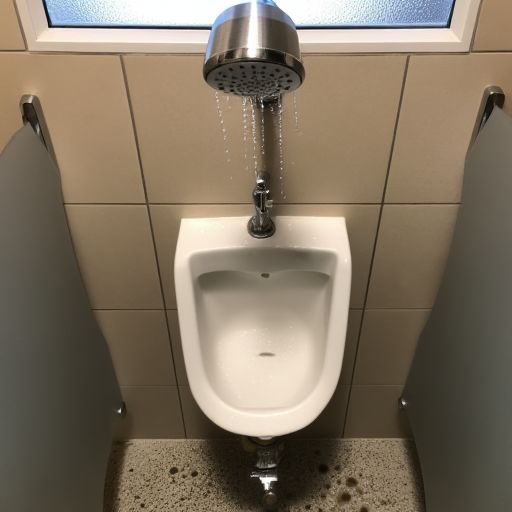}
        \end{tabularx}
        \vspace{-4pt}
        \caption{\textbf{DPG prompt 200:} A showerhead dripping onto a urinal in a restroom with grey dividers}
        \vspace{4pt}
    \end{subfigure}

    \begin{subfigure}[t]{1\linewidth}
        \begin{tabularx}{\linewidth}{XXXX}
            \includegraphics[width=\linewidth]{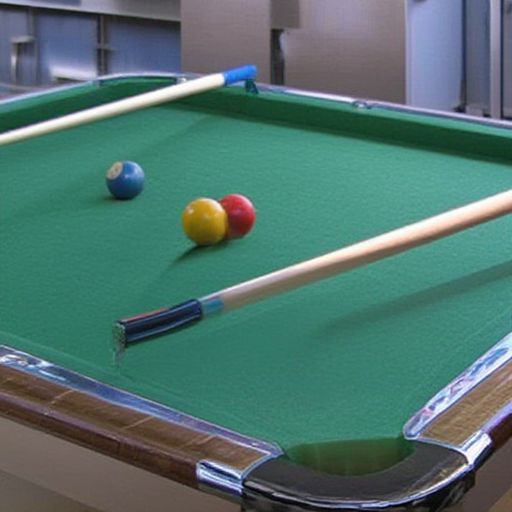} &
            \includegraphics[width=\linewidth]{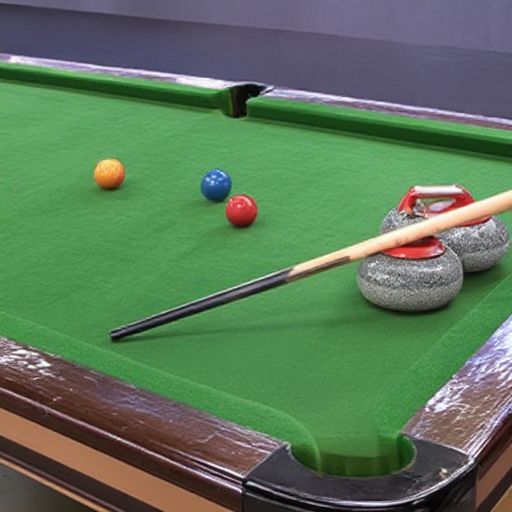} &
            \includegraphics[width=\linewidth]{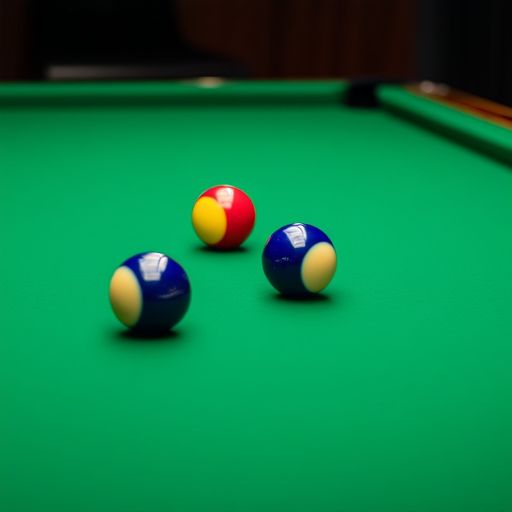} &
            \includegraphics[width=\linewidth]{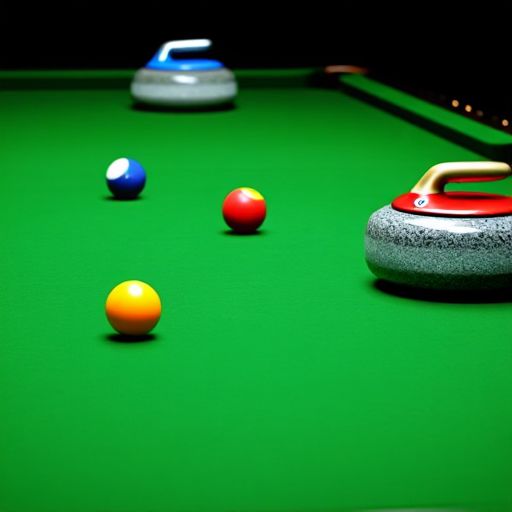}
        \end{tabularx}
        \vspace{-4pt}
        \caption{\textbf{DPG prompt 220:} Red, yellow, and blue billiard balls rolling on a table beside curling stones}
        \vspace{4pt}
    \end{subfigure}

    \begin{subfigure}[t]{1\linewidth}
        \begin{tabularx}{\linewidth}{XXXX}
            \includegraphics[width=\linewidth]{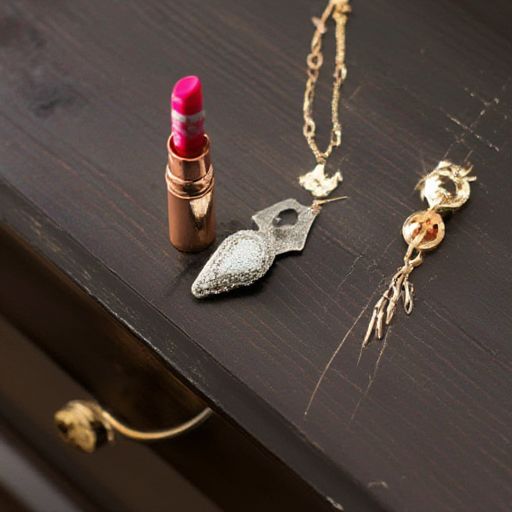} &
            \includegraphics[width=\linewidth]{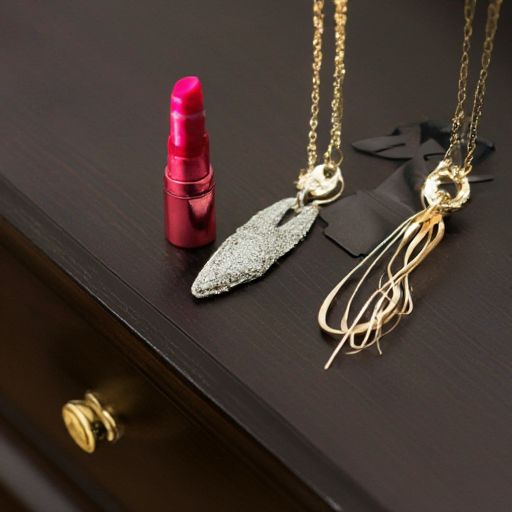} &
            \includegraphics[width=\linewidth]{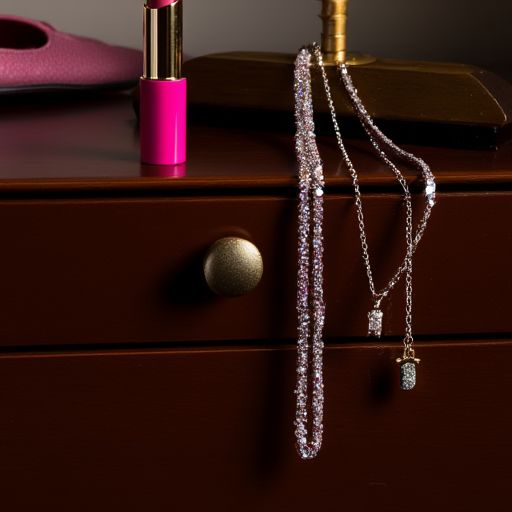} &
            \includegraphics[width=\linewidth]{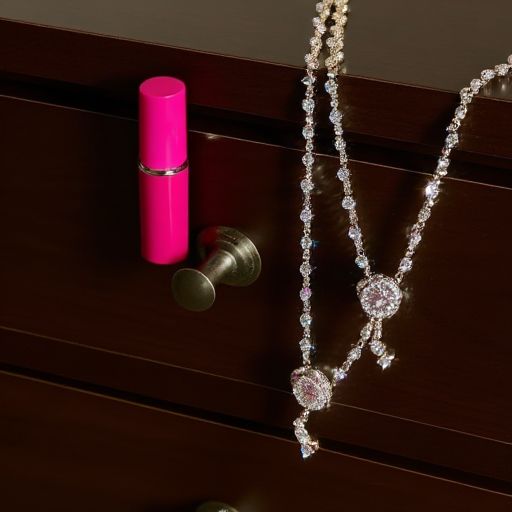}
        \end{tabularx}
        \vspace{-4pt}
        \caption{\textbf{DPG prompt 230:} A pink lipstick and two sparkling necklaces on a dark wooden dresser}
        \vspace{4pt}
    \end{subfigure}

    \begin{subfigure}[t]{1\linewidth}
        \begin{tabularx}{\linewidth}{XXXX}
            \includegraphics[width=\linewidth]{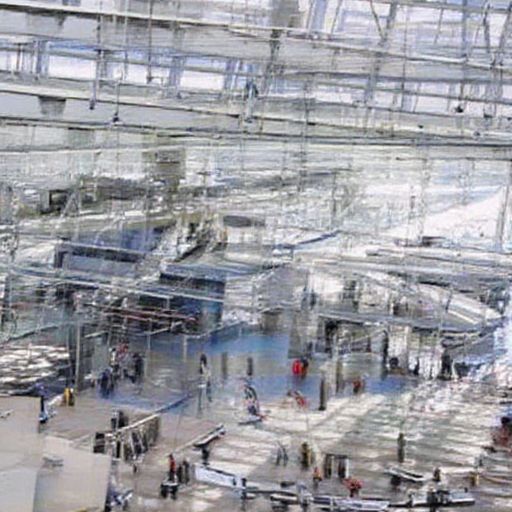} &
            \includegraphics[width=\linewidth]{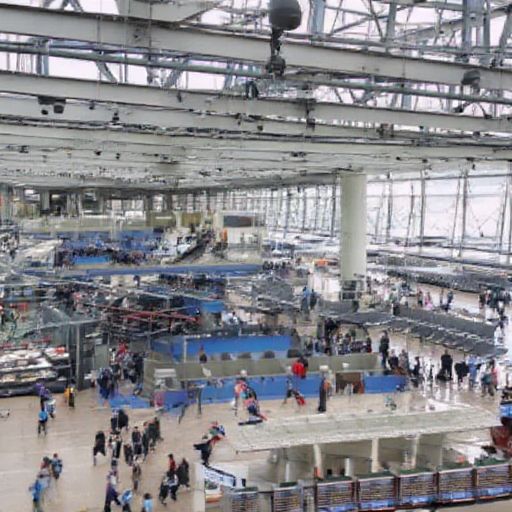} &
            \includegraphics[width=\linewidth]{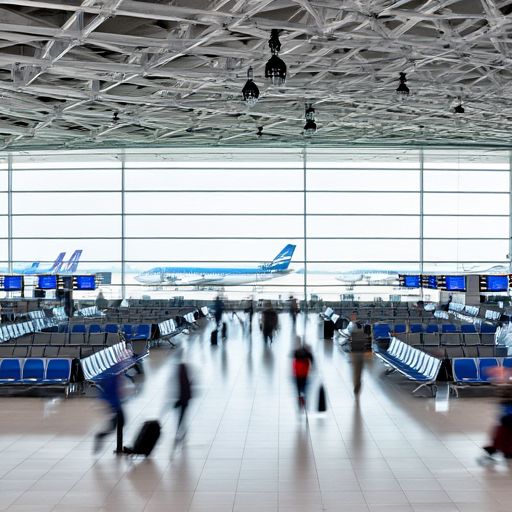} &
            \includegraphics[width=\linewidth]{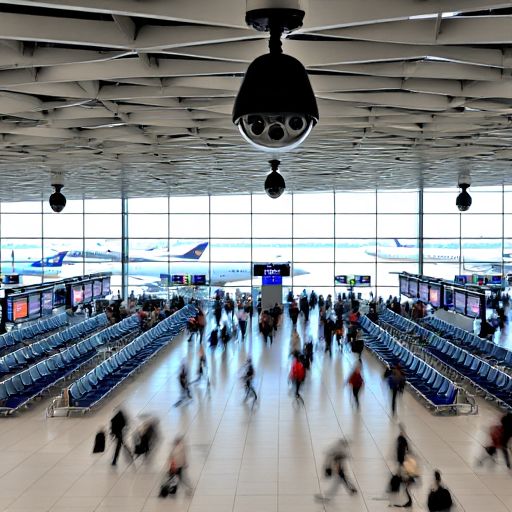}
        \end{tabularx}
        \vspace{-4pt}
        \caption{\textbf{DPG prompt 250:} A busy airport terminal with windows showing airplanes, seating areas, flight displays, and black surveillance cameras overhead.}
        \vspace{4pt}
    \end{subfigure}

    \captionsetup{justification=centering}
    \caption{
    Curated text-to-image comparisons on DPG benchmark prompts. \\
    Prompts are shortened for paper presentation. \\
    (part 4 of 4)
    }
    \label{fig:t2i_dpg}
\end{figure}

%% file: fig/t2i_failure.tex
\begin{figure}[H]
    \centering
    \captionsetup{justification=raggedright,singlelinecheck=false}
    \setlength{\tabcolsep}{5pt}
    \scriptsize
    \begin{tabularx}{0.75\linewidth}{|X|X|X}
        FM & CAFM & CAFM+CFG4
    \end{tabularx}
    \setlength{\tabcolsep}{1pt}
    \begin{subfigure}[t]{0.75\linewidth}
        \begin{tabularx}{\linewidth}{XXX}
            \includegraphics[width=\linewidth]{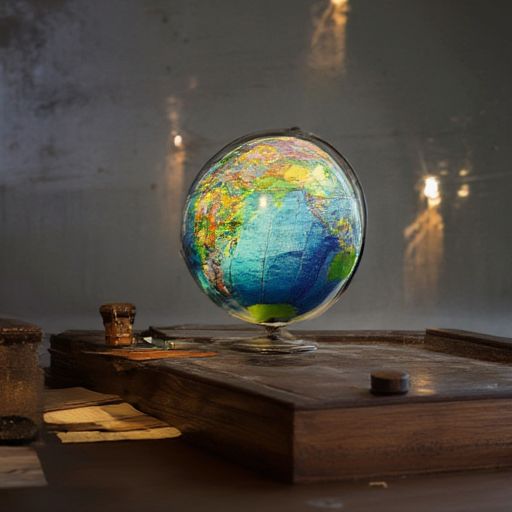} &
            \includegraphics[width=\linewidth]{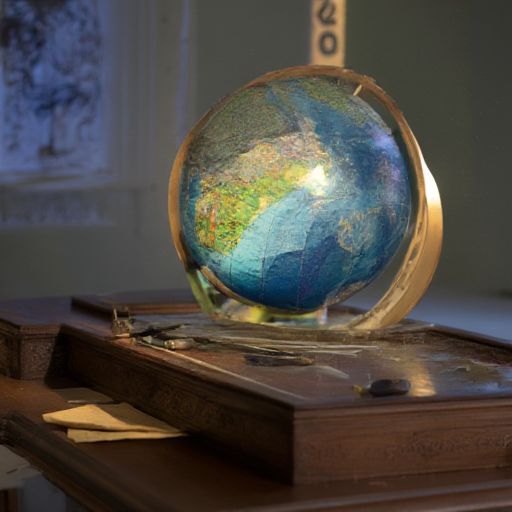} &
            \includegraphics[width=\linewidth]{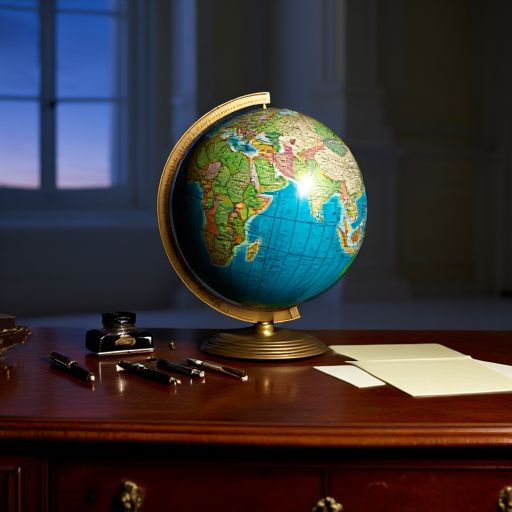}
        \end{tabularx}
        \vspace{-4pt}
        \caption{\textbf{DPG prompt 3:} An antique mahogany desk with a detailed globe and vintage pens}
        \vspace{4pt}
    \end{subfigure}

    \begin{subfigure}[t]{0.75\linewidth}
        \begin{tabularx}{\linewidth}{XXX}
            \includegraphics[width=\linewidth]{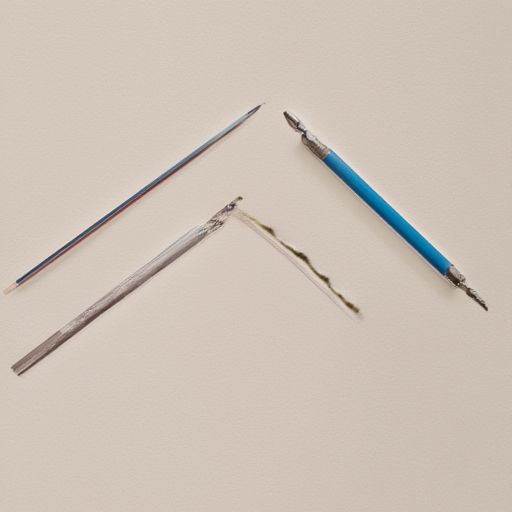} &
            \includegraphics[width=\linewidth]{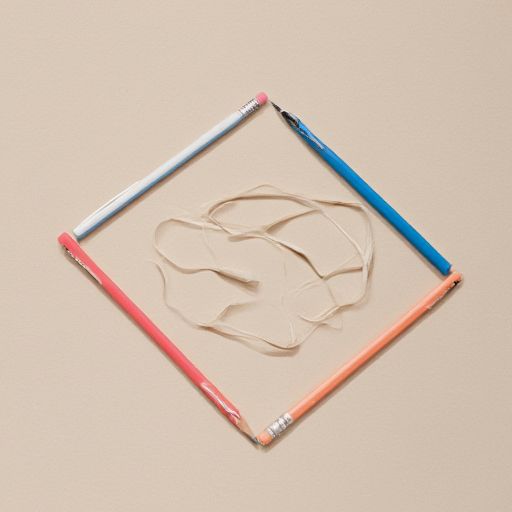} &
            \includegraphics[width=\linewidth]{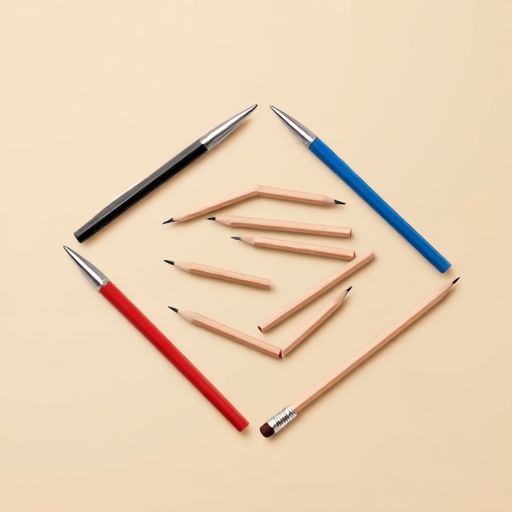}
        \end{tabularx}
        \vspace{-4pt}
        \caption{\textbf{DPG prompt 4:} Four pens forming a rectangle on a beige desk, with five pencils arranged in a circle at the center}
        \vspace{4pt}
    \end{subfigure}

    \begin{subfigure}[t]{0.75\linewidth}
        \begin{tabularx}{\linewidth}{XXX}
            \includegraphics[width=\linewidth]{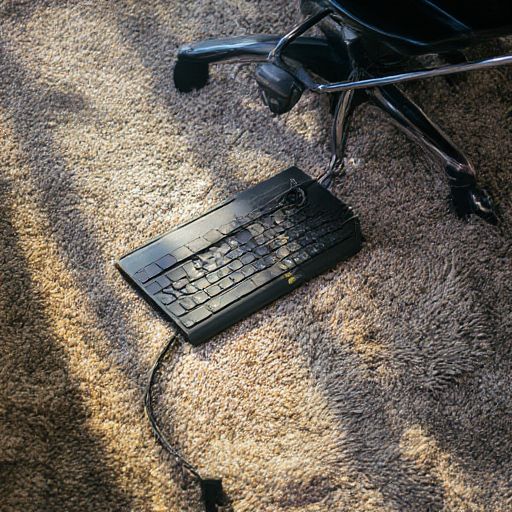} &
            \includegraphics[width=\linewidth]{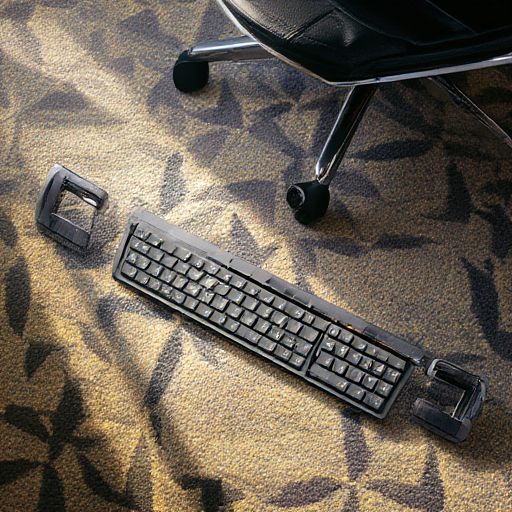} &
            \includegraphics[width=\linewidth]{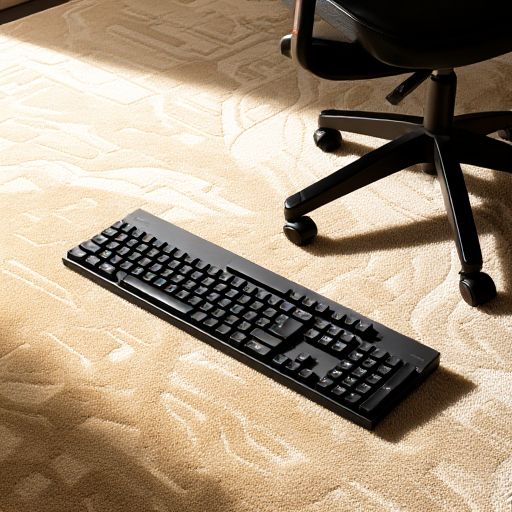}
        \end{tabularx}
        \vspace{-4pt}
        \caption{\textbf{DPG prompt 13:} A black keyboard resting diagonally on a beige carpet in a sunlit home office, with a nearby office chair.}
        \vspace{4pt}
    \end{subfigure}

    \begin{subfigure}[t]{0.75\linewidth}
        \begin{tabularx}{\linewidth}{XXX}
            \includegraphics[width=\linewidth]{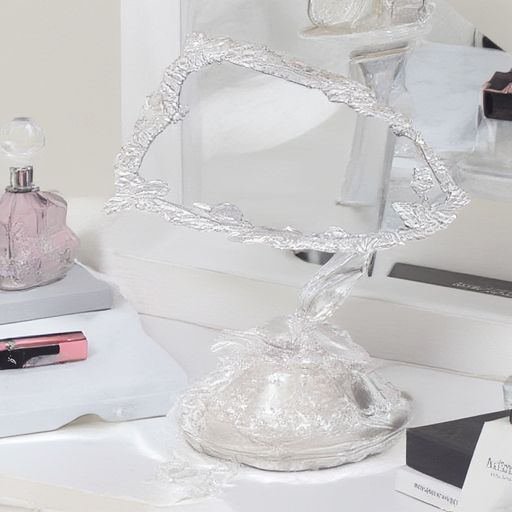} &
            \includegraphics[width=\linewidth]{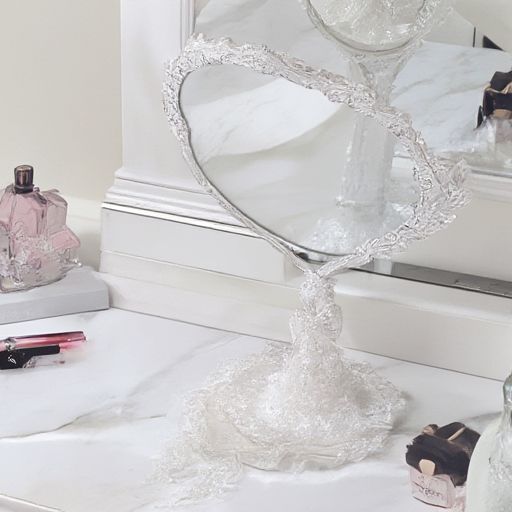} &
            \includegraphics[width=\linewidth]{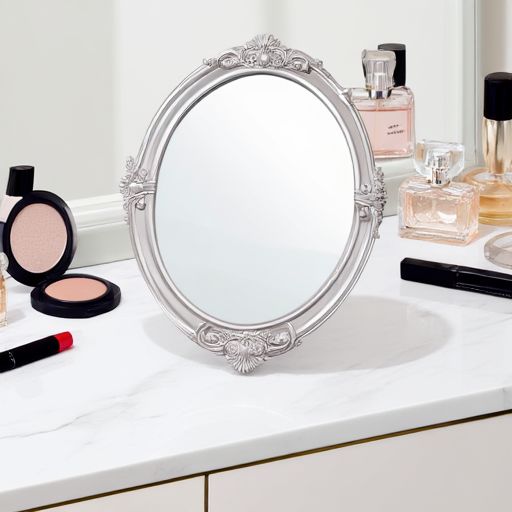}
        \end{tabularx}
        \vspace{-4pt}
        \caption{\textbf{DPG prompt 15:} An ornate silver makeup mirror on a white marble vanity surrounded by cosmetics and perfume bottles in natural light.}
        \vspace{4pt}
    \end{subfigure}
    
    \captionsetup{justification=centering}
    \caption{
    Failure cases for guidance-free text-to-image generation.
    }
    \label{fig:t2i_failure}
\end{figure}

%% file: fig/norm.tex
\begin{figure}
    \centering
    \begin{subfigure}[t]{0.46\linewidth}
        \includegraphics[width=\linewidth]{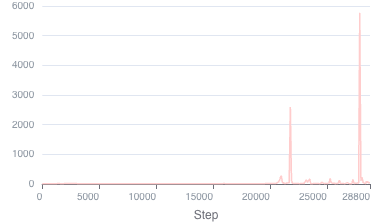}
        \caption{LayerNorm}
    \end{subfigure}
    \hfill
    \begin{subfigure}[t]{0.46\linewidth}
        \includegraphics[width=\linewidth]{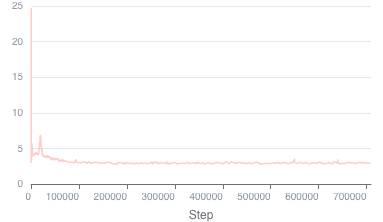}
        \caption{RMSNorm}
    \end{subfigure}
    
    \caption{Discriminator gradient norm.}
    \label{fig:norm}
\end{figure}